\documentclass[journal]{IEEEtran}
\usepackage{amsfonts}
\usepackage{graphicx}
\usepackage{subfigure}
\usepackage{multirow}
\usepackage{amssymb}
\usepackage{amsmath}
\usepackage{diagbox}
\usepackage{bbding}
\usepackage{makecell}
\usepackage{xcolor}
\usepackage{float}  
\usepackage[citecolor=blue, colorlinks]{hyperref}
\usepackage{subfiles}
\usepackage{booktabs}
\usepackage{soul}

\newcommand{\revise}[1]{\textcolor{black}{#1}}

\begin{document}

\title{3D Shape Completion on Unseen Categories: \\A Weakly-supervised Approach}

\author{Lintai Wu, Junhui Hou,~\emph{Senior Member, IEEE}, Linqi Song, and Yong Xu,~\emph{Senior Member, IEEE}
\thanks{This project was supported by the Hong Kong Research Grants Council under Grant 11219422 and Grant 11219324. \textit{(Corresponding author: Junhui Hou)}}
\thanks{L. Wu is with the Department of Computer Science, City University of Hong Kong, Hong Kong SAR, and also with the Bio-Computing Research Center, Harbin Institute of Technology, Shenzhen, Shenzhen 518055, Guangdong,
China. Email: lintaiwu2-c@my.cityu.edu.hk.}
\thanks{J. Hou and L. Song are with the Department of Computer Science, City University of Hong Kong, Hong Kong SAR. Email: jh.hou@cityu.edu.hk.}
\thanks{Y. Xu is with  the Bio-Computing Research Center,
Harbin Institute of Technology, Shenzhen, Shenzhen 518055, Guangdong,
China, and also with the Shenzhen Key Laboratory of Visual Object Detection and Recognition, Shenzhen, Guangdong
518055, China. Email: yongxu@ymail.com.}}

\maketitle

\begin{abstract}
3D shapes captured by scanning devices are often incomplete due to occlusion.  3D shape completion methods have been explored to tackle this limitation. However, most of these methods are only trained and tested on a subset of categories, resulting in poor generalization to unseen categories. In this paper, we propose a novel weakly-supervised framework to reconstruct the complete shapes from unseen categories. We first propose an end-to-end prior-assisted shape learning network that leverages data from the seen categories to infer a coarse shape. Specifically, we construct a prior bank consisting of representative shapes from the seen categories. Then, we design a multi-scale pattern correlation module for learning the complete shape of the input by analyzing the correlation between local patterns within the input and the priors at various scales. In addition, we propose a self-supervised shape refinement model to further refine the coarse shape. Considering the shape variability of 3D objects across categories, we construct a category-specific prior bank to facilitate shape refinement. Then, we devise a voxel-based partial matching loss and leverage the partial scans to drive the refinement process. Extensive experimental results show that our approach is superior to state-of-the-art methods by a large margin.
We will make the source code publicly available at \url{https://github.com/ltwu6/WSSC}.

\end{abstract}


\IEEEpeerreviewmaketitle

\section{Introduction}

\IEEEPARstart In recent years, with the popularity of 3D scanning equipment, 3D sensing technology has developed rapidly. 3D shapes, such as point clouds, meshes, and 3D voxels, have become increasingly valuable in various downstream tasks, such as SLAM, virtual reality, and autonomous driving \cite{liu2022perceptual,cui2021deep,zhu2022nice,ye2021meta,lei2023s,lohesara2023headset,tian2023superudf}.
However, obtaining high-quality 3D shapes directly from 3D scanning devices is challenging in practical applications. Due to self-occlusion as well as occlusion from environmental obstacles, the captured raw 3D data typically suffer from incompleteness, which may significantly impact downstream task performances.

To address the issues mentioned above, a number of deep learning-based shape completion methods have been proposed \cite{mittal2022autosdf,raopatchcomplete,yan2022shapeformer,xu2023cp3,yu2023adapointr,wu2023leveraging,wu2023scoda,ma2023collaborative,zhu2023csdn}. Existing shape completion methods typically adopt encoder-decoder architectures, which take a partial shape as input and aim to recover its complete geometric structure. During training, these methods optimize shape similarity between predicted shapes and corresponding complete ground truth using partial-complete pairs from specific categories. The inference process is also restricted to the same categories used in training.

In practical scenarios, however, many categories of partial scans lack complete ground truth due to limited resources such as human labor and time. Consequently, supervised shape completion methods exhibit poor generalization to unseen categories. To solve this problem, there have emerged some unsupervised shape completion methods \cite{xu2023unsupervised,li2020self,hong2023acl,wen2021cycle4completion,zhang2021unsupervised,cai2022learning}. These unsupervised shape completion methods can be divided into two types according to their training data. The first type employs data from the target categories for both training and testing, while the second type uses data from the same category but different domains for training. The former ignores shape prior information from the existing data categories, while the latter only focuses on completing data from different domains rather than different categories. Given partial-complete pairs from a set of categories, we argue that leveraging their prior knowledge can enable the completion of partial shapes from novel categories.

Currently, PatchComplete \cite{raopatchcomplete} is the only shape completion method specifically designed for unseen categories, based on the observation that 3D objects from different categories may share similar local structures, although they differ in global shapes. Therefore, PatchComplete learns patch-based local priors from training categories, which are further applied to infer complete shapes from unseen categories. However, PatchComplete still suffers from the following limitations. \textbf{First}, it requires four separate training sessions. Three patch-based prior-learning reconstruction models are trained individually to infer the attention scores between priors and inputs in Euclidean space, and then these three models are fused to train a decoder. This process is computationally expensive and time-consuming. \textbf{Second}, PatchComplete uses unified priors to train all categories, disregarding the fact that each category's data possesses its own distinct patterns. Unified priors struggle to accurately capture the shapes of each individual category. \textbf{Third}, PatchComplete is solely trained using data from seen categories without leveraging partial scans of unseen categories to learn their complete shapes.

In this paper, we propose a weakly-supervised method to infer complete shapes from unseen categories. 
We first devise an end-to-end Prior-assisted Shape Learning Network (PSLN) to predict a coarse complete shape in a supervised manner. \revise{Specifically, we construct a prior bank composed of some representative complete shapes from the seen categories in the training data. The PSLN consumes as input a partial shape along with all the priors stored in this bank.} Next, we design a Multi-scale Pattern Correlation (MPC) module to capture multi-scale local patterns associated with the input shape from the priors. Subsequently, we leverage these learned local patterns to effectively reconstruct the complete shape of the partial input. Different from PatchComplete, our PSLN computes the correlation between the priors and partial shape at the feature level instead of in Euclidean space. This design enables our model to be trained end-to-end in a single pass. Furthermore, since the input and priors may exhibit similar local patterns at varying scales, such as the resemblance between the large lampstand of some streetlights and the small legs of chairs, we extract priors' local patterns of different scales and compute the correlations between them and partial input at each layer of MPC. Experimental results demonstrate that our PSLN achieves superior performance compared to PatchComplete while significantly reducing computational and storage requirements.

Moreover, inspired by self-supervised point cloud completion methods \cite{mittal2021self,wang2021cascaded,gu2020weakly}, we leverage the information of partial scans from unseen categories to refine the coarse outputs. These self-supervised point cloud methods use the partial scans of the input object along with the partial matching loss to guide the model in predicting the complete shape. However, as the loss functions for self-supervised point cloud completion are unsuitable for voxel grids, we propose a voxel-based partial matching loss to guide the refinement process. Additionally, we believe that objects within a category exhibit more similar local patterns than those across different categories. Therefore, we design specific priors for each category during the refinement stage. 

Our approach utilizes complete shapes from seen categories and partial scans from unseen categories as supervision. Since this supervisory information is indirect and partial in guiding the complete shapes from unseen categories, our approach can be regarded as a form of weakly-supervised method.

In summary, the main contributions of this paper are two-fold:

\begin{itemize}

	\item We introduce a multi-scale Prior-assisted Shape Learning Network that effectively leverages priors from seen categories to reconstruct coarse complete shapes from unseen categories. This model effectively capitalizes on the multi-scale local patterns in the shapes of the seen categories. Moreover, the proposed model is end-to-end, requiring only a one-time training process.
 
	\item We propose a \revise{Category-specific Shape Refinement approach} that refines the coarse shapes with partial scans from unseen categories in a self-supervised manner. This model takes into account the shape characteristics of different categories and accurately performs category-specific refinement for the coarse shapes.

\end{itemize}

The remainder of this paper is organized as follows. Section \ref{related_work} reviews the previous supervised and unsupervised shape completion and reconstruction approaches. Section \ref{proposed_method} introduces our methods in detail. Section \ref{experiments} introduces our experimental settings, results, and analyses. Section \ref{conclusion} concludes this work.

\section{Related Work}
\label{related_work}

\subsection{Supervised 3D Shape Completion and Reconstruction}
The 3D shape completion and reconstruction methods can be divided into various categories according to the data representations, such as voxel grid \cite{sun2022patchrd, dai2017shape,tatarchenko2017octree,raopatchcomplete,chibane2020implicit,mittal2022autosdf,wallace2019few,cheng2023sdfusion}, point cloud \cite{yuan2018pcn,yang2018foldingnet,tchapmi2019topnet,xiang2021snowflakenet,xie2020grnet,yu2021pointr,zhang2021view,li2023proxyformer,zhou2022seedformer,zhang2023hyperspherical}, mesh \cite{dai2019scan2mesh,tang2019skeleton,tang2021skeletonnet} and continuous implicit functions \cite{chabra2020deep,park2019deepsdf,peng2020convolutional,wang2023lp}.

Voxel grids can be seen as an extension of the 2D-pixel representation to 3D image grids. It is widely used due to its simple and regular representation. The grids are filled with occupancy values or distance field values. 
The occupancy value of a voxel is typically binary, with a value of 0 representing an empty voxel and a value of 1 indicating an occupied voxel. In certain cases, the occupancy value can be represented as a continuous probability distribution, taking on values between 0 and 1, which reflect the probability that the voxel is occupied. 
Tatarchenko et al. \cite{tatarchenko2017octree} introduced a novel deep neural network architecture, termed Octree Generating Network (OGN), which can efficiently generate high-resolution 3D outputs via an octree-based representation. By utilizing hierarchical octree structures to represent 3D data, OGN is able to perform sparse computations and process large-scale 3D data efficiently. Sun et al. \cite{sun2022patchrd} proposed to exploit the local patches of partial inputs to infer the missing parts since the local patterns of an object may repeat. They first use a network to retrieve the most correlated local patches from inputs and then deform and blend them into the final outputs. \revise{Similar to \cite{sun2022patchrd}, we also use a binary voxel grid to represent 3D objects. However, unlike existing voxel-based methods, our approach focuses on completing partial scans of unseen categories. Moreover, we leverage priors from the training set to learn the shapes from unseen categories in a weakly-supervised manner. }

Distance field values encode the distance from a point in space to the closest surface of an object. Dai et al. \cite{dai2017shape} utilized a 3D-Encoder Predictor Network to generate an initial low-resolution shape prediction. Subsequently, a 3D shape synthesis method was devised to refine the preliminary output by integrating shape priors obtained from a complete shape database. Chibane et al. \cite{chibane2020implicit} proposed a 3D shape reconstruction and completion model named Implicit Feature Networks (IF-Nets), which can consume various 3D inputs, preserve objects' details and reconstruct articulated humans. They use multi-scale 3D vectors to encode 3D shapes and predict the occupancy at the feature space. \revise{Park et al. \cite{park2019deepsdf} introduced a technique for learning continuous signed distance functions (SDFs) to represent 3D shapes by training a deep neural network to map 3D points to a signed distance value.}
Mittal et al. \cite{mittal2022autosdf} introduced AutoSDF, a method that leverages an autoencoder to acquire a compact latent representation of 3D shapes, which in turn generates a Signed Distance Function (SDF) that encapsulates shape priors. The generated SDF is subsequently employed to steer the 3D completion, reconstruction, and generation process. 
Rao et al. \cite{raopatchcomplete} proposed that although 3D objects may differ in global shapes, they could share similar local structures across different categories. To address this, the authors proposed a framework named PatchComplete that learns patch-based local priors from training categories and leverages these priors to infer complete shapes from unseen categories. Nevertheless, on the one hand, the training process of PatchComplete is time-consuming. On the other hand, PatchComplete used a unified seen-category prior for all categories of data, which limited its generation ability.

Point cloud refers to a set of points that are spatially organized in 3D space. There exist three predominant methodologies for generating missing parts of point clouds. The first approach assumes that the 3D surfaces of point clouds can be obtained from the 2D manifold by performing successive mappings, known as folding-based methods. For instance, FoldingNet \cite{yang2018foldingnet} uses an encoder to extract the partial point cloud's global features and then deploys a 2D grid in conjunction with the global feature to reconstruct the complete point cloud via a decoder. However, FoldingNet is unable to fold a 2D grid into a complex shape. 
Instead of employing a fixed 2D grid as seeds, Yang et al. \cite{yang2021progressive} utilized a network to learn a high-dimensional seed to generate point clouds. Similarly, Pang et al. \cite{pang2021tearingnet} argued that the fixed 2D grid fails to capture the complex topologies of point clouds and used a network to learn the topology of the point cloud and tore the grid into patches to fit the shape of the object. 
The primary drawback of folding-based methods is that they predetermine a topology for the target object, which restricts the generated complete shape. Therefore, some approaches directly generate points in 3D space, known as point-based methods, instead of using a 2D manifold. Tchapmi et al. \cite{tchapmi2019topnet} designed a rooted-tree decoder to generate points as the tree grows hierarchically. Inspired by the process of snowflake generation, Xiang et al. \cite{xiang2021snowflakenet} proposed to split each parent point into multiple child points in a coarse-to-fine manner. Furthermore, they introduced a skip-transformer block to capture the relationship between the features of child points and parent points.

Mesh is a structural arrangement of vertices, edges, and faces that are interlinked to define a 3D shape. Dai et al. \cite{dai2019scan2mesh} proposed Scan2Mesh, an approach that leverages a generative neural network to infer the topology and geometry of a mesh from a partial and noisy range scan. Tang et al. \cite{tang2021skeletonnet} proposed SkeletonNet, a technique that can be trained to reconstruct 3D object surfaces from 2D RGB images whilst maintaining the topology of the mesh. This is accomplished by first predicting the object's skeleton and then using it to generate a surface mesh.

Continuous implicit functions represent objects as mathematical functions, such as neural network models, that implicitly describe the object's surface as the zero level-set of the function. Chabra et al. \cite{chabra2020deep} proposed to split the shape's surface into local regions and employ independent local latent codes to represent them. To solve the imbalanced distributions between different local regions, Wang et al. \cite{wang2023lp} proposed to represent an object with clusters of local regions along with multiple decoders, where each decoder is responsible for one cluster with the same pattern.

\begin{figure*}[htbp]	
	\centering
	\includegraphics[width=1.0\textwidth]{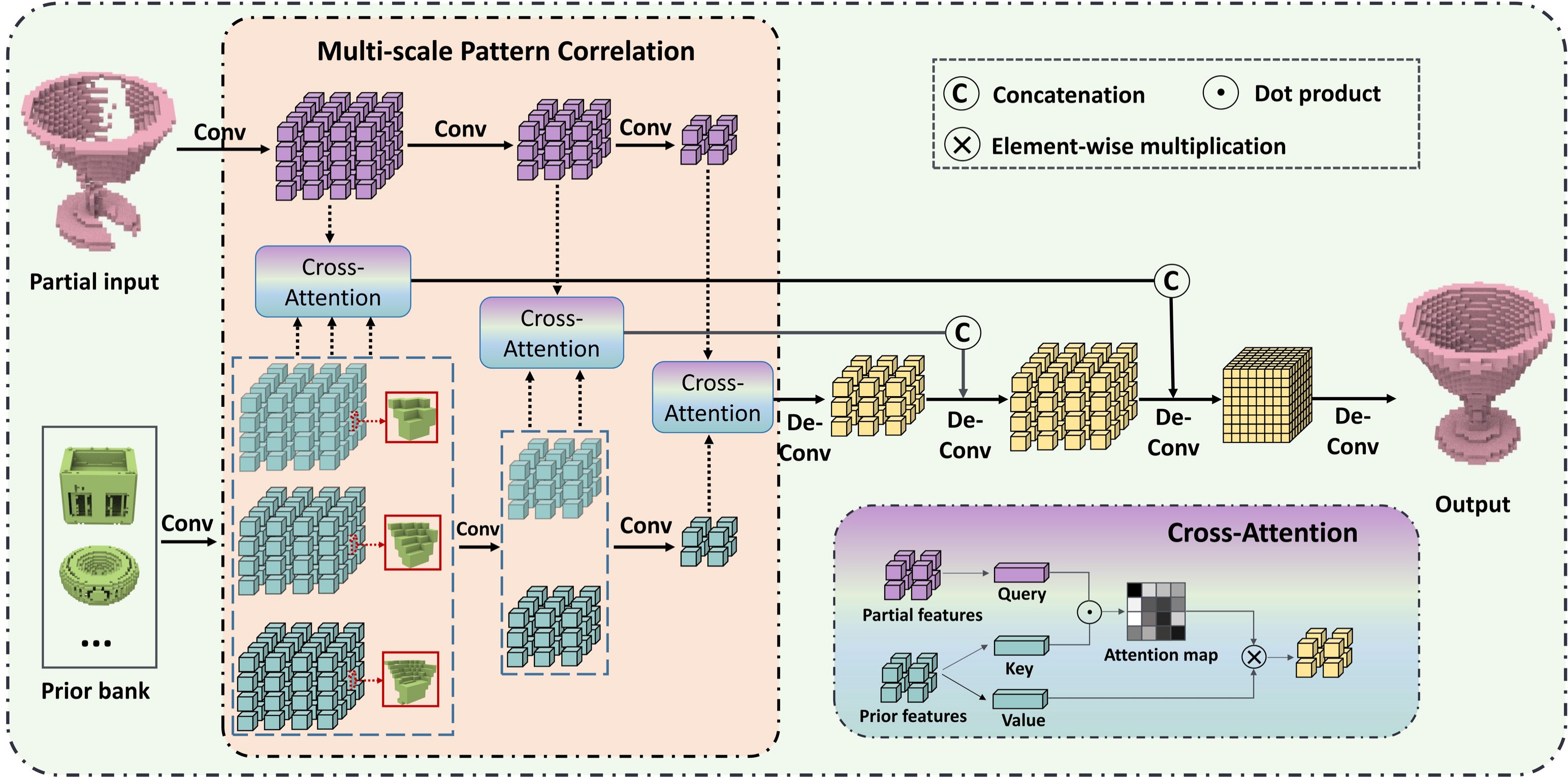}
	\caption{Flowchart of the Prior-assisted Shape Learning Network. ``Conv" and ``De-Conv" denote the convolution and de-convolution operation, respectively. The PSLN leverages a Multi-scale Pattern Correlation (MPC) module to capture the associated patterns for each local region of the partial input from the priors and subsequently use these patterns to infer the complete shape hierarchically. The MPC consists of two encoders and three Cross-Attention modules. The prior encoder utilizes multiple kernels of varying sizes to capture local structures at different scales. The Cross-Attention module identifies associated local patterns by calculating attention scores between the multi-scale local features of the partial input and the priors.}
	\label{overall}
\end{figure*}

\subsection{Unsupervised 3D Shape Completion and Reconstruction}
Unsupervised methods can be classified into cross-domain and in-domain methods based on whether the training and testing data come from the same domain or not.
Cross-domain methods \cite{chen2019unpaired,wen2021cycle4completion,wen2021cycle4completion,zhang2021unsupervised,cai2022learning}, also known as unpaired methods, do not necessitate the supervision that is provided by paired complete shapes and partial inputs during the training phase. Rather, they rely on a large dataset of complete and clean shapes to train a model that can capture prior 3D shape information. Subsequently, the pre-trained model is utilized to guide the completion of partial shapes. Note that the training and test data belong to the same categories but different domains. Chen et al. \cite{chen2019unpaired} introduced the first cross-domain unsupervised method for point cloud completion. They pre-trained two Auto-encoders, one for partial inputs and the other for complete point clouds, which were obtained from distinct data distributions. Next, they employed a generator and a discriminator to transform the global feature of the partial input and distinguish between the transformed global feature and that of the complete point clouds. The completion results were improved by minimizing the difference between the global feature of the partial input and that of the complete point cloud. Wen et al. \cite{wen2021cycle4completion} extended this method by simultaneously learning the transformation from partial inputs to complete point clouds and its inverse transformation to enhance the model's 3D shape understanding ability. Zhang et al. \cite{zhang2021unsupervised} introduced GAN Inversion to the point cloud completion task, which involved pre-training a GAN with a complete and clean dataset, followed by searching for a global feature that was best matched with the partial input in the latent space of GAN and optimizing it. Finally, Cai et al. \cite{cai2022learning} proposed a model that learned a unified latent space for partial and complete point clouds by modeling the incompleteness as the occlusion degree of the global feature of complete point clouds. The model's ability to understand the complete shape was improved by training it in the unified latent space. 
These methods can achieve good performance on test data from the same category as the training set, but they cannot generalize well to data from different categories.

The in-domain methods \cite{xu2023unsupervised,li2020self,hong2023acl,gu2020weakly,wang2021cascaded,mittal2021self} do not use partial-complete pairs from other domains, and the data categories for the training and test set are consistent. In this paper, we mainly introduce the self-supervised point cloud methods that we have used. Mittal et al. \cite{mittal2021self} proposed the first self-supervised point cloud completion method, PointPnCNet. PointPnCNet learns to infer complete shapes from partial scans with complete ground-truth shapes as supervision. It randomly removes a region from the partial scan, employs neural networks to predict the missing part, and leverages partial matching loss and partial scans as supervision. On the one hand, the network tends to extract common patterns among shapes in the same category. On the other hand, the model is unable to distinguish between synthetic removal and natural incompleteness. As a result, PointPnCNet is capable of successfully predicting complete shapes. Wang et al. \cite{wang2021cascaded} proposed a two-branch cascaded refinement network for point cloud completion. In order to learn to predict complete shapes without ground truth, they remove a portion of the area from partial inputs and combine them with the MixUp \cite{achituve2021self} strategy as inputs and supervisions.
Gu et al. \cite{gu2020weakly} proposed a weakly-supervised model that can simultaneously predict complete shapes and 6-DoF poses with the help of multi-view partial scans. Nonetheless, the performances of these in-domain unsupervised methods are constrained as they only utilize a limited number of partial shapes from fixed categories for training.

\section{Proposed Method}
\label{proposed_method}
We represent the shapes as 3D voxel grids. Denote by \(\mathbf{S}_{s}\in\mathbb{R}^{N\times N\times N}\) the partial shape of an object from the seen categories and \(\mathbf{G}\in\mathbb{R}^{N\times N\times N}\) the corresponding ground truth, \(\mathbf{S}_{u}\in\mathbb{R}^{N\times N\times N}\) the partial shape from unseen categories, where \(N \) represents the resolution of the 3D voxels. Our goal is to find a mapping \(f(\mathbf{S}_{u}|\mathbf{S}_{s}, \mathbf{G})=\mathbf{O}\) to make the reconstructed shape \(\mathbf{O}\) as close as possible to its ground truth.
\revise{Our approach consists of two stages, i.e., prior-assisted Coarse Shape Learning (CoSL) and Category-specific Shape Refinement (CaSR), which will be detailed as follows.}

\subsection{\revise{Prior-assisted Coarse Shape Learning}}
In real life, many 3D objects have similar local patterns. For example, most desks, benches, and lamps have sticks as support. These local structures are important for understanding the shapes of objects from unseen categories and can be used to infer missing parts of these objects. \revise{Inspired by this, we propose a Prior-assisted Shape Learning Network (PSLN) that leverages shape priors from seen categories to complete the partial scans from unseen categories.}

Specifically, we first construct a prior bank \({\mathbf{B}}\) that consists of several representative complete shapes from training data. Following PatchComplete, we use the mean-shift clustering algorithm \cite{cheng1995mean} to select \(M\) complete shapes from training data as priors. These priors contain typical local structures from each category in the training data.

Next, we design PSLN, an end-to-end network, to learn the local patterns of the partial shape from the priors.
As depicted in Figure \ref{overall},  we start by using two encoders to extract local features of the partial input and priors. \revise{Both of these encoders are composed of four 3D convolutional layers.
The features of the partial input and priors at the \(i\)-th layer are denoted by \(\mathbf{X}_i\in\mathbb{R}^{N_{i}\times N_{i}\times N_{i}\times C_i}\) and \(\mathbf{Y}_{i,m}\in\mathbb{R}^{N_{i}\times N_{i}\times N_{i}\times C_i}\), respectively, where \(m=1,2,...,M\), \(C\) represents the feature dimension.}
Then, we use cross-attention \cite{vaswani2017attention} to compute the weights between \(\mathbf{X}_i\) and \(\mathbf{Y}_{i,m}\) to represent their correlation. To be specific, we first split \(\mathbf{X}_i\) and \(\mathbf{Y}_{i,m}\) into vectors of size \(1\times 1\times 1\times C_i\). We define the vectors of \(\mathbf{X}_i\) as ``queries" and the vectors of \(\mathbf{Y}_{i,m}\) as ``keys" and ``values". Then, we compute the dot products of each query with all keys and divide them by \(C_i/2\). Next, we adopt a Softmax function to obtain the weights, i.e., 
\begin{equation}
w_{i,m}=\texttt{Softmax}\left(\frac{\mathbf{X}_i\cdot \mathbf{Y}_{i,m}}{d/2}\right)\,.
\end{equation}
Finally, we assemble the prior embeddings by
\begin{equation}
\mathbf{F}_{i}=\sum_{m=1}^M{w_{i,m}\cdot \mathbf{V}_{i,m}}\, ,
\end{equation}
where \(\mathbf{V}\) represents the ``values". \revise{In practice, \(\mathbf{V}\) is equal to \(\mathbf{Y}\), and we only compute \(\mathbf{F}_{\{2,3,4\}}\) for the features outputted by the last three encoder layers.}

Additionally, we observe that local patterns of partial shapes and priors may be similar at different scales. For example, the sticks of a chair backrest may have a shape similar to that of the pole of a street lamp, but their sizes are significantly different. \revise{Although PatchComplete comprises three stages to encode multi-resolution features, it maintains the same scale for partial shapes and priors across all resolutions. As a result, it overlooks the similarity of local structures at different scales.} To this end, we use convolutional kernels of various sizes to extract multi-scale features of priors and then calculate the attention between all multi-scale prior features and the features of the partial input. At the first layer of the prior encoder, we apply convolutional kernels of sizes \(7\times 7 \times 7\), \(5\times 5 \times 5\) and \(3\times 3 \times 3\), respectively. At the second layer, the sizes of convolutional kernels are \(5\times 5 \times 5\) and \(3\times 3 \times 3\). At the last layer, we only use a single convolutional kernel of size \(3\times 3 \times 3\) due to the large receptive field of this layer. The multi-scale features of each layer are concatenated and passed to the next layer. 
We call these blocks Multi-scale Structure Learning unit (MSL). 

The assembled prior embeddings \(\mathbf{F}_{i}\) are then decoded into the complete shape hierarchically.  Specifically,  we first directly decode \(\mathbf{F}_{4}\)  by a deconvolution layer.  Then \(\mathbf{F}_{3}\) and  \(\mathbf{F}_{2}\) are respectively concatenated with the decoded features and processed by two deconvolution layers. The model outputs the coarse complete shape after one more deconvolution layer and a Softmax layer. The training process of the PSLN is driven by optimizing L1 loss between the reconstructed coarse shape and its corresponding ground-truth complete shape. 

\revise{\textbf{Remark}. Although our PSLN follows the assumption of PatchComplete, which leverages shape priors to complete the partial scans from unseen categories, the prior-utilization strategy and network architecture of our PSLN differ from PatchComplete. Specifically, PatchComplete employs a prior-learning strategy. Concretely, it trains a set of learnable shape priors and learns patch-based attention maps between these priors and partial shapes during the patch-learning stage. The target shape is generated by directly fusing the patch volumes of the learned shape priors in 3D Euclidean space. By contrast, PSLN leverages a prior-assisted strategy that uses fixed-value shape priors without requiring learning to assist in the completion of partial scans. Our goal is only to learn the multi-scale structural correlation between partial scans and priors at the feature level, then decode the correlated prior features to infer the complete shape.}

\subsection{Category-specific Shape Refinement}
According to the idea of self-supervised point cloud completion methods \cite{mittal2021self,wang2021cascaded,gu2020weakly}, complete shapes can be inferred from the partial shapes of the target categories. These methods take a partial input and use partial shapes of the input object along with the partial matching loss to guide the model in reconstructing the complete shape. In this way, the model can discover many common patterns belonging to the shapes of the category and thus infer the complete shapes of the input.  Intuitively, one can apply the point-based partial matching loss,  which is a unidirectional CD loss defined as 
\begin{equation}
\label{point_partial_loss}
\begin{split}
	L_{part}(\mathbf{P}_{in}, \mathbf{P})=\frac{1}{|\mathbf{P}_{in}|}\sum_{\mathbf{p}_{in}\in \mathbf{P}_{in}} \mathop{\texttt{min}}\limits_{\mathbf{p}\in \mathbf{P}}||\mathbf{p}_{in}-\mathbf{p}||\,,
\end{split}
\end{equation}
to promote the existing parts of the predicted shape \(\mathbf{P}\) to be close to the input partial shape \(\mathbf{P}_{in}\).   
However, unlike the point-based representation, voxel representations do not impose constraints on the number of occupied voxels, and directly applying the partial L1 loss to voxel data tends to result in most voxels being occupied (see Figure \ref{ablation_visual_loss}).

To this end, we propose a voxel-based partial matching loss. First, we compute the partial L1 loss to constrain the existing parts of the predicted shape to be close to the input partial shape. Denote \(\mathbf{S}\) the partial scan and \(\mathbf{O}\) the refined result. The resolution of \(\mathbf{S}\) and  \(\mathbf{O}\) is \(N\), and the voxel value is 1 if this voxel is occupied; otherwise, the voxel value is 0. 
The partial L1 loss \(L_p\) is computed as
\begin{equation}
\label{l1_partial_loss}
    L_p=\frac{1}{N^3}\sum_{i=0}^{N^3}(|\mathbf{O}_i-\mathbf{S}_i|*\mathbf{A}_i),
\end{equation}
where \(\mathbf{A}\) is the binary mask for \(\mathbf{S}\) that indicates whether each voxel of \(\mathbf{S}\) is occupied or not. In practice, \(\mathbf{A}\) is equal to \(\mathbf{S}\).

Although the partial L1 loss can constrain the existing parts of the predicted result to be close to the partial input, it cannot guide the result to be a reasonable and complete shape. Due to the lack of constraints on the number of occupied voxels, the model tends to fill most voxels with “1”. The point-based partial matching loss can successfully guide the predicted shape because the predicted point number is fixed. That means the point cloud completion models should only predict the positions of a fixed number of points. Inspired by this, we fix the number of foreground voxels for all predicted results. Specifically, we design an occupancy loss that sums up all the predicted voxel values and makes them close to a predefined occupied voxel number \(H\). 
Since the number of occupied voxels varies across different objects, a fixed value of \(H\) for all objects may significantly compromise shape integrity or result in redundancy, particularly if \(H\) deviates significantly from the actual number of occupied voxels. To address this issue, we adaptively set \(H\) of the complete target shape to be \(\alpha\) times the occupied voxel number of the corresponding partial input:
\begin{equation}
    L_{s}=|\sum_{i=0}^{N^3} \mathbf{O}_i - H|, {\rm with}~H=\alpha \sum_{i=0}^{N^3} \mathbf{S}_i.
\end{equation}

Additionally, \(L_s\) can calculate the actual number of occupied voxels only when the voxel value is 0 or 1. However, in practice, the value of each voxel outputted by the network falls between 0 and 1 due to the Softmax function. To make the predicted voxel values close to 0 or 1, we propose a variance loss that computes the variance of the predicted voxel values and makes them larger as much as possible:
\begin{equation}
    L_v=\frac{1}{\texttt{var}(\mathbf{O})}\,.
\end{equation}

\begin{table*}[htbp]
	\centering
	\renewcommand\arraystretch{1.25}
	\caption{Quantitative Comparisons of different methods on the ShapeNet dataset in terms of IoU, F1, and CD($\times 100$). The best results under each metric are highlighted in bold. ``$\uparrow$" (resp. ``$\downarrow$") indicates the higher (resp. lower), the better.}
	\label{shapenet_results}
	\begin{tabular}{c|c|c|c|c|c|c|c|c|c|c}
		\toprule[1.2pt]
		Metric & Methods             & Average & Bag & Lamp & Bathtub & Bed & Basket	& Printer	& Laptop	& Bench \\ \hline 
            \multirow{6}*{IoU $\uparrow$} &
		Few-Shot \cite{wallace2019few}  & 0.218 & 0.193 & 0.149 & 0.278 & 0.172 &0.335 &0.184 &0.144 &  0.291  \\
		~ & IF-Nets \cite{chibane2020implicit}       & 0.197 & 0.138 & 0.150 &  0.276 & 0.178 & 0.237& 0.123& 0.250& 0.226 \\
		~ & AutoSDF \cite{mittal2022autosdf}       & 0.216 & 0.237 &  0.201 & 0.200 & 0.198 & 0.187 & 0.182 & 0.309 & 0.216   \\ 
             ~ & SDFusion \cite{cheng2023sdfusion}       &  0.311 & 0.342 & 0.304 & 0.299 & 0.299 & 0.264 & 0.324 & 0.309 & 0.344\\
		~ & PatchComplete \cite{raopatchcomplete}       & 0.429 & 0.391 & 0.374 & 0.493 & 0.368 & 0.490	& 0.396 & 0.461 & 0.456 \\  

            ~ & Ours & \textbf{0.474} & \textbf{0.450} & \textbf{0.380} & \textbf{0.542} & \textbf{0.426} & \textbf{0.509}	& \textbf{0.458} & \textbf{0.533} & \textbf{0.492} \\ \hline
            \multirow{6}*{F1 $\uparrow$} &
		Few-Shot \cite{wallace2019few}  & 0.345 &0.316  & 0.250 & 0.429 & 0.288 & 0.478& 0.306&0.244 & 0.445   \\
		~ & IF-Nets \cite{chibane2020implicit}       & 0.316 & 0.234 & 0.249 & 0.420  & 0.296 & 0.362	& 0.212	&0.394 	& 0.357\\
            ~ & AutoSDF \cite{mittal2022autosdf}       & 0.335 & 0.355 & 0.319 & 0.323 & 0.321 & 0.304 & 0.290 & 0.451 & 0.338  \\ 
            ~ & SDFusion \cite{cheng2023sdfusion}       & 0.466 & 0.504 & 0.458 & 0.457 & 0.455 & 0.412 & 0.483 & 0.458 & 0.504 \\
            ~ & PatchComplete \cite{raopatchcomplete}       & 0.587 & 0.552 & 0.534 & 0.650 & 0.530 & 0.637	& 0.555 & 0.617 & 0.616 \\
            ~ & Ours & \textbf{0.630} & \textbf{0.615} & \textbf{0.540} & \textbf{0.692} & \textbf{0.590} & \textbf{0.656}	& \textbf{0.618} & \textbf{0.683} & \textbf{0.648} \\ \hline
            \multirow{6}*{CD $\downarrow$}  &
		Few-Shot \cite{wallace2019few}  & 8.13 &6.05 & 12.11  & 5.75 & 9.70 & 5.58 &8.88 &11.39 & 5.61 \\
		~ & IF-Nets \cite{chibane2020implicit}       & 13.37  &  8.83 &  23.51 & 6.16  & 8.75  & 10.12 &  10.12	& 24.04 &  15.45\\
            ~ & AutoSDF \cite{mittal2022autosdf}       & 11.36 & 11.65  & 11.12 & 9.97 & 11.28 & 11.66 & 17.98 & 7.49 & 9.75  \\ 
            ~ & SDFusion \cite{cheng2023sdfusion}       & 7.54 & 8.81 & 7.74  & 6.32 & 7.73 & 8.90 & 11.48 & 5.34 & 4.00\\
            ~ & PatchComplete \cite{raopatchcomplete}    & 3.67 &3.98 &\textbf{4.97} &2.70&4.36  &2.82& 4.85 & 2.90 & 2.77  \\
            ~ & Ours & \textbf{3.02} & \textbf{2.75} & 5.01 & \textbf{2.18} & \textbf{3.40} & \textbf{2.53} & \textbf{3.54} & \textbf{2.18}	& \textbf{2.57} 
			 \\ \bottomrule[1.2pt]
	\end{tabular}
\end{table*}

The voxel-based partial matching loss is the weighted sum of partial L1 loss, occupancy loss, and variance loss:
\begin{equation}
    L_{VPM} = L_p + \gamma_1 L_{s} + \gamma_2 L_v \,,
\end{equation}
where \(\gamma_1\) and \(\gamma_2\) are the weights  of \(L_s\) and \(L_v\), respectively.

While the voxel-based partial matching loss facilitates the model in reconstructing the missing parts of the inputs by learning the typical patterns of each category, the model may predict significantly deviated results in these regions due to the lack of supervision. On the other hand, the coarse shapes, while not completely accurate, do exhibit a general similarity to the complete shapes of the inputs.
In order to guide the model in predicting more reasonable missing parts for the partial input, we use the missing parts of the coarse shapes outputted by CoSL as weak supervision. Specifically, we first remove the regions of the coarse shape that overlap with the partial input. We refer to the remaining part as the missing part and denote it as \(\mathbf{T}\). Then, we compute the partial L1 loss for \(\mathbf{T}\) and the predicted result. We term this loss as coarse shape loss and denote it as \(L_m\). Note that we only assign a small weight \(\lambda\) for this loss term. Overall, the total loss of our CaSR is:
\begin{equation}
    L=L_{VPM}+\lambda L_m \,.
\end{equation}

\begin{figure}[htbp]
	\centering  
	\subfigbottomskip=1pt 
        \setcounter{subfigure}{0}
	%\quad
	\subfigure[]{
		\includegraphics[width=0.22\linewidth]{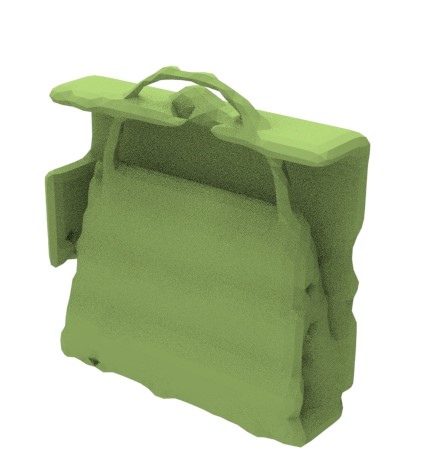}}
	\subfigure[]{
		\includegraphics[width=0.22\linewidth]{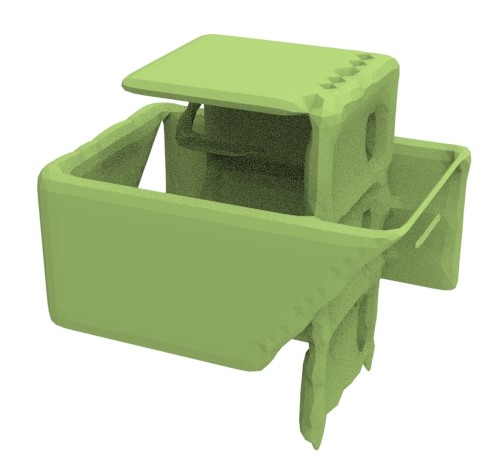}}
	\subfigure[]{
		\includegraphics[width=0.22\linewidth]{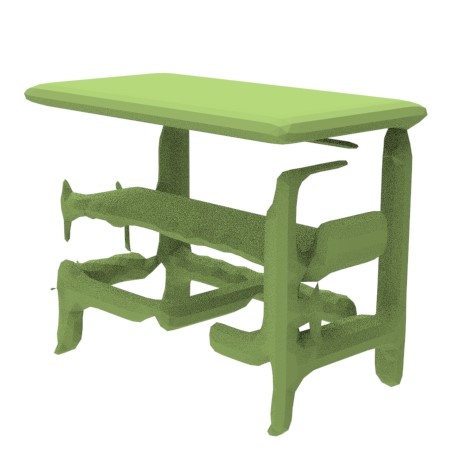}}
    \subfigure[]{
		\includegraphics[width=0.22\linewidth]{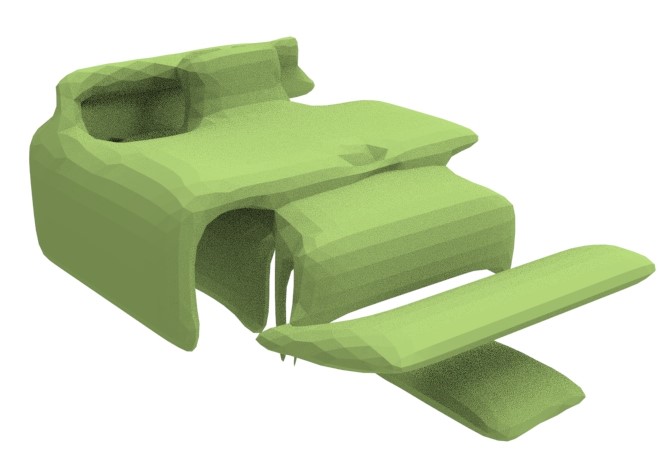}}
	\caption{Visualization of some category-specific priors. These priors contain typical structures specific to their respective categories. (a) bag, (b) basket,  (c) bench, (d) printer.}
	\label{visual_prior}
\end{figure}

We adopt the same framework as PSLN in the CoSL stage to refine our coarse outputs. The inputs are coarse shapes outputted by CoSL and category-specific priors.  PatchComplete only utilizes fixed priors from seen categories to infer the shapes of unseen categories. However, these fixed priors are not easily adaptable to all unseen data categories since each data category has its unique shape characteristics. For example, lamps typically exhibit a stick-shaped lampstand, while desks commonly feature a flat desktop and pillar-shaped legs. To address this issue, we introduce category-specific priors in the refinement stage, replacing the fixed priors derived from the training data. Specifically, to make use of more partial shapes, we first calculate the overlapping area between each pair of partial shapes from the target category. Then, we select M pairs with the least overlapping areas and concatenate them to construct the prior bank. Note that all partial shapes within this bank originate from distinct objects. In this way, the prior bank for each category contains various structures that are specific to that category. We visualize some priors in Figure \ref{visual_prior}.

\begin{table*}[htbp]
	\centering
	\renewcommand\arraystretch{1.25}
	\caption{Quantitative Comparisons of different methods on the ScanNet dataset in terms of IoU, F1, and CD($\times 100$). The best results under each metric are highlighted in bold. ``$\uparrow$" (resp. ``$\downarrow$") indicates the higher (resp. lower), the better.}
	\label{scannet_results}
	\begin{tabular}{c|c|c|c|c|c|c|c|c}
		\toprule[1.2pt]
		Metric & Methods             & Average & Bag & Lamp & Bathtub & Bed & Basket	& Printer \\ \hline 
            \multirow{6}*{IoU $\uparrow$} &
		Few-Shot \cite{wallace2019few}  & 0.245 & 0.199 & 0.239 & 0.259 &0.237  & 0.317&0.220  \\
		~ & IF-Nets \cite{chibane2020implicit}       & 0.222 & 0.133 & 0.183 & 0.281 & 0.235 & 0.343	& 0.156	 \\
		~ & AutoSDF \cite{mittal2022autosdf}       & 0.212 & 0.200 & 0.262 & 0.219 & 0.206 & 0.233 & 0.150   \\ 
            ~ & SDFusion \cite{cheng2023sdfusion}       & 0.322 & 0.355 & 0.348 & 0.324 &  0.359 & 0.287 & 0.258 \\
		~ & PatchComplete \cite{raopatchcomplete}       & 0.370 & 0.351 & \textbf{0.369} & 0.407 & 0.337 & 0.423	& 0.336  \\   
            ~ & Ours &\textbf{0.432} & \textbf{0.384} & 0.362 & \textbf{0.529}&\textbf{0.448} &\textbf{0.523} & \textbf{0.343}  \\ \hline
            \multirow{6}*{F1 $\uparrow$} &
		Few-Shot \cite{wallace2019few}  &0.389  &0.330  &0.382  & 0.408 &0.380  &0.474 &0.357   \\
		~ & IF-Nets \cite{chibane2020implicit}       &0.352  &0.233  &0.304  & 0.431  &0.376  & 0.501	& 0.266	\\
            ~ & AutoSDF \cite{mittal2022autosdf}       & 0.338 & 0.325 & 0.404 & 0.347 & 0.331 & 0.367 &  0.254  \\ 
            ~ & SDFusion \cite{cheng2023sdfusion}       & 0.477 & 0.517 & 0.509 & 0.480 &  0.520 & 0.436 & 0.402 \\
            ~ & PatchComplete \cite{raopatchcomplete}       & 0.532 & 0.515 & \textbf{0.533} & 0.569 & 0.498 & 0.582	& 0.497 \\
            ~ & Ours & \textbf{0.592} & \textbf{0.550} &0.526  & \textbf{0.682} & \textbf{0.613} & \textbf{0.674}	& \textbf{0.504}  \\ \hline
            \multirow{6}*{CD $\downarrow$} &
		Few-Shot \cite{wallace2019few}  & 7.07 & 7.67 & 7.54 & 6.46 & 8.25 &4.76 & 7.74 \\
		~ & IF-Nets \cite{chibane2020implicit}       &9.46 &10.26 &15.76 &7.19 &9.13 & 4.89	& 9.51 \\
            ~ & AutoSDF \cite{mittal2022autosdf}       &10.18 &10.49 & 6.96 &9.19 & 10.31  & 9.00 & 15.10  \\ 
            ~ & SDFusion \cite{cheng2023sdfusion}       & 7.50 & 7.41 & 5.40 & 5.96 &  5.94 & 8.45 & 11.83 \\
            ~ & PatchComplete \cite{raopatchcomplete}       & 4.21 & 4.70 & \textbf{3.84} & 3.63 & 4.37  & 3.62 & 5.09 \\
            ~ & Ours & \textbf{3.81} & \textbf{4.41} & 3.93 & \textbf{2.99} & \textbf{3.54} & \textbf{3.11} & \textbf{4.92}
			 \\ \bottomrule[1.2pt]
	\end{tabular}
\end{table*}

\section{Experiments} 
\label{experiments}

\subsection{Dataset and Evaluation Metric}
\label{dataset_metric}
\subsubsection{Dataset} To ensure a comprehensive evaluation of our method, we have conducted experiments on both synthetic and real-world data. The synthetic data we utilize are derived from ShapeNet \cite{chang2015shapenet}. The partial shapes are obtained by scanning the objects from ShapeNet at random angles. \revise{Following PatchComplete, we train our model with data from 18 categories (i.e., table, chair, sofa, cabinet, bookshelf, piano, microwave, stove, file cabinet, trash bin, bowl, display, keyboard, dishwasher, washing machine, pots, faucet, and guitar) and subsequently test it on data from other 8 unseen categories (i.e., bathtub, lamp, bed, bag, printer, laptop, bench, and basket). The training set and test set comprise 3202 and 1325 models, respectively, each of which consisted of 4 partial shapes scanned from random viewpoints.} 

For real-world data, we used real scans from ScanNet \cite{dai2017scannet}, which provides both partial scans and their complete ground truth. \revise{In this dataset, the training set contains 7537 models from 8 categories (i.e., chair, table, sofa, trash bin, cabinet, bookshelf, file cabinet, and monitor), while the test set contains 1191 models from 6 unseen categories (i.e.,  bathtub, lamp, bed, bag, basket, and printer). Each model has only one partial scan.}

The partial and complete shapes from both synthetic and real-world data are represented as voxel grids with the resolution of \(32\times 32\times 32\).
\subsubsection{Evaluation Metric}
We use the Intersection over Union (IoU) and F1 score to evaluate the similarity between the predicted voxel grids and the ground truth. The IoU measures the overlap between two sets of data, and the F1 score is the harmonic mean of precision and recall. In addition, following \cite{sun2022patchrd}, we convert the predicted voxel grids to point clouds and then employ the Chamfer distance (CD) to compare the differences between the predicted point sets and the ground truth.

\subsection{Implementation Details}
The resolution \(N\) of the voxel grids is 32 in our experiment. The number of priors \(M\) is 112.  The weight of the occupied voxel number \(\alpha\) is 2.5 for all categories. \(\lambda\) for \(L_m\) is 0.5. \(\gamma_1\) and \(\gamma_2\) of \(L_s\) and \(L_v\) are 1e-5 and 1e-4, respectively. 

Our model is trained from scratch on both the ShapeNet and ScanNet datasets without leveraging any pre-trained models. Four partial scans are utilized to compute the voxel-based partial matching loss for ShapeNet, while one partial scan is used for ScanNet.
We used Adam Optimizer to train our model. The batch size is 10. The initial learning rate of CoSL and CaSR is 1e-3 and 1e-4, respectively. The learning rate exponentially decays by 0.5 for every 50 epochs, and the models are trained for 120 epochs for both stages. 
The training process is conducted on an NVIDIA RTX 3090 GPU with Intel(R) Xeon(R) CPUs.

We train the CoSL on the unseen categories and determine the hyperparameters based on the IoU between the predicted results and the corresponding complete shapes from the seen categories. Subsequently, we train the CaSR on the unseen categories and select the hyperparameters using the IoU between the predicted results and the partial inputs, as complete shapes are typically unavailable for unseen categories in practical scenarios. The complete shapes from the unseen categories of our dataset are solely employed for evaluation purposes. In Section \ref{hyper_selections}, we present experimental results to validate the efficacy of this strategy.

\subsection{Comparison with State-of-the-Art Methods}
\label{section_comparison}

\begin{figure*}[htbp]
	\centering  
	\subfigbottomskip=1pt 
	\subfigure{
		\includegraphics[width=0.113\linewidth]{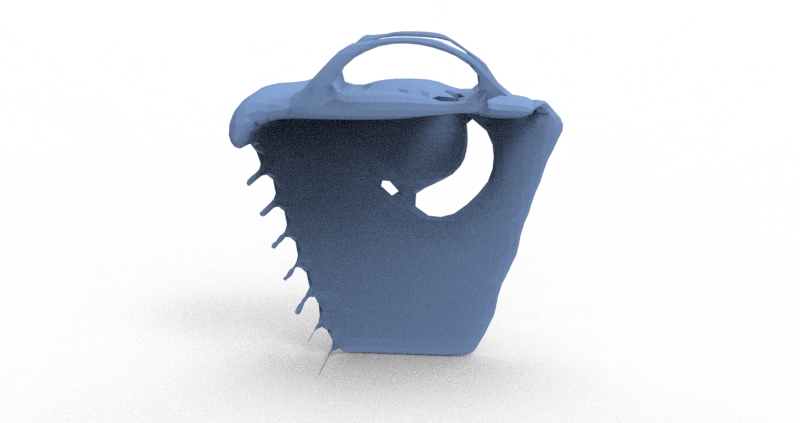}}
	\subfigure{
		\includegraphics[width=0.113\linewidth]{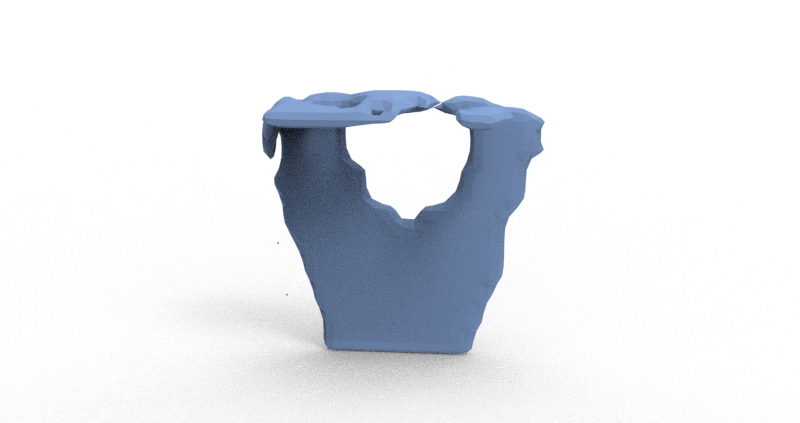}}
	\subfigure{
		\includegraphics[width=0.113\linewidth]{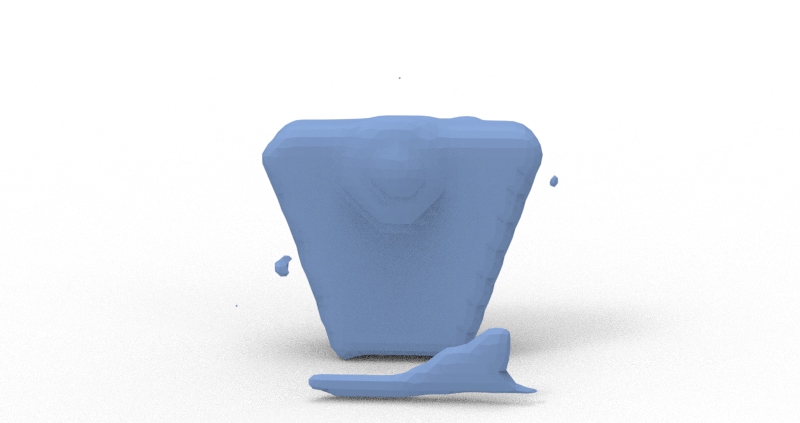}}
	\subfigure{
		\includegraphics[width=0.113\linewidth]{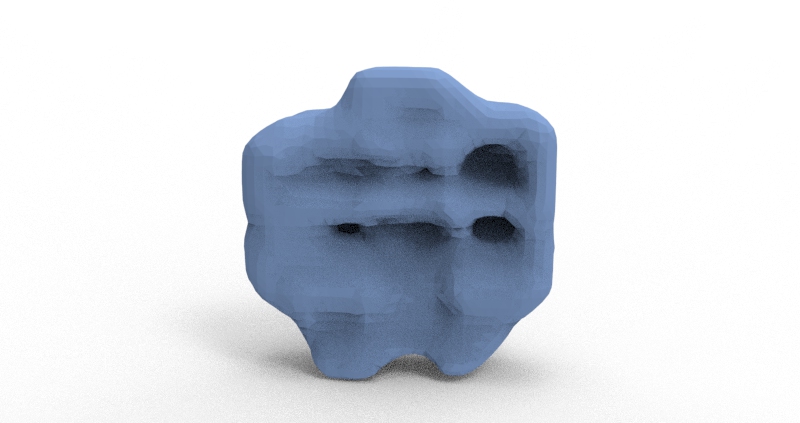}}
        \subfigure{
		\includegraphics[width=0.113\linewidth]{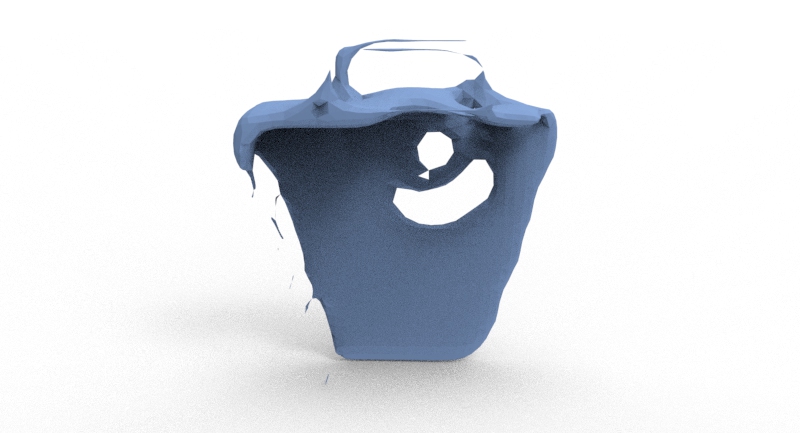}}
	\subfigure{
		\includegraphics[width=0.113\linewidth]{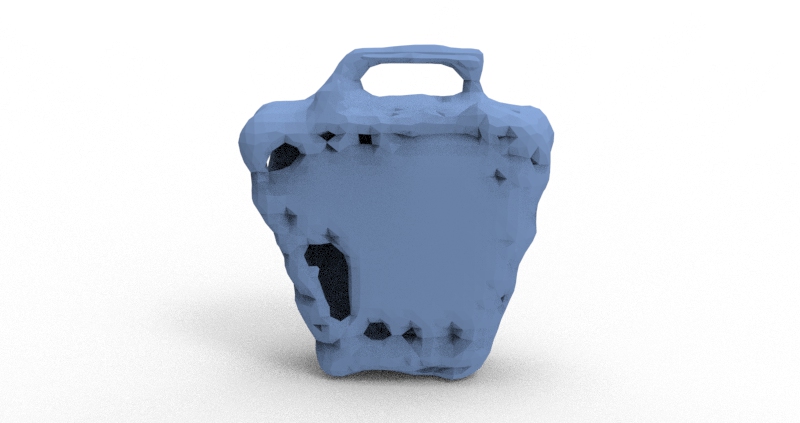}}
	\subfigure{
		\includegraphics[width=0.113\linewidth]{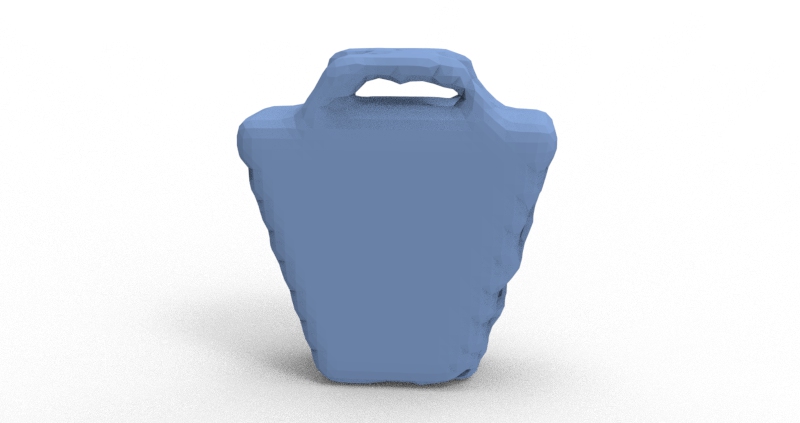}}
	\subfigure{
		\includegraphics[width=0.113\linewidth]{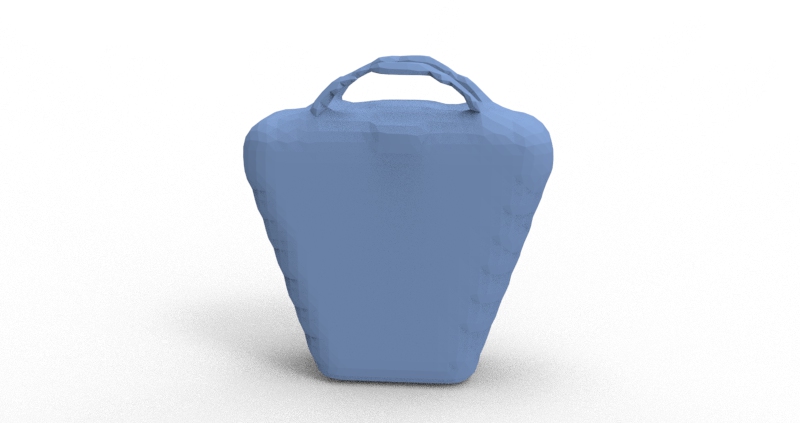}}
	\\
	\subfigure{
		\includegraphics[width=0.113\linewidth]{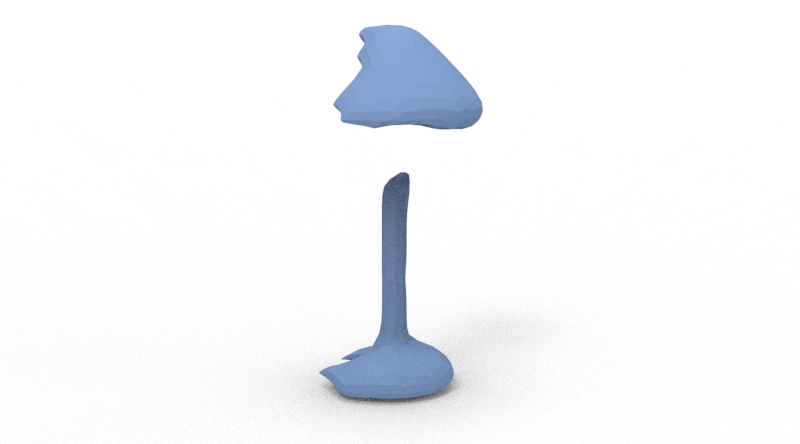}}
	\subfigure{
		\includegraphics[width=0.113\linewidth]{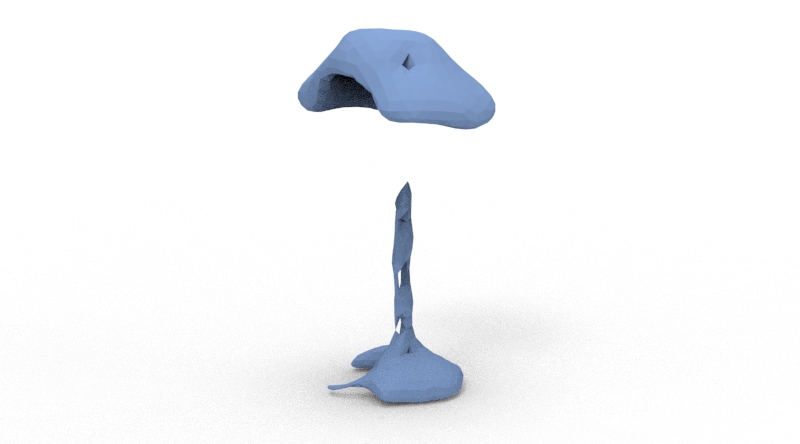}}
	\subfigure{
		\includegraphics[width=0.113\linewidth]{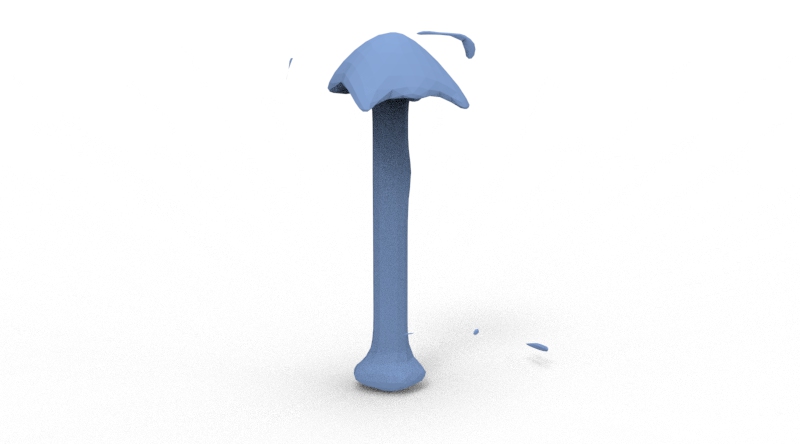}}
	\subfigure{
		\includegraphics[width=0.113\linewidth]{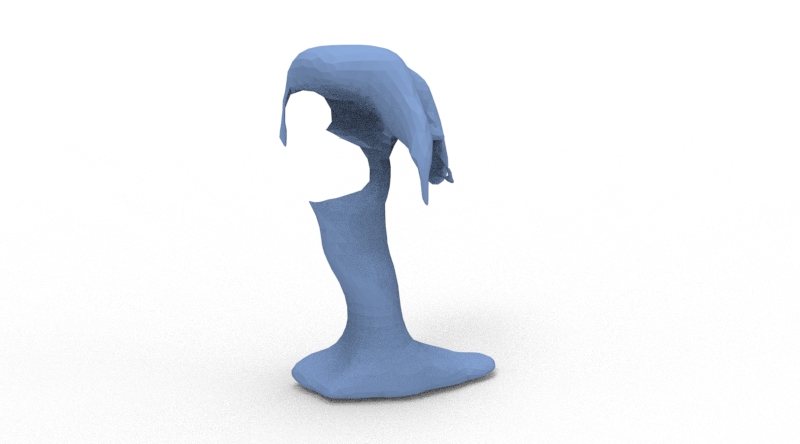}}
        \subfigure{
		\includegraphics[width=0.113\linewidth]{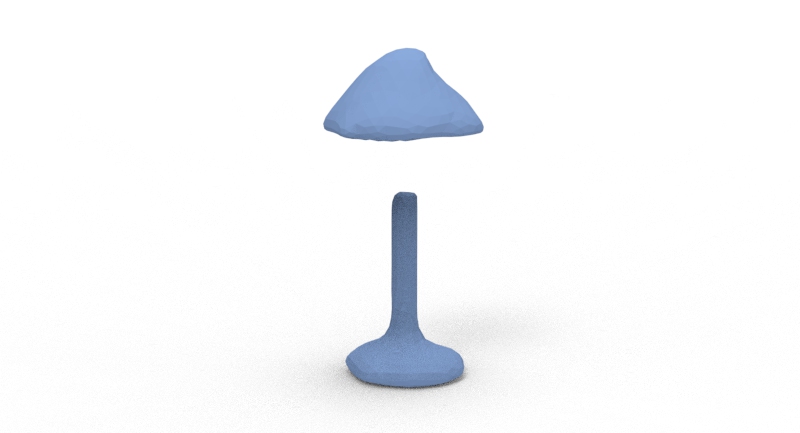}}
	\subfigure{
		\includegraphics[width=0.113\linewidth]{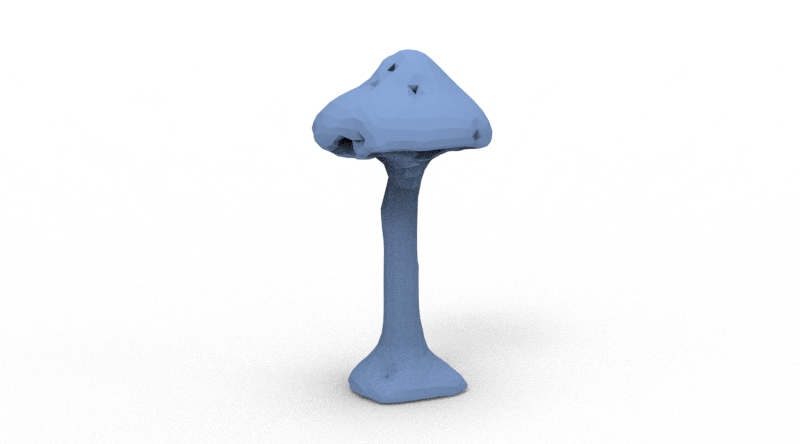}}
	\subfigure{
		\includegraphics[width=0.113\linewidth]{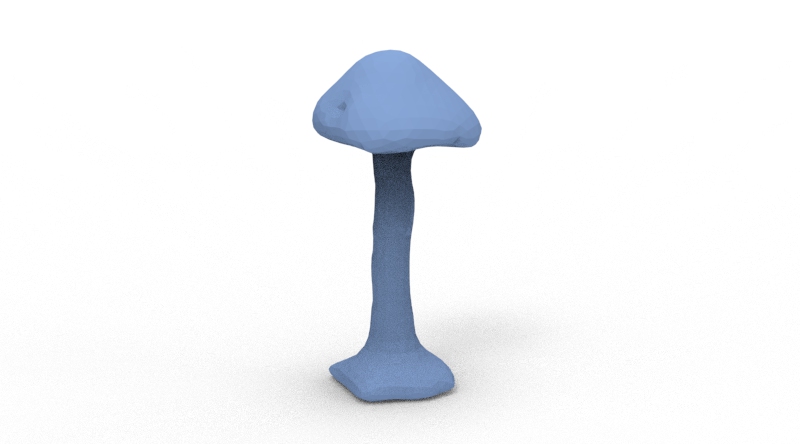}}
	\subfigure{
		\includegraphics[width=0.113\linewidth]{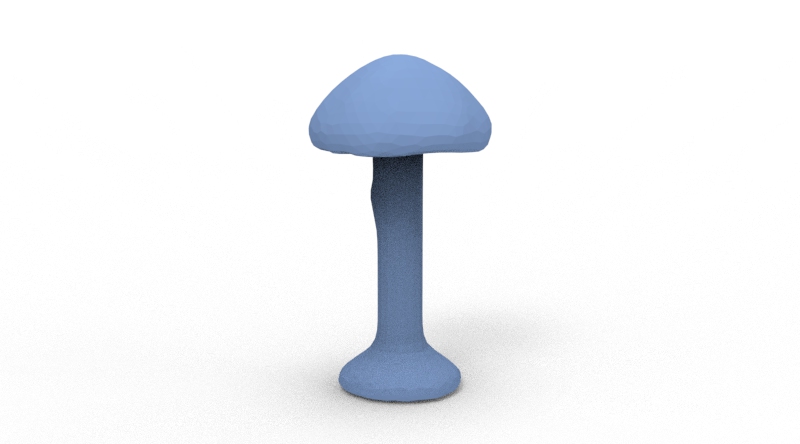}}
	\\
	\subfigure{
		\includegraphics[width=0.113\linewidth]{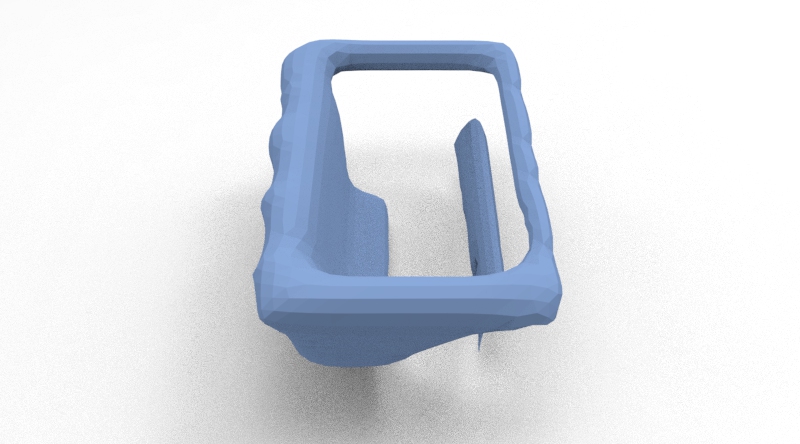}}
	\subfigure{
		\includegraphics[width=0.113\linewidth]{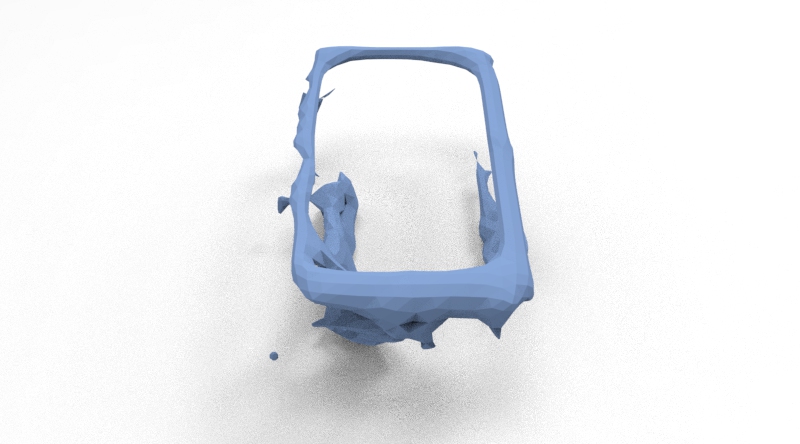}}
	\subfigure{
		\includegraphics[width=0.113\linewidth]{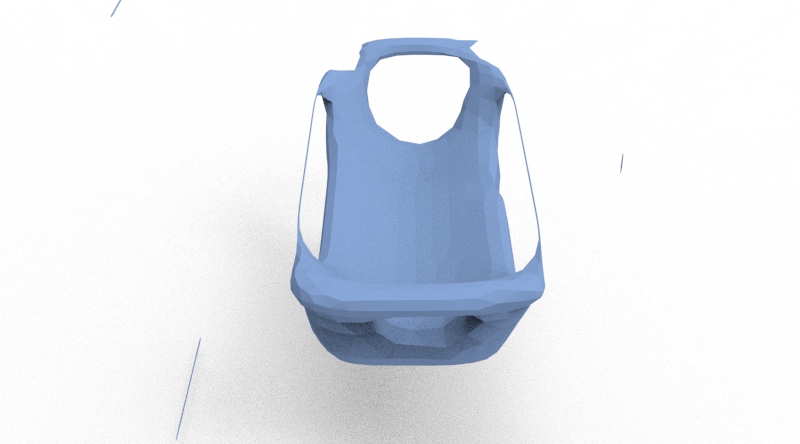}}
	\subfigure{
		\includegraphics[width=0.113\linewidth]{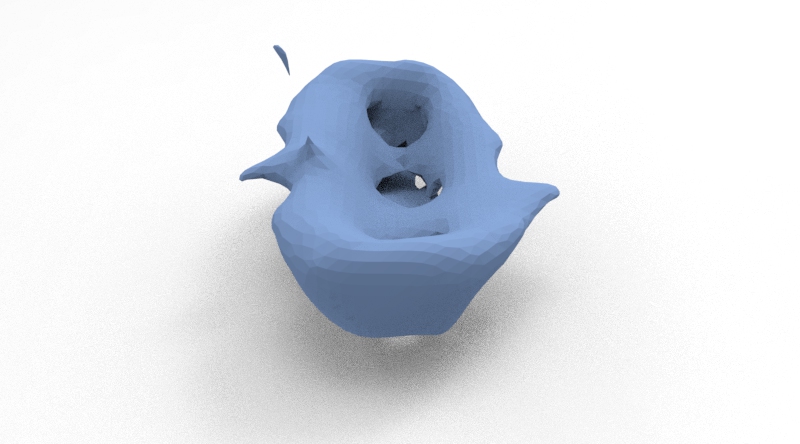}}
        \subfigure{
		\includegraphics[width=0.113\linewidth]{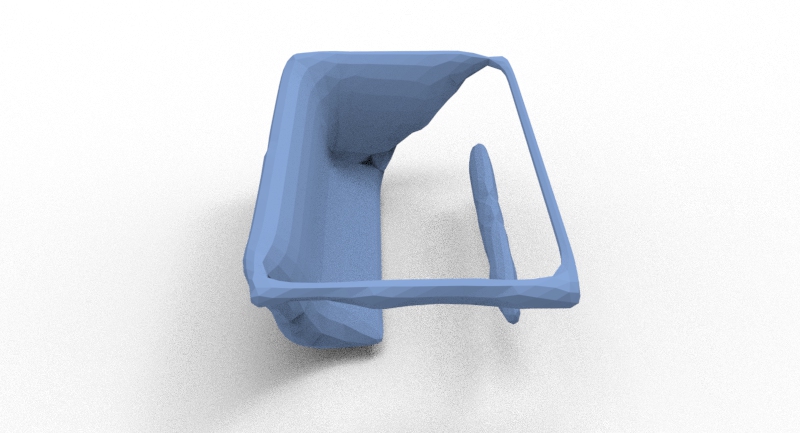}}
	\subfigure{
		\includegraphics[width=0.113\linewidth]{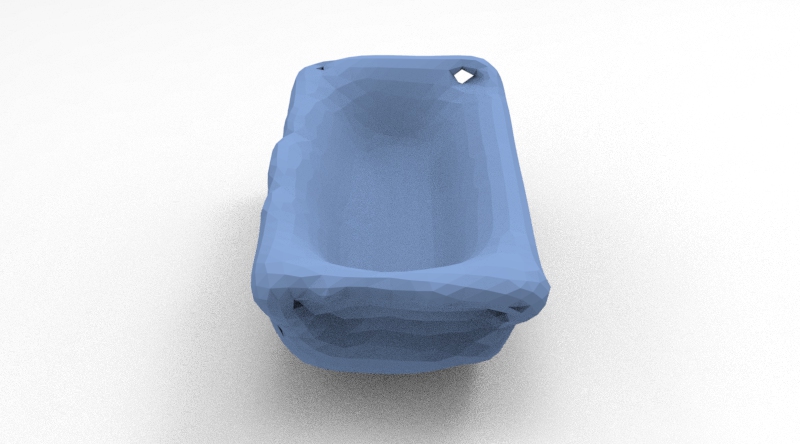}}
	\subfigure{
		\includegraphics[width=0.113\linewidth]{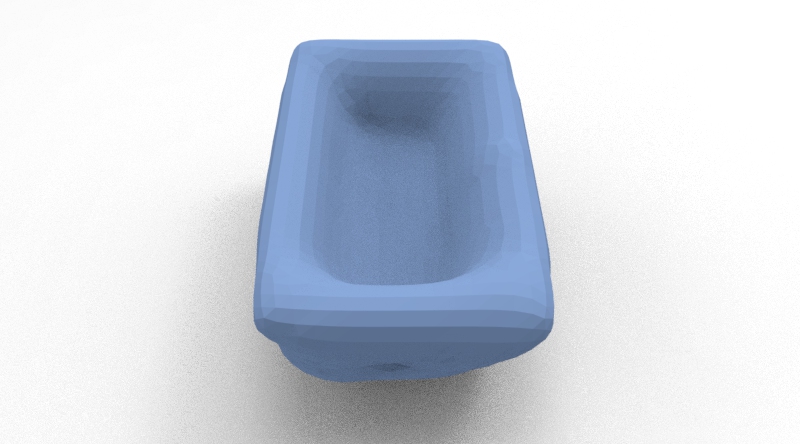}}
	\subfigure{
		\includegraphics[width=0.113\linewidth]{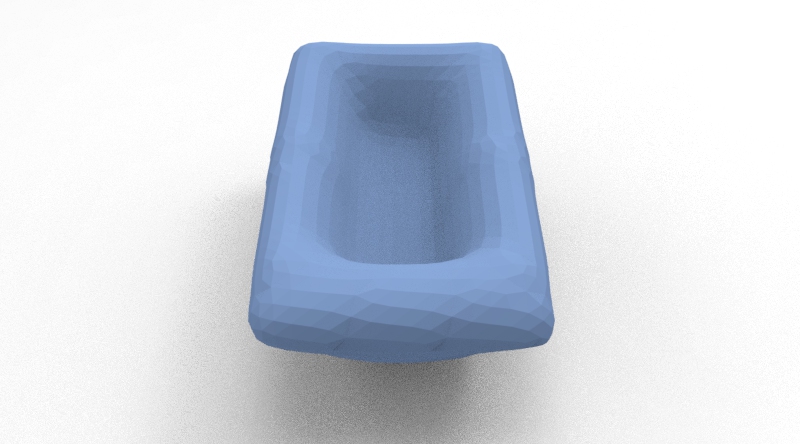}}
	\\
	\subfigure{
		\includegraphics[width=0.113\linewidth]{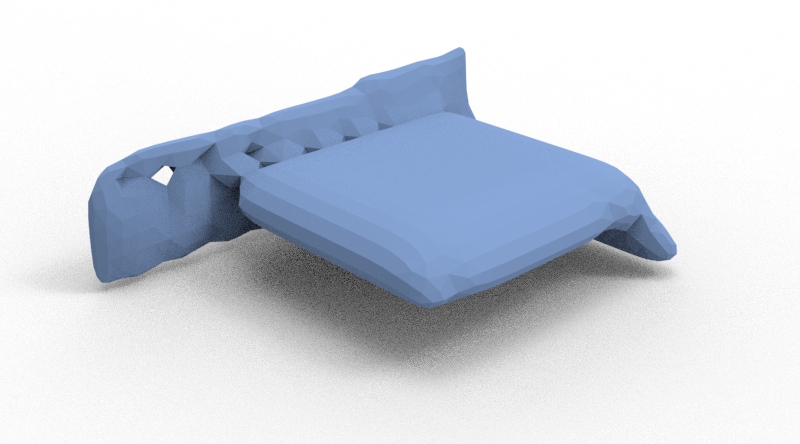}}
	\subfigure{
		\includegraphics[width=0.113\linewidth]{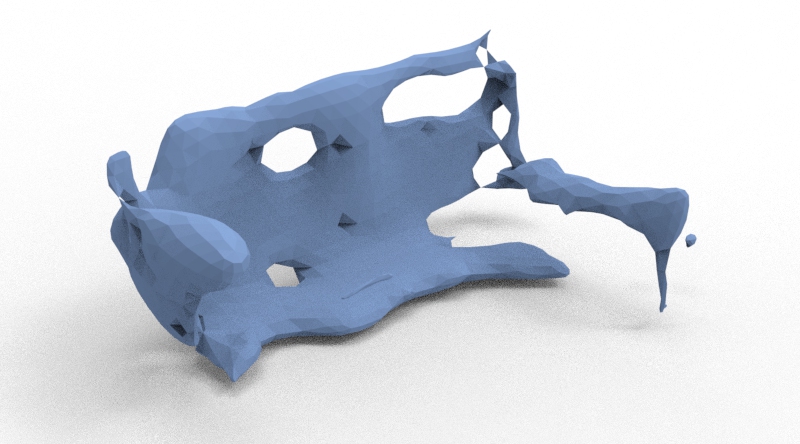}}
	\subfigure{
		\includegraphics[width=0.113\linewidth]{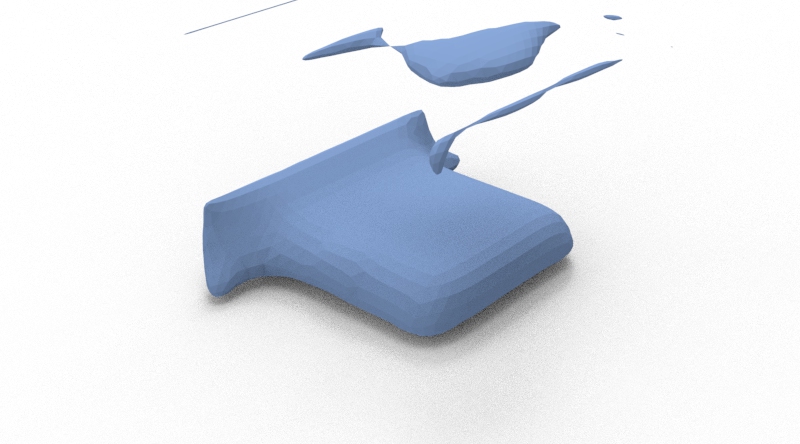}}
	\subfigure{
		\includegraphics[width=0.113\linewidth]{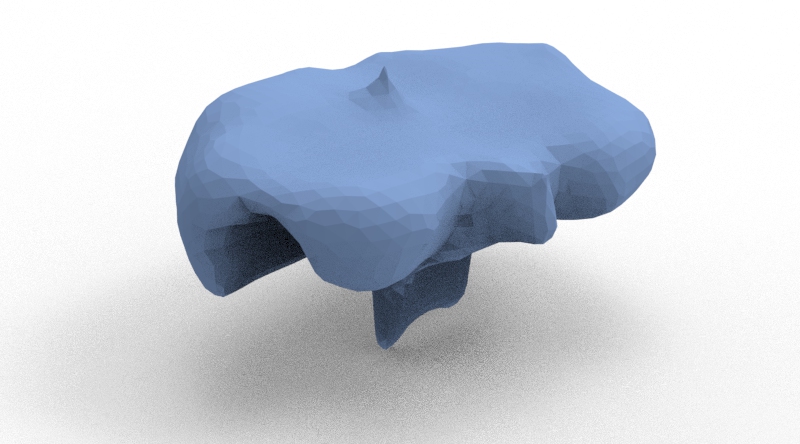}}
        \subfigure{
		\includegraphics[width=0.113\linewidth]{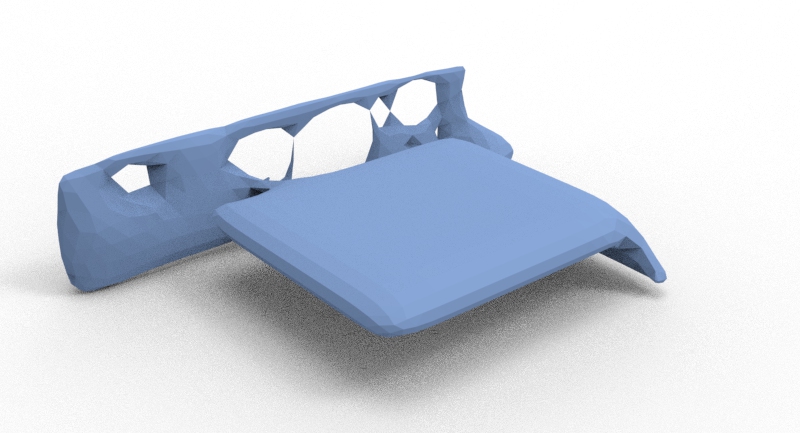}}
	\subfigure{
		\includegraphics[width=0.113\linewidth]{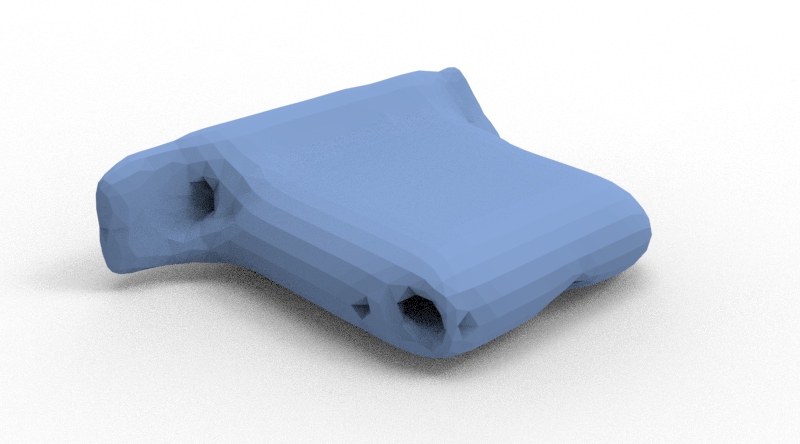}}
	\subfigure{
		\includegraphics[width=0.113\linewidth]{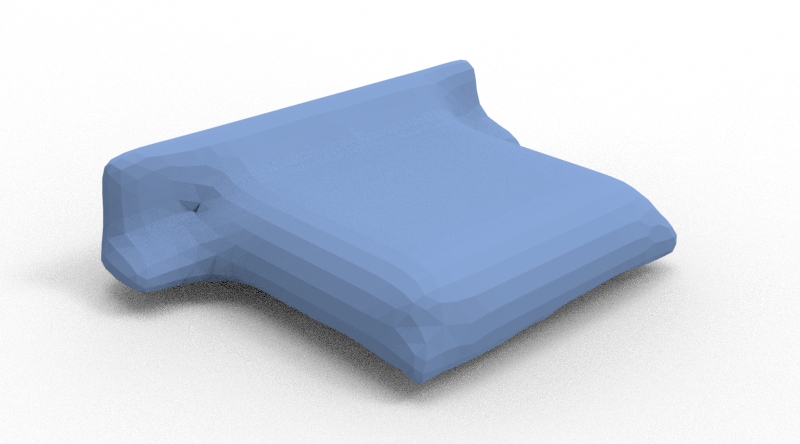}}
	\subfigure{
		\includegraphics[width=0.113\linewidth]{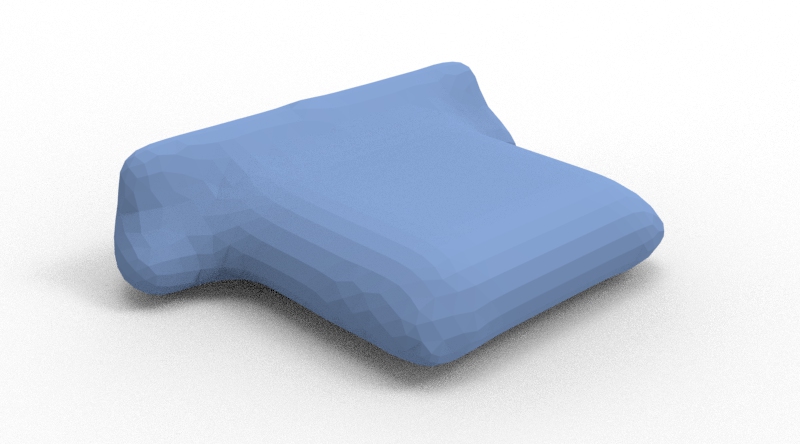}}
	\\
	\subfigure{
		\includegraphics[width=0.113\linewidth]{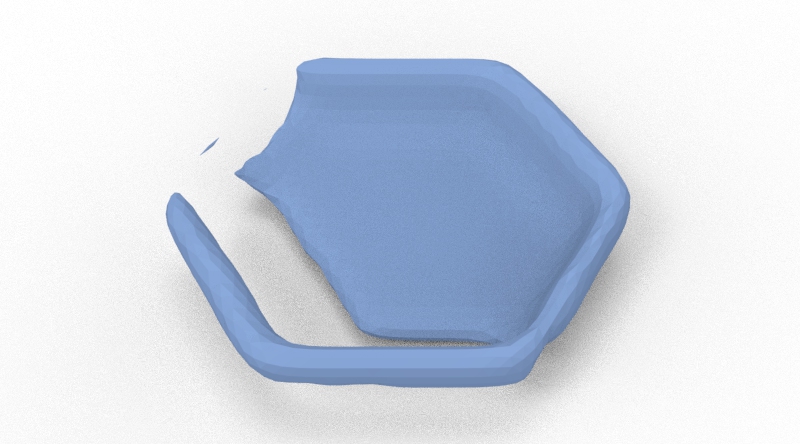}}
	\subfigure{
		\includegraphics[width=0.113\linewidth]{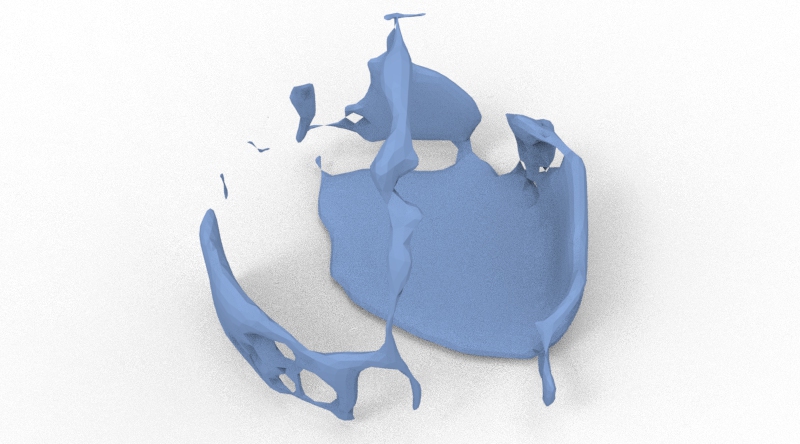}}
	\subfigure{
		\includegraphics[width=0.113\linewidth]{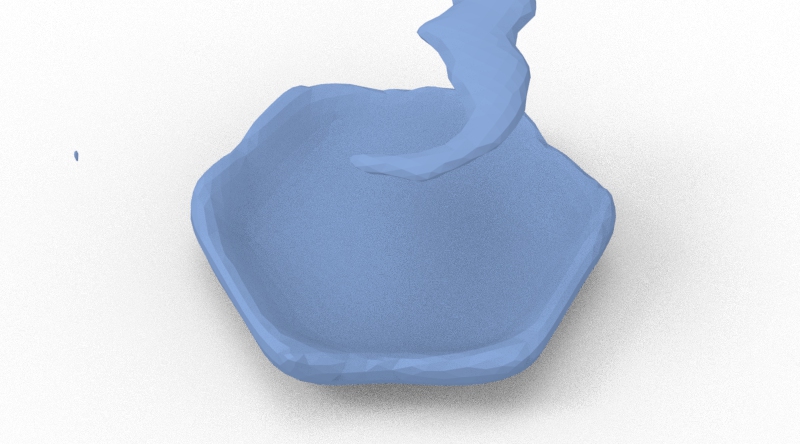}}
	\subfigure{
		\includegraphics[width=0.113\linewidth]{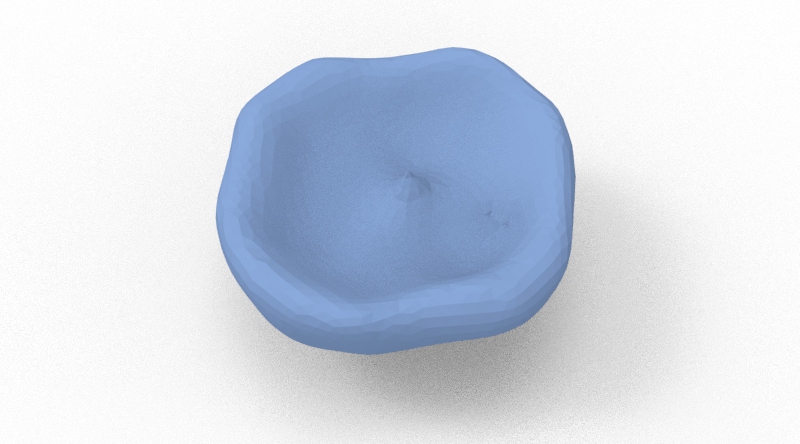}}
        \subfigure{
		\includegraphics[width=0.113\linewidth]{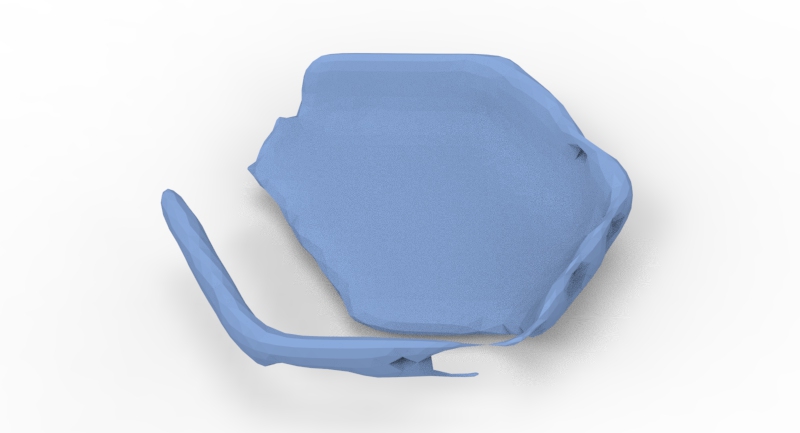}}
	\subfigure{
		\includegraphics[width=0.113\linewidth]{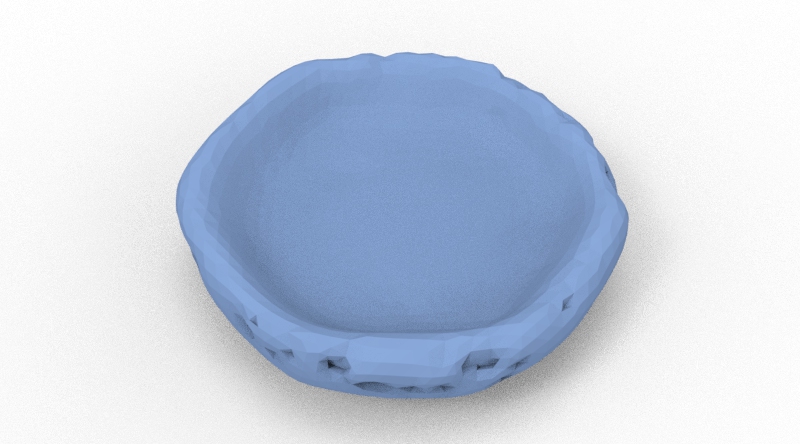}}
	\subfigure{
		\includegraphics[width=0.113\linewidth]{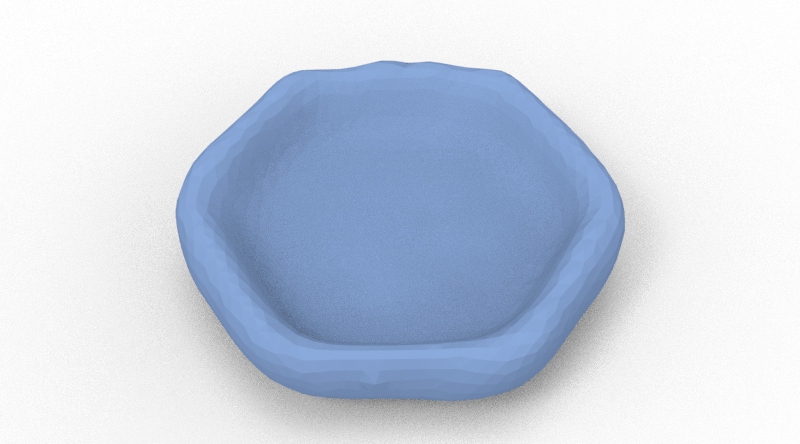}}
	\subfigure{
		\includegraphics[width=0.113\linewidth]{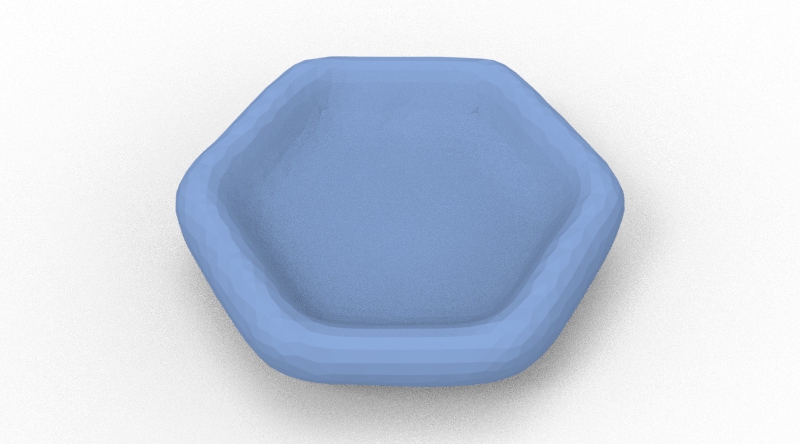}}
	\\
	\subfigure{
		\includegraphics[width=0.113\linewidth]{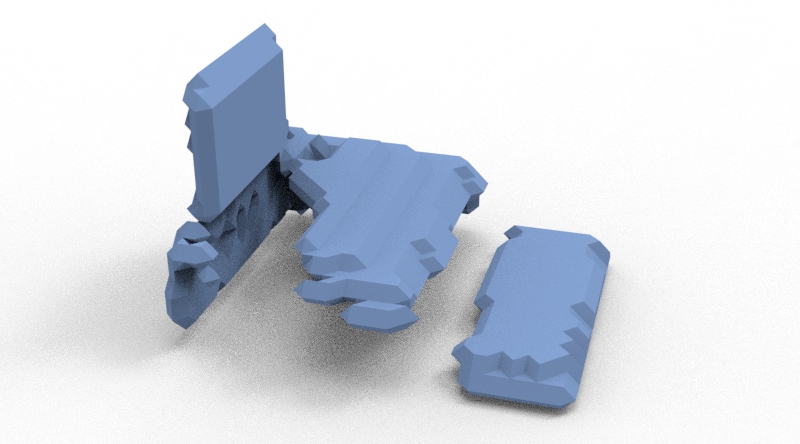}}
	\subfigure{
		\includegraphics[width=0.113\linewidth]{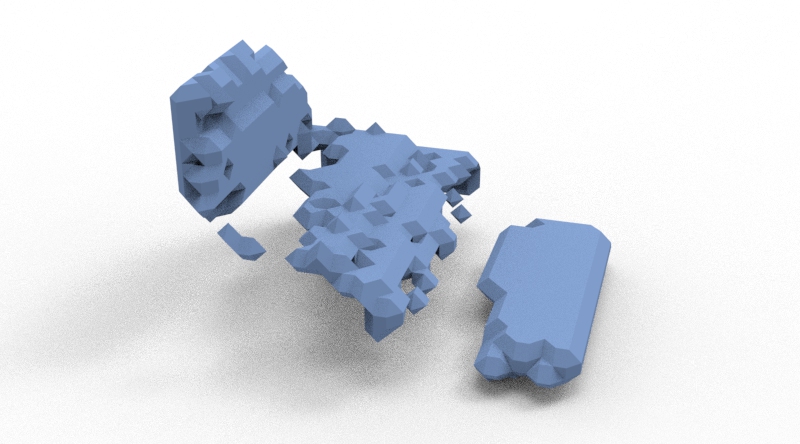}}
	\subfigure{
		\includegraphics[width=0.113\linewidth]{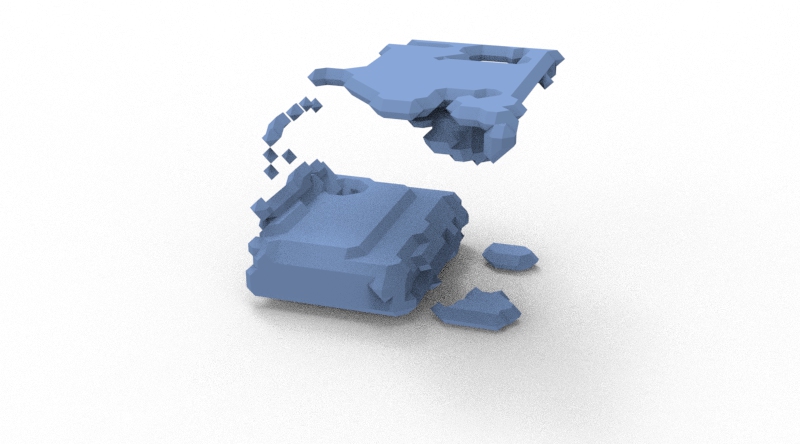}}
	\subfigure{
		\includegraphics[width=0.113\linewidth]{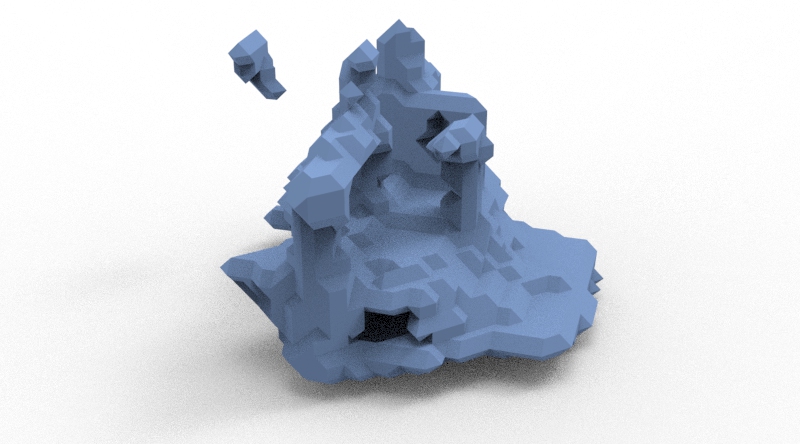}}
        \subfigure{
		\includegraphics[width=0.113\linewidth]{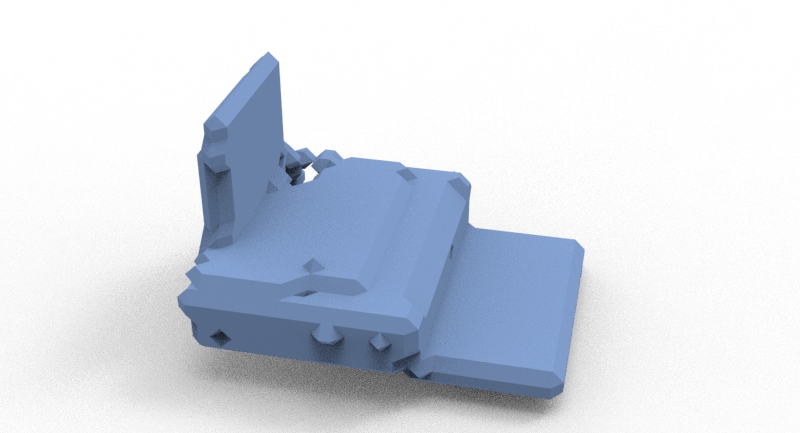}}
	\subfigure{
		\includegraphics[width=0.113\linewidth]{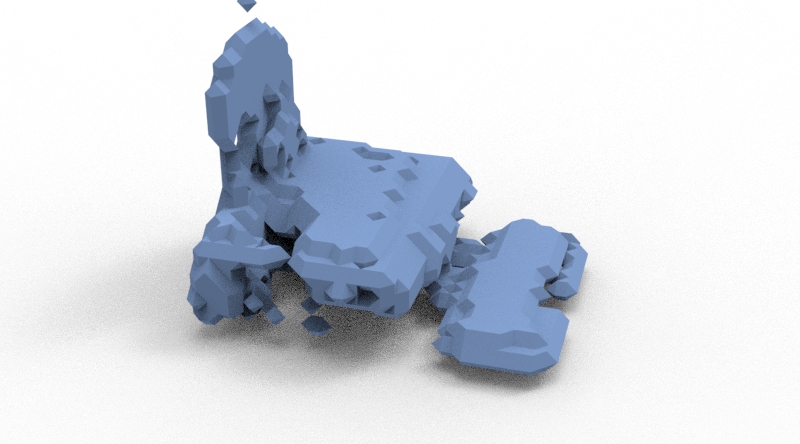}}
	\subfigure{
		\includegraphics[width=0.113\linewidth]{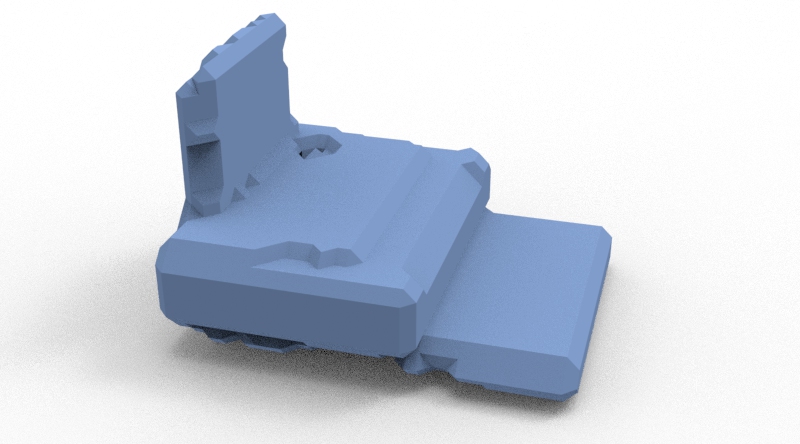}}
	\subfigure{
		\includegraphics[width=0.113\linewidth]{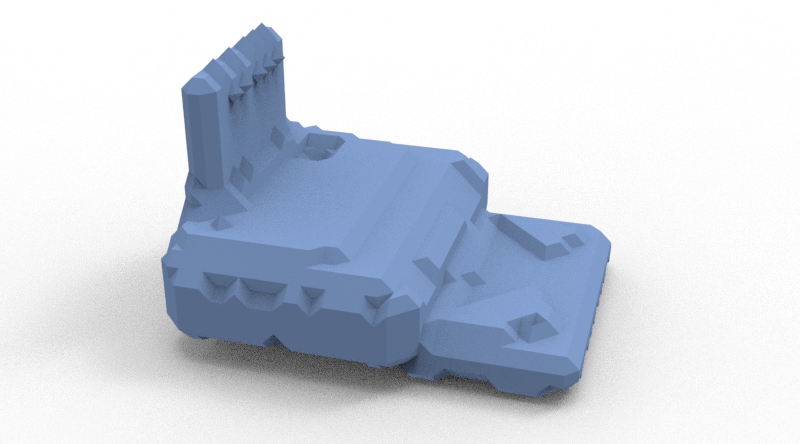}}
	\\
	\subfigure{
		\includegraphics[width=0.113\linewidth]{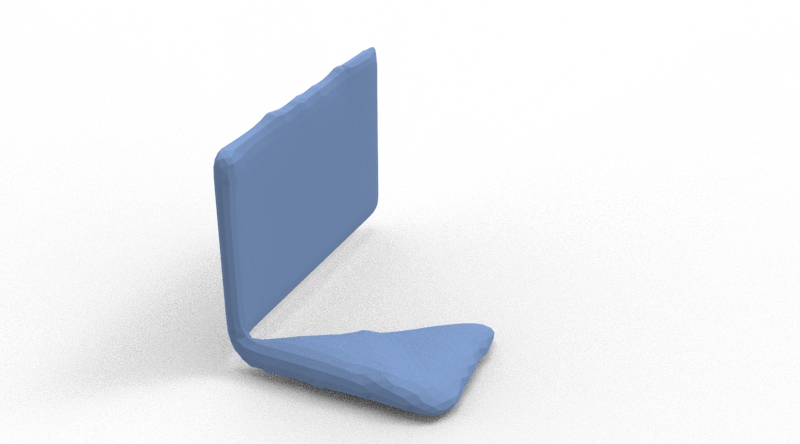}}
	\subfigure{
		\includegraphics[width=0.113\linewidth]{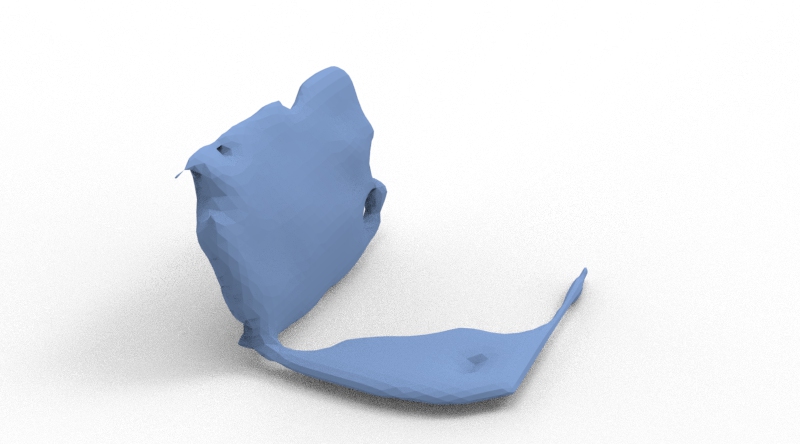}}
	\subfigure{
		\includegraphics[width=0.113\linewidth]{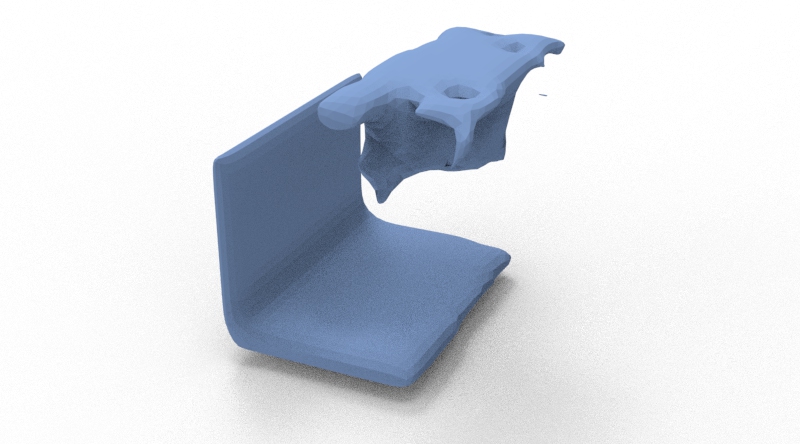}}
	\subfigure{
		\includegraphics[width=0.113\linewidth]{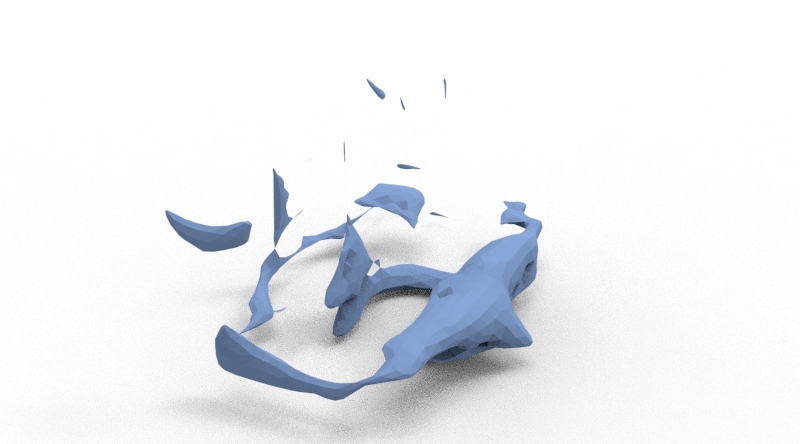}}
        \subfigure{
		\includegraphics[width=0.113\linewidth]{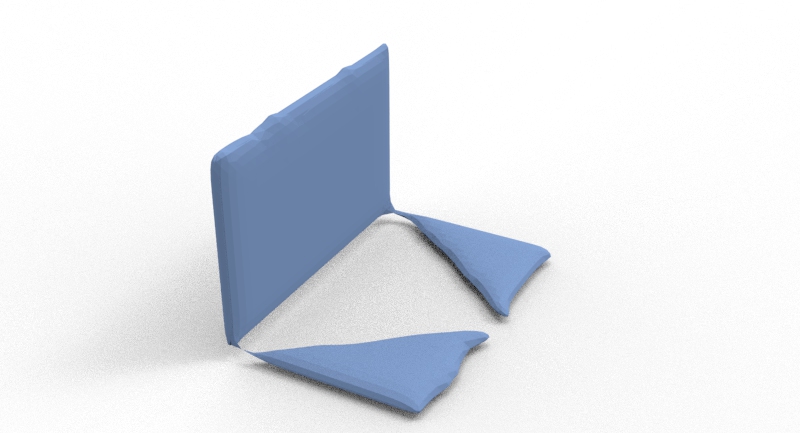}}
	\subfigure{
		\includegraphics[width=0.113\linewidth]{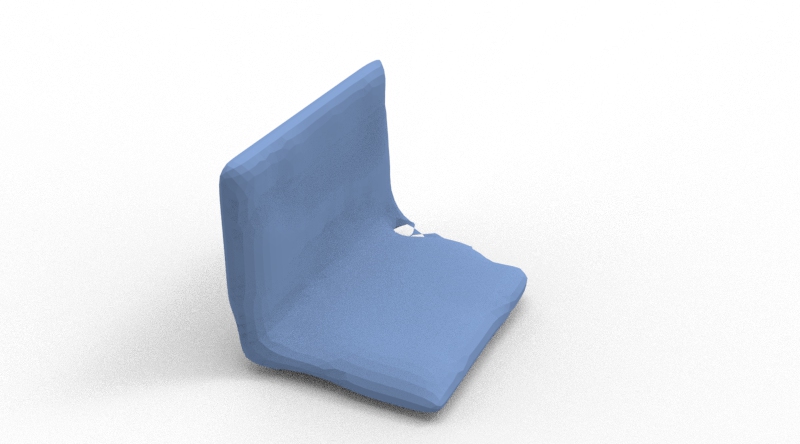}}
	\subfigure{
		\includegraphics[width=0.113\linewidth]{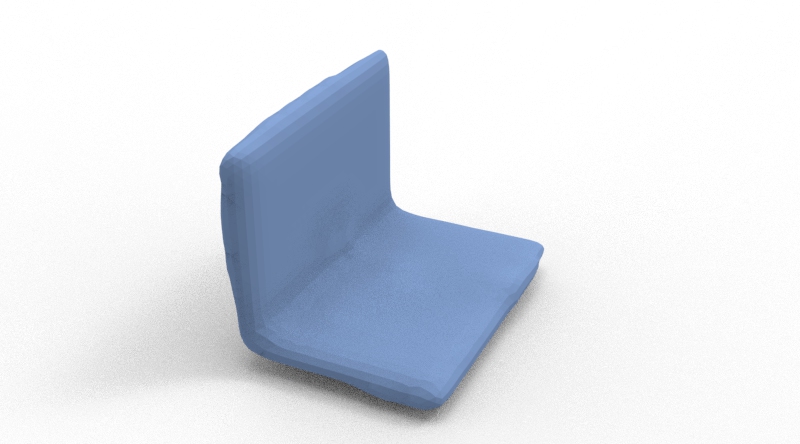}}
	\subfigure{
		\includegraphics[width=0.113\linewidth]{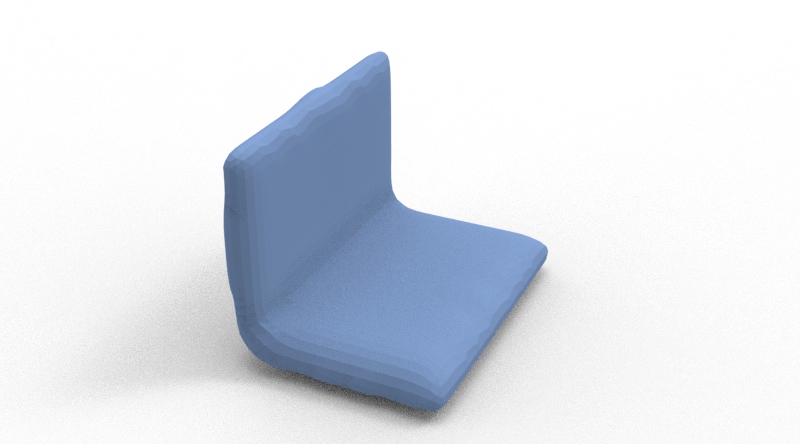}}
	\\
	\setcounter{subfigure}{0}
	%\quad
	\subfigure[]{
		\includegraphics[width=0.113\linewidth]{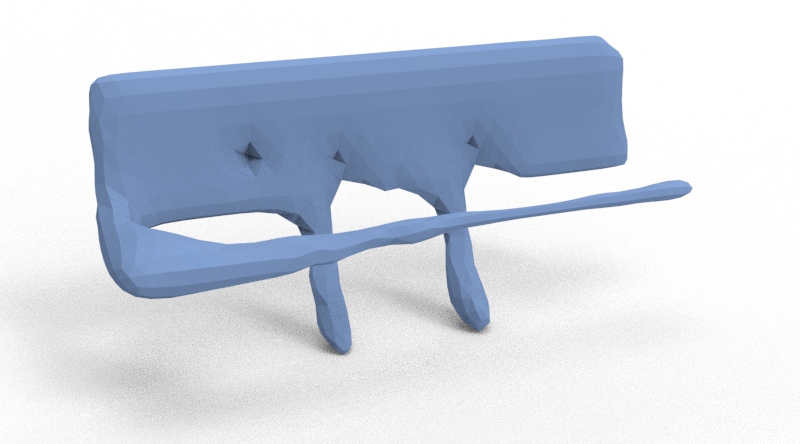}}
	\subfigure[]{
		\includegraphics[width=0.113\linewidth]{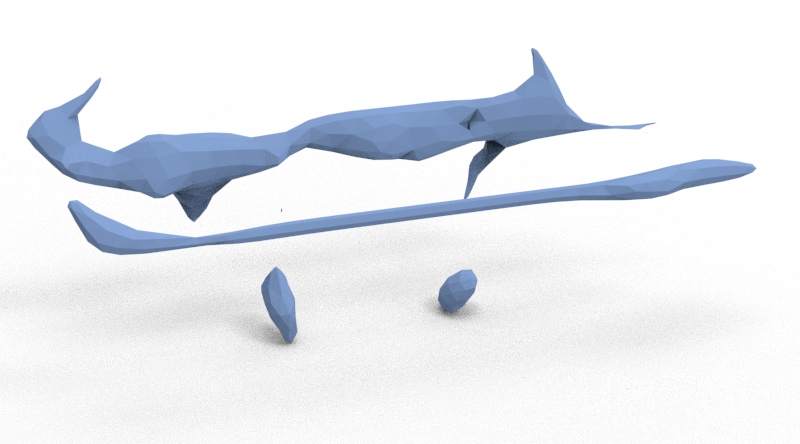}}
	\subfigure[]{
		\includegraphics[width=0.113\linewidth]{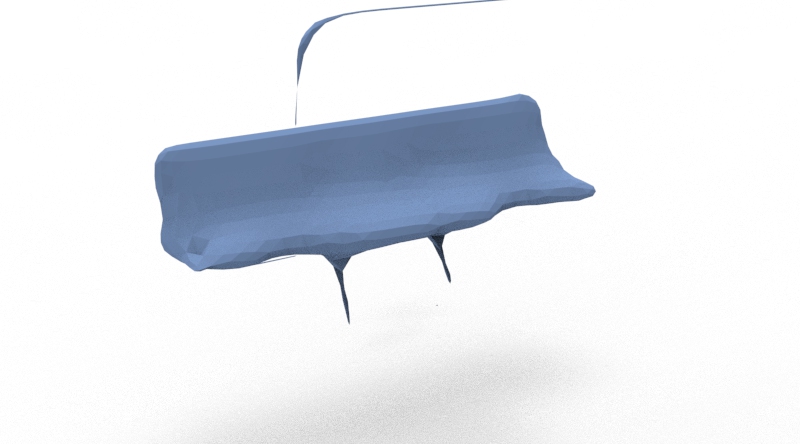}}
	\subfigure[]{
		\includegraphics[width=0.113\linewidth]{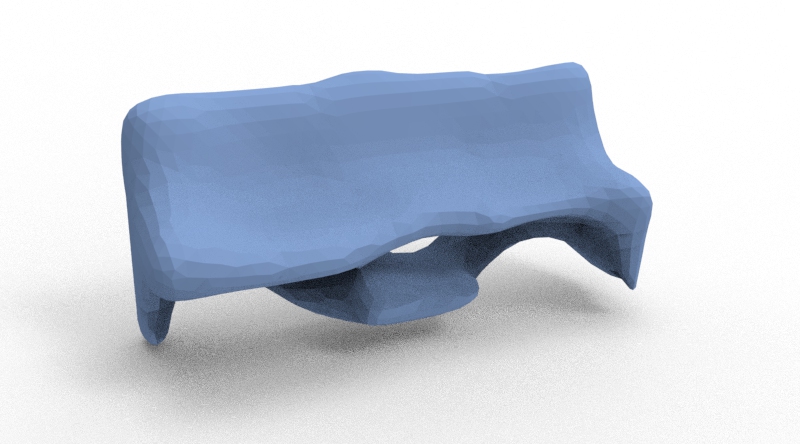}}
        \subfigure[]{
		\includegraphics[width=0.113\linewidth]{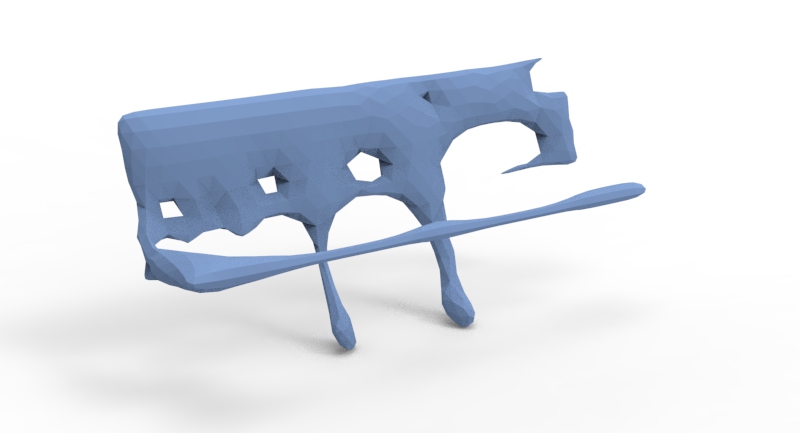}}
	\subfigure[]{
		\includegraphics[width=0.113\linewidth]{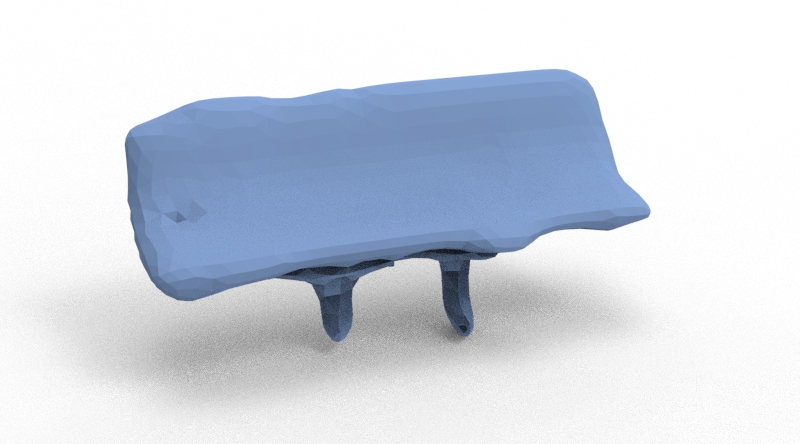}}
	\subfigure[]{
		\includegraphics[width=0.113\linewidth]{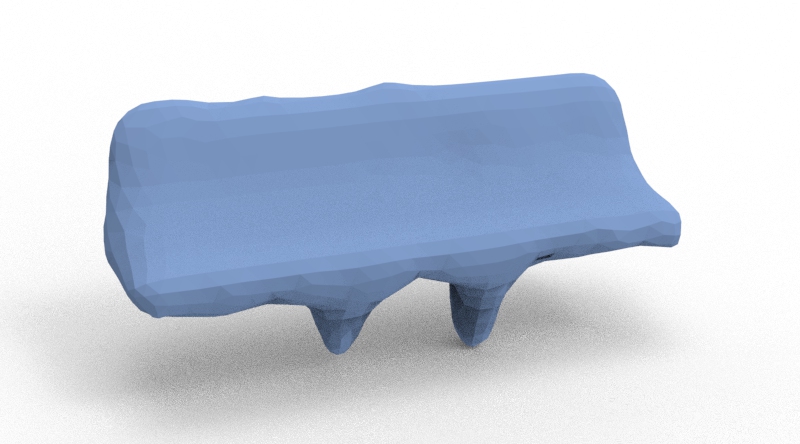}}
	\subfigure[]{
		\includegraphics[width=0.113\linewidth]{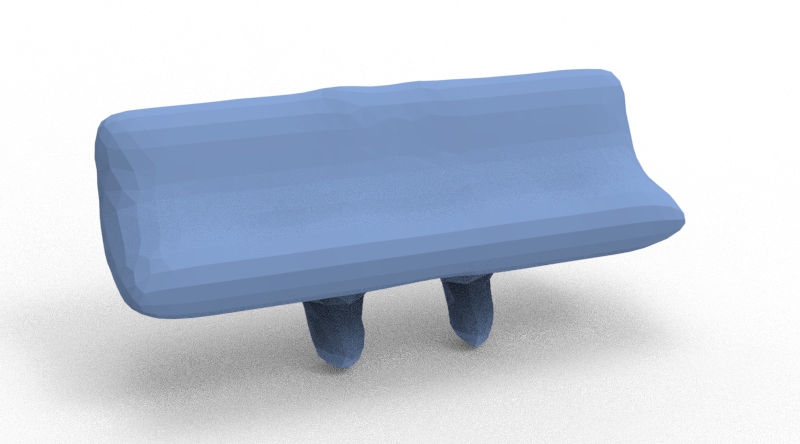}}
	\caption{Comparison of the visual results on the ShapeNet dataset. From top to bottom, the categories of the objects are bag, lamp, bathtub, bed, basket, printer, laptop, and bench. (a) Partial input (b) AutoSDF. (c) IFNet. (d) Fewshot. (e) SDFusion. (f) PatchComplete. (g) Ours. (h) Ground truth.}
	\label{visual_shapenet}
\end{figure*}

\begin{figure*}[htbp]
	\centering  
	\subfigbottomskip=1pt 
	\subfigure{
		\includegraphics[width=0.113\linewidth]{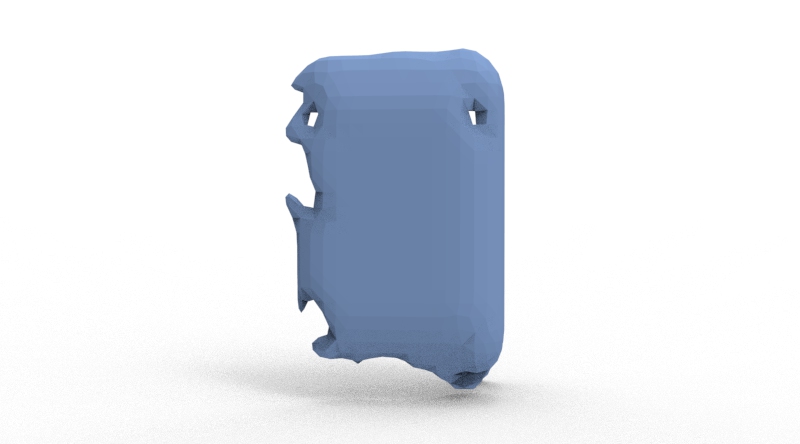}}
	\subfigure{
		\includegraphics[width=0.113\linewidth]{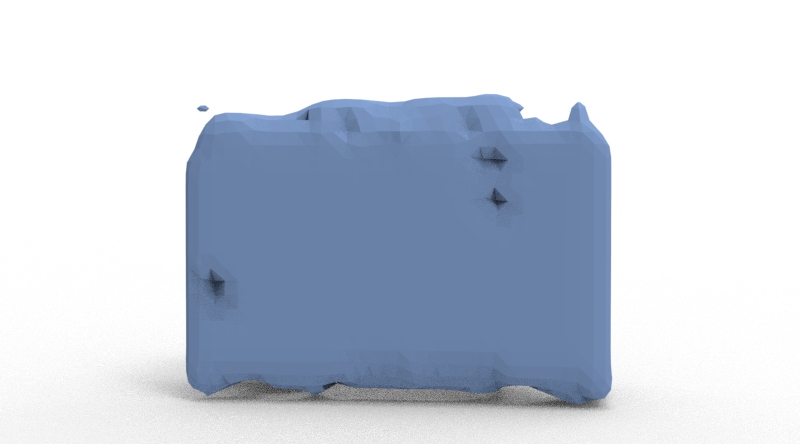}}
	\subfigure{
		\includegraphics[width=0.113\linewidth]{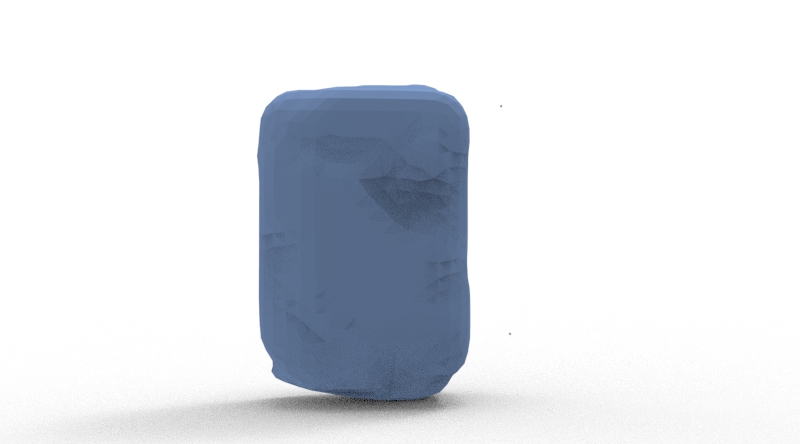}}
	\subfigure{
		\includegraphics[width=0.113\linewidth]{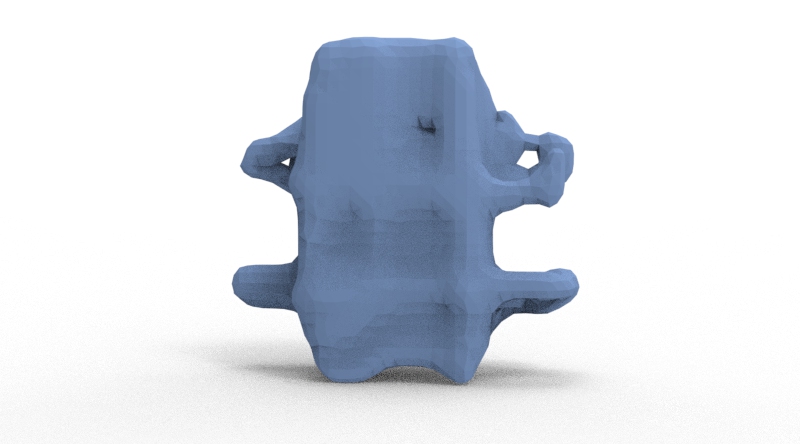}}
        \subfigure{
		\includegraphics[width=0.113\linewidth]{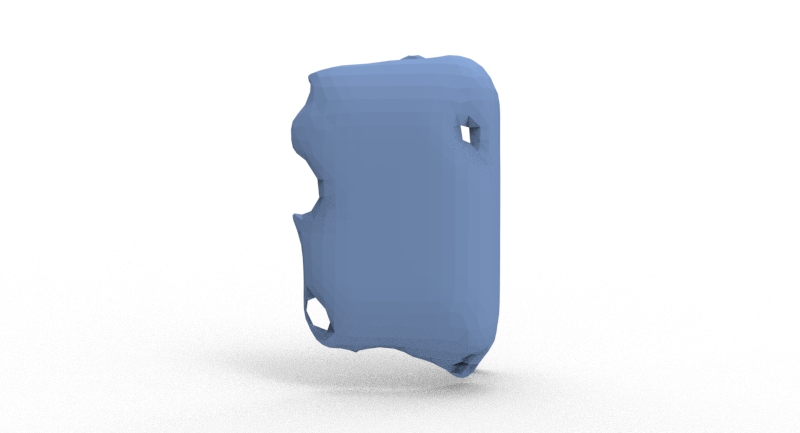}}
	\subfigure{
		\includegraphics[width=0.113\linewidth]{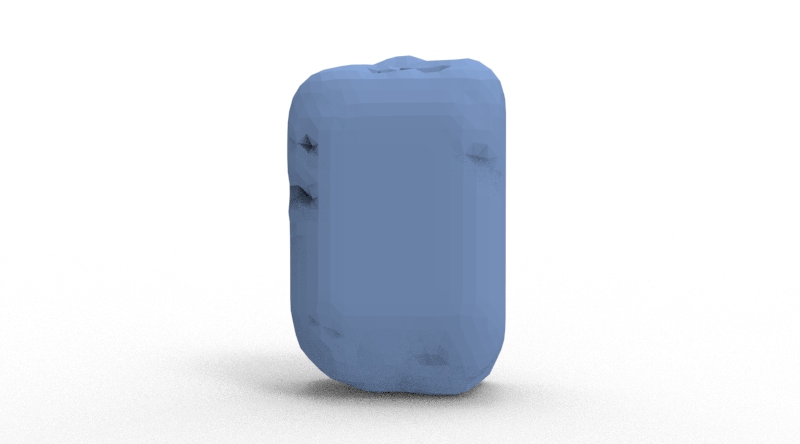}}
	\subfigure{
		\includegraphics[width=0.113\linewidth]{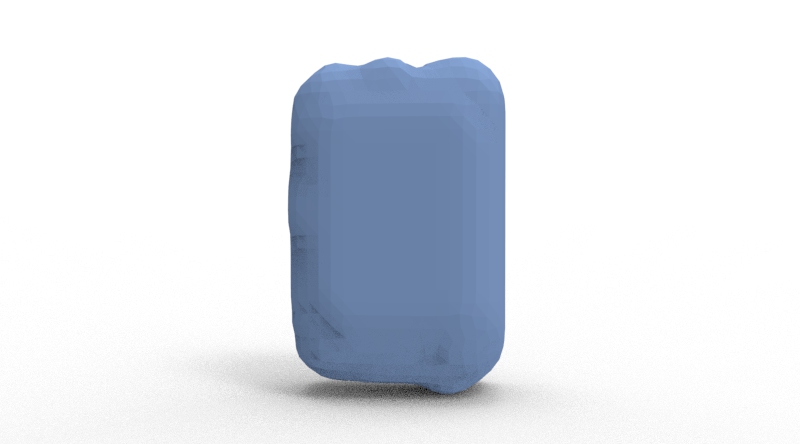}}
	\subfigure{
		\includegraphics[width=0.113\linewidth]{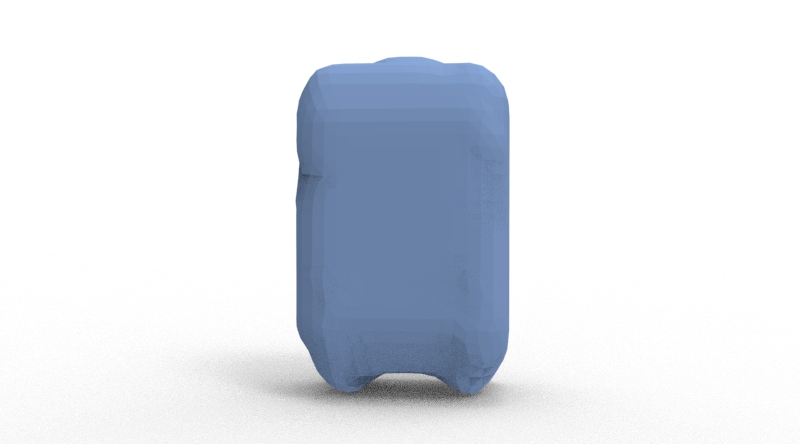}}
	\\
	\subfigure{
		\includegraphics[width=0.113\linewidth]{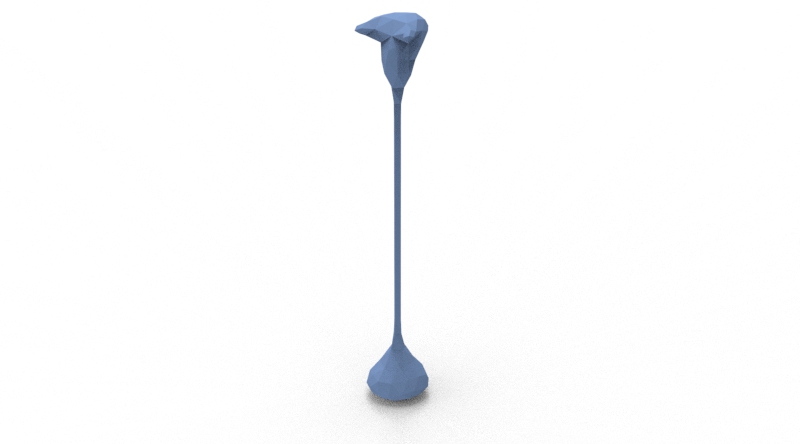}}
	\subfigure{
		\includegraphics[width=0.113\linewidth]{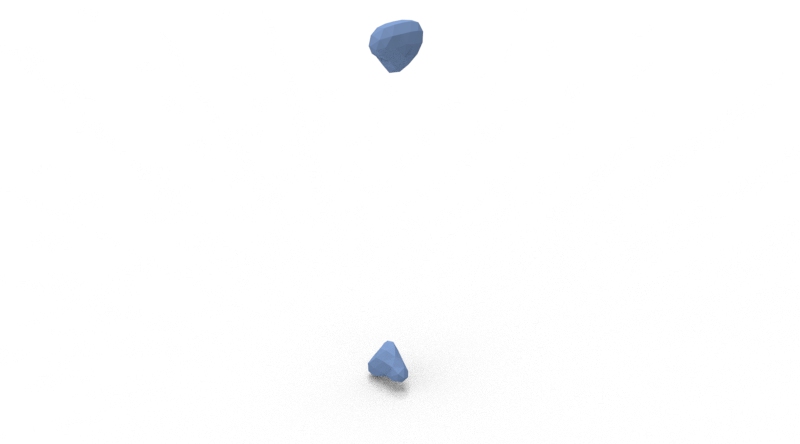}}
	\subfigure{
		\includegraphics[width=0.113\linewidth]{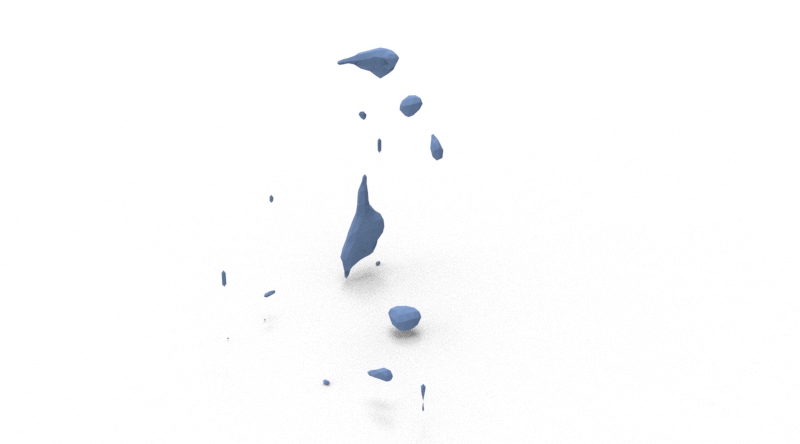}}
	\subfigure{
		\includegraphics[width=0.113\linewidth]{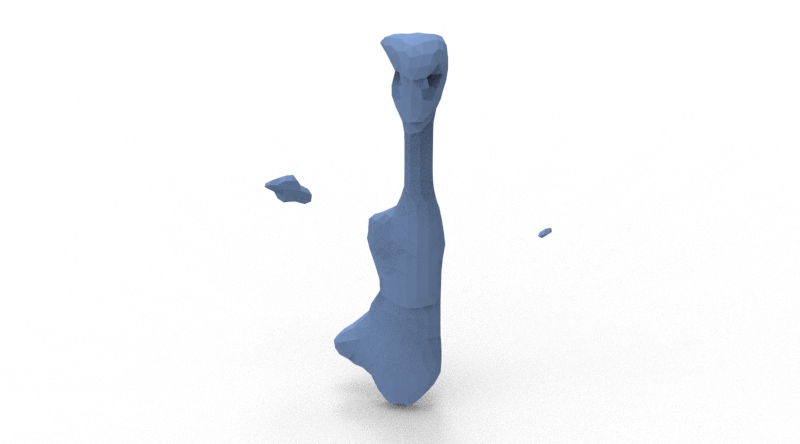}}
        \subfigure{
		\includegraphics[width=0.113\linewidth]{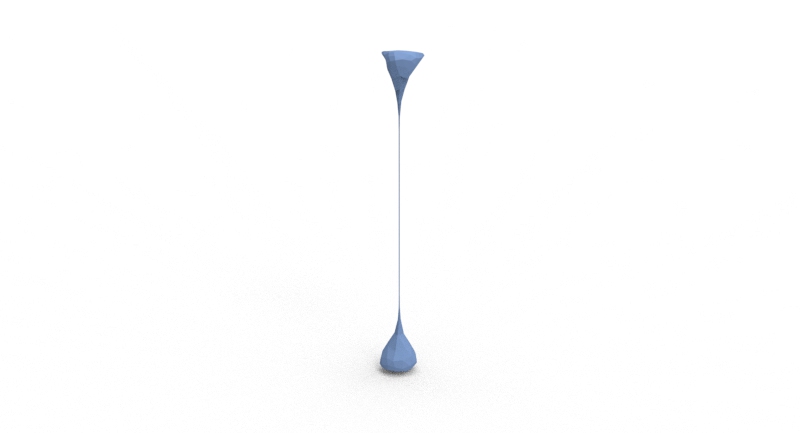}}
	\subfigure{
		\includegraphics[width=0.113\linewidth]{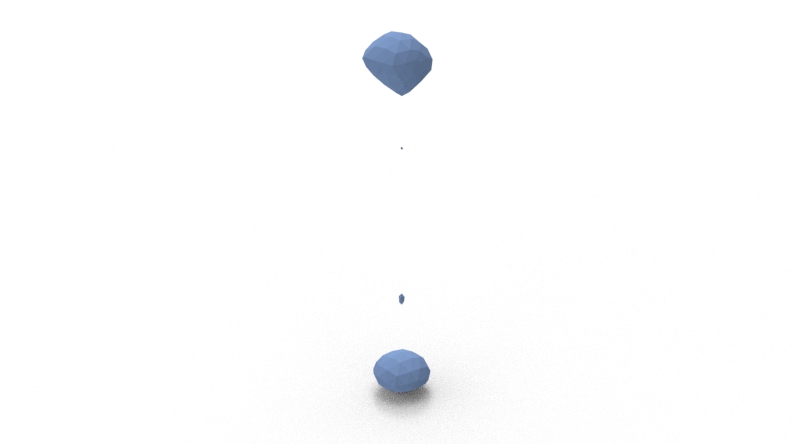}}
	\subfigure{
		\includegraphics[width=0.113\linewidth]{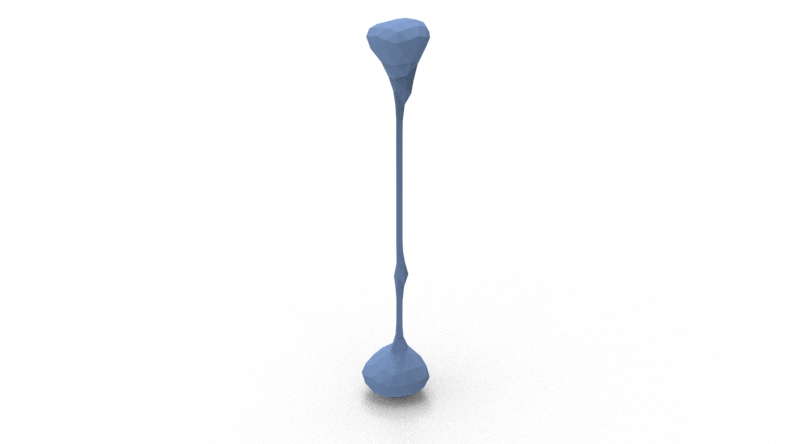}}
	\subfigure{
		\includegraphics[width=0.113\linewidth]{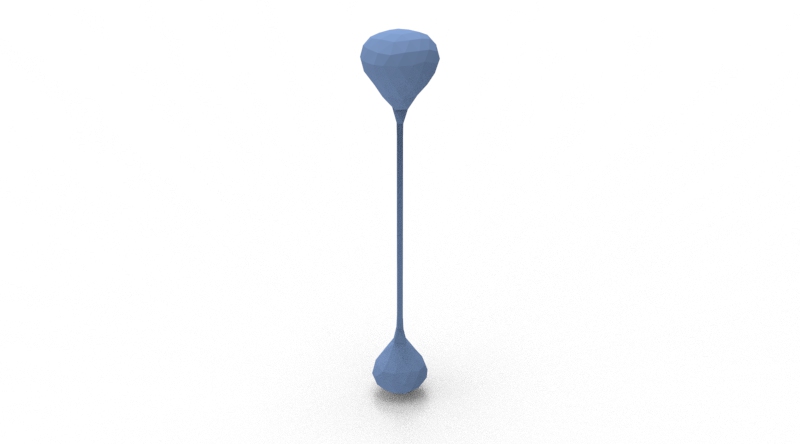}}
	\\
	\subfigure{
		\includegraphics[width=0.113\linewidth]{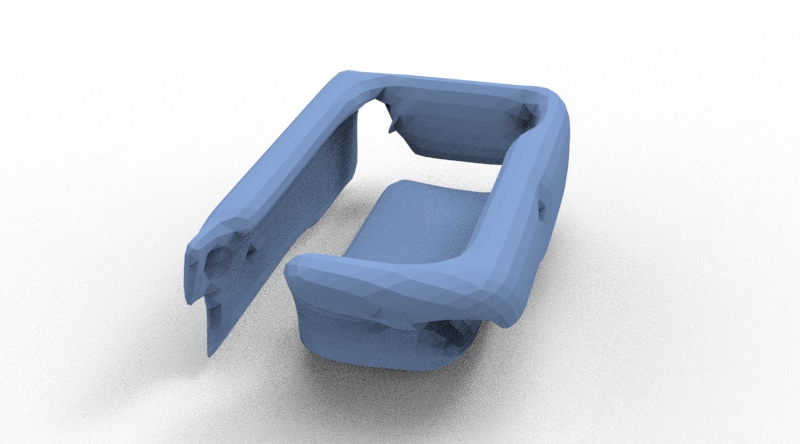}}
	\subfigure{
		\includegraphics[width=0.113\linewidth]{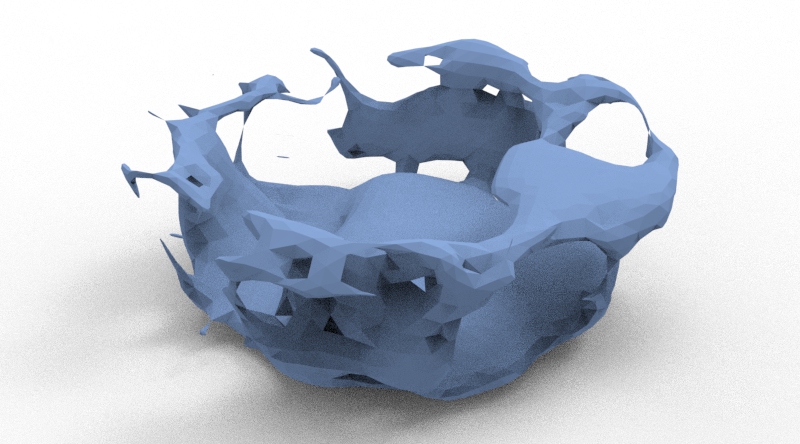}}
	\subfigure{
		\includegraphics[width=0.113\linewidth]{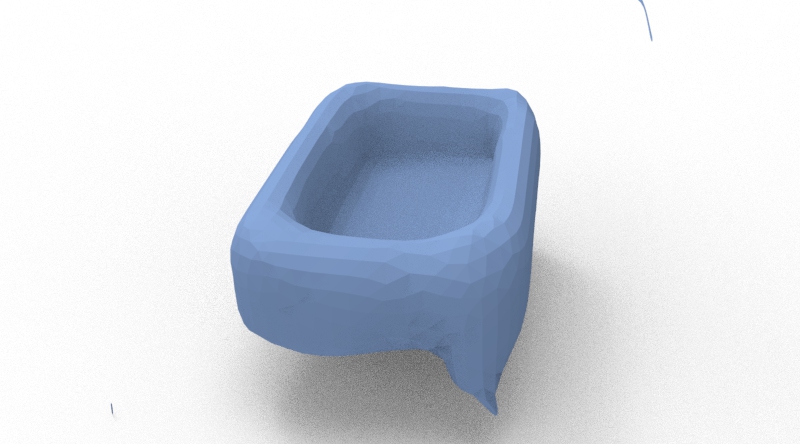}}
	\subfigure{
		\includegraphics[width=0.113\linewidth]{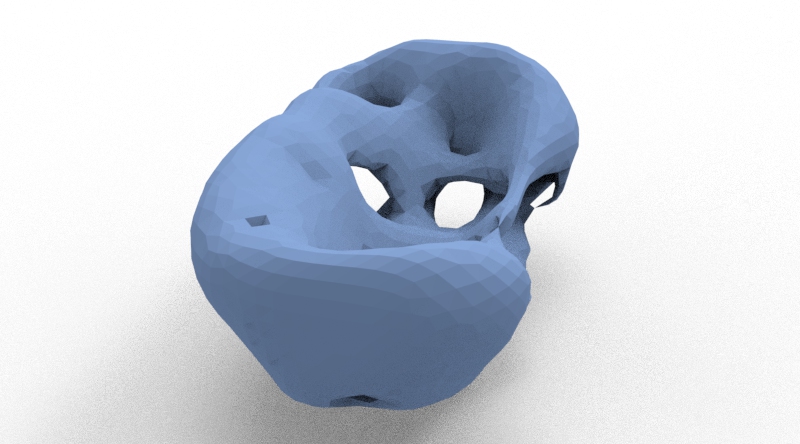}}
        \subfigure{
		\includegraphics[width=0.113\linewidth]{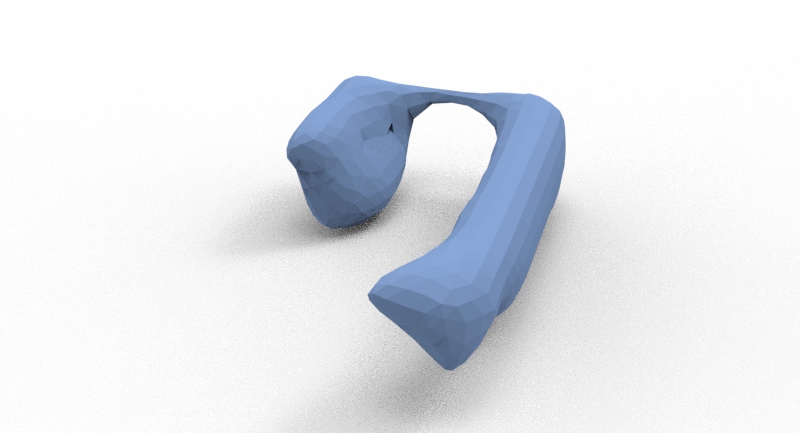}}
	\subfigure{
		\includegraphics[width=0.113\linewidth]{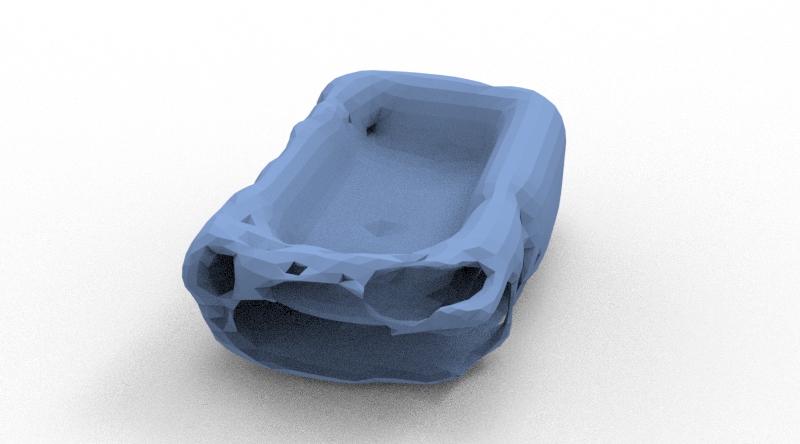}}
	\subfigure{
		\includegraphics[width=0.113\linewidth]{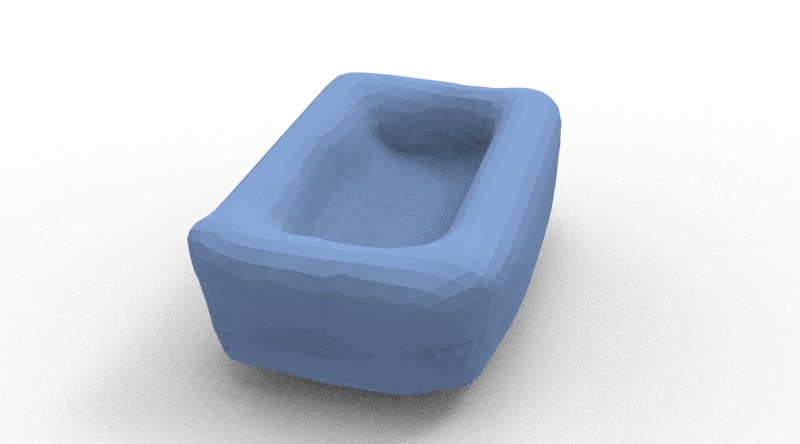}}
	\subfigure{
		\includegraphics[width=0.113\linewidth]{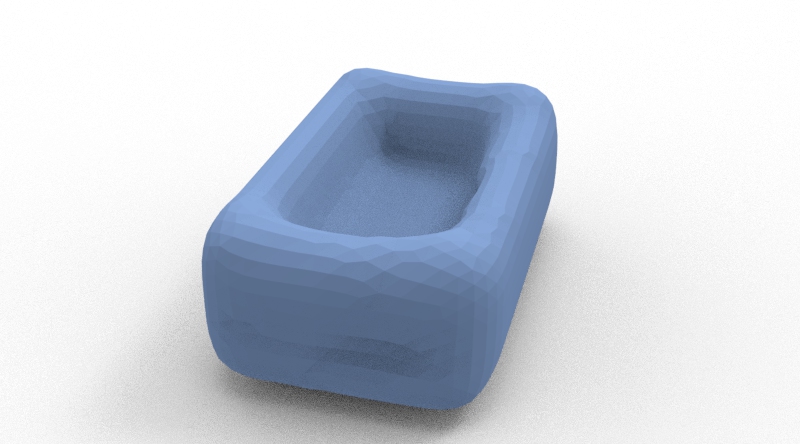}}
	\\
	\subfigure{
		\includegraphics[width=0.113\linewidth]{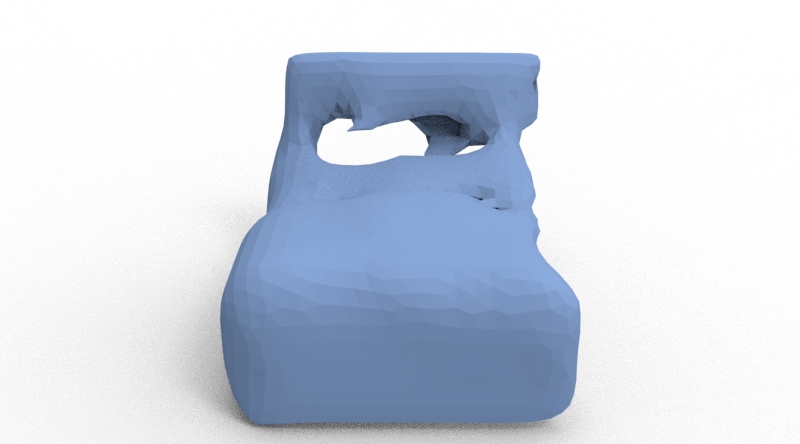}}
	\subfigure{
		\includegraphics[width=0.113\linewidth]{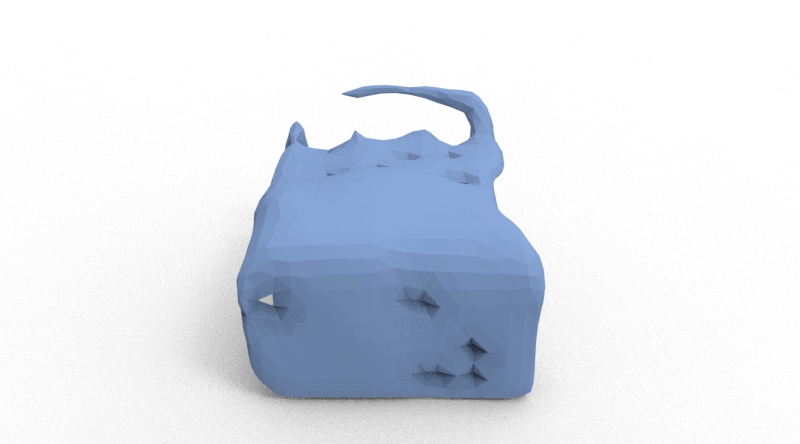}}
	\subfigure{
		\includegraphics[width=0.113\linewidth]{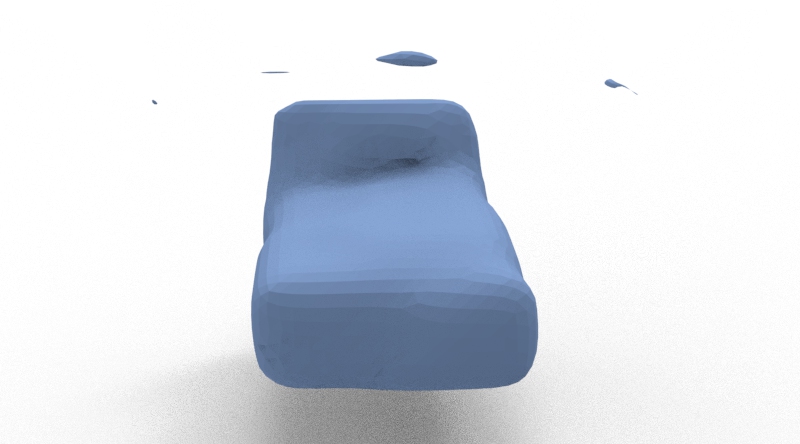}}
	\subfigure{
		\includegraphics[width=0.113\linewidth]{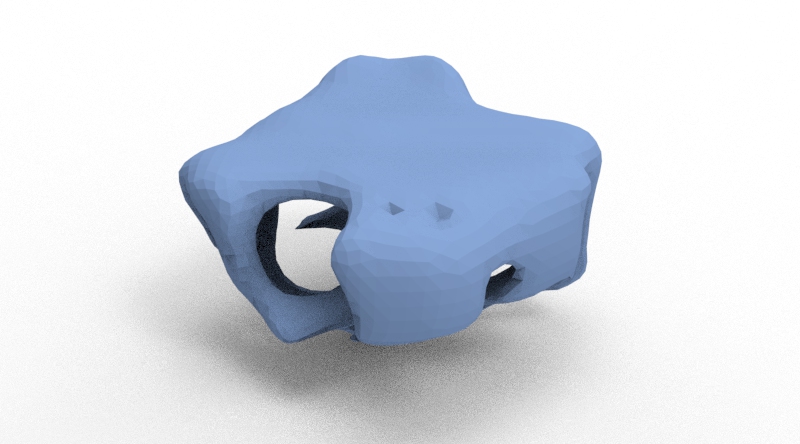}}
        \subfigure{
		\includegraphics[width=0.113\linewidth]{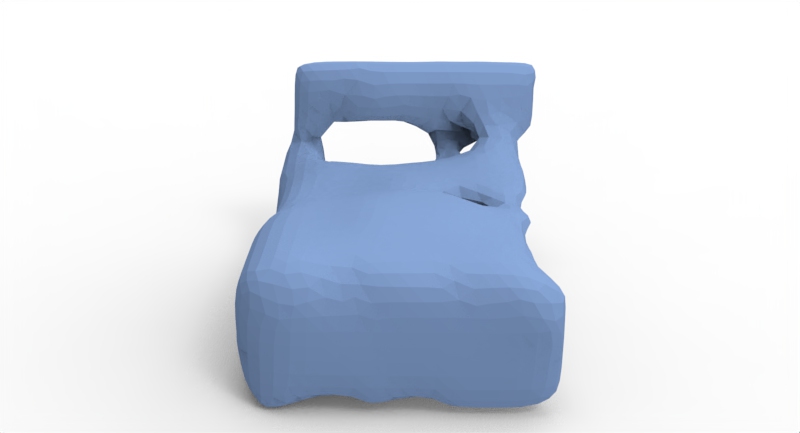}}
	\subfigure{
		\includegraphics[width=0.113\linewidth]{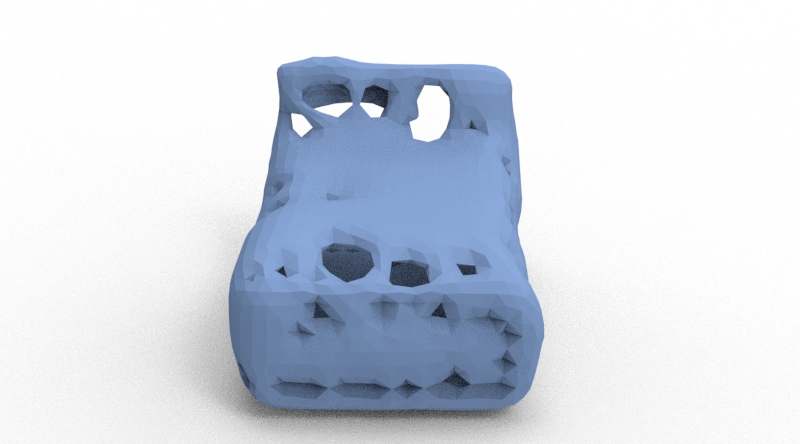}}
	\subfigure{
		\includegraphics[width=0.113\linewidth]{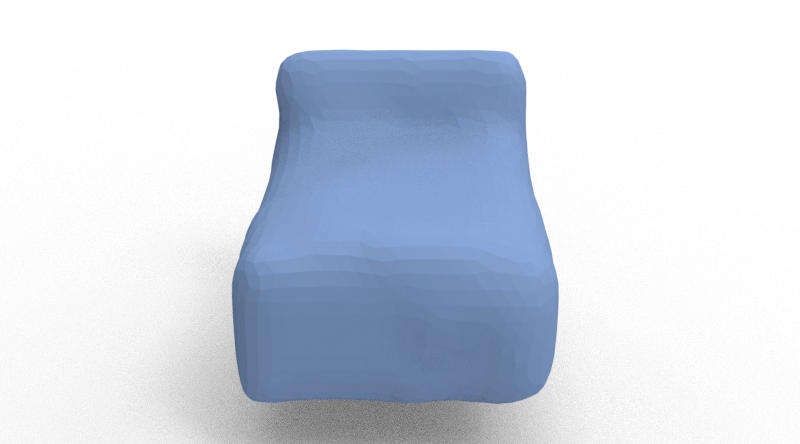}}
	\subfigure{
		\includegraphics[width=0.113\linewidth]{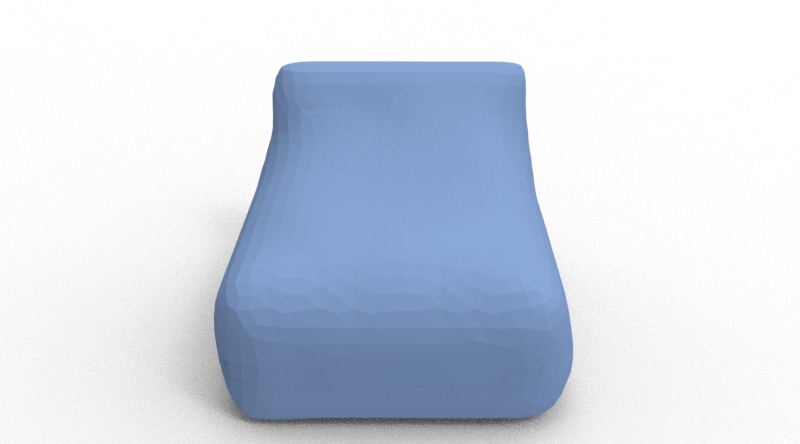}}
	\\
	\subfigure{
		\includegraphics[width=0.113\linewidth]{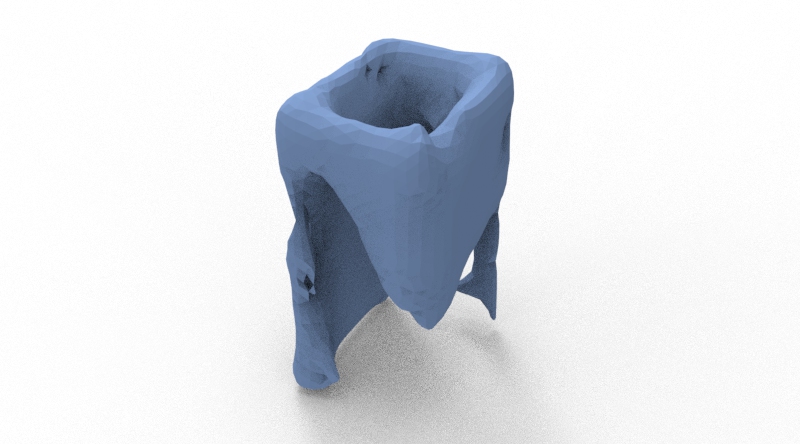}}
	\subfigure{
		\includegraphics[width=0.113\linewidth]{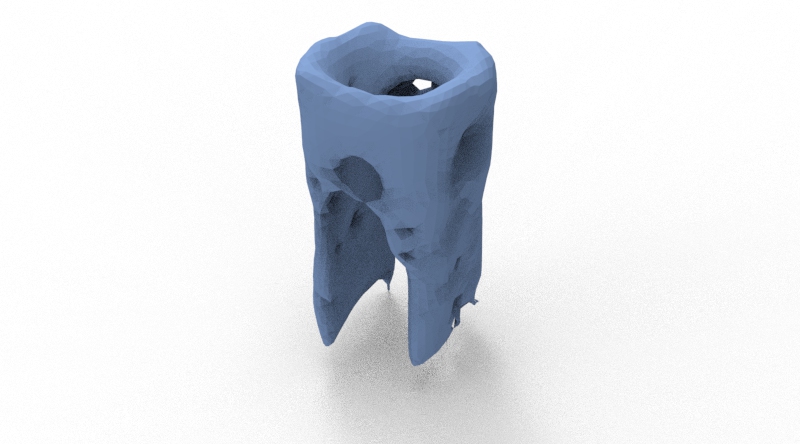}}
	\subfigure{
		\includegraphics[width=0.113\linewidth]{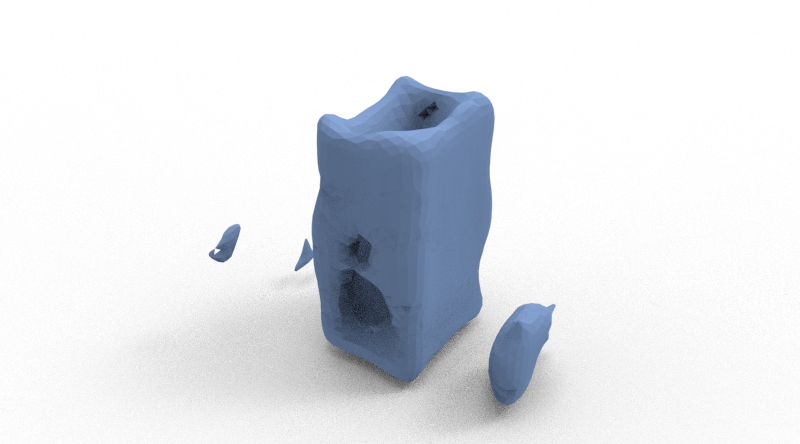}}
	\subfigure{
		\includegraphics[width=0.113\linewidth]{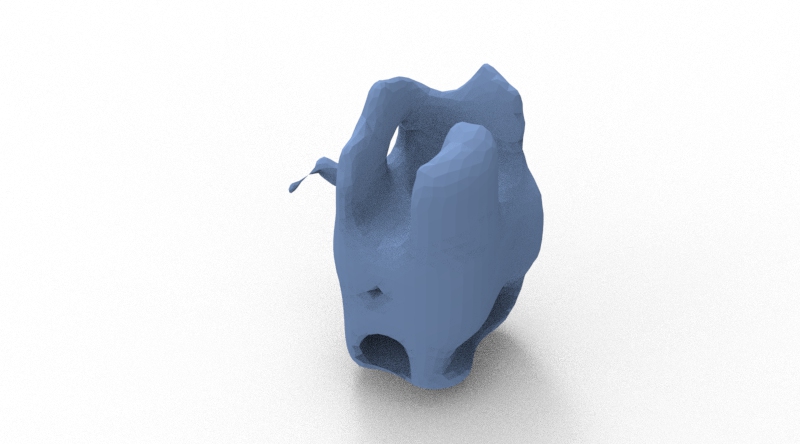}}
  \subfigure{
		\includegraphics[width=0.113\linewidth]{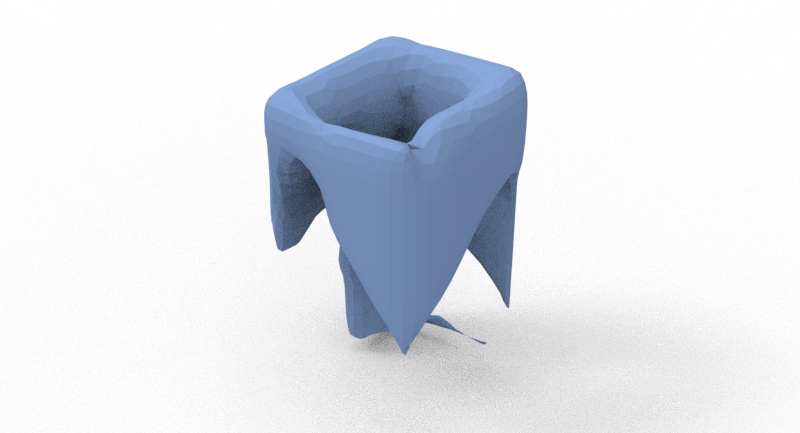}}
	\subfigure{
		\includegraphics[width=0.113\linewidth]{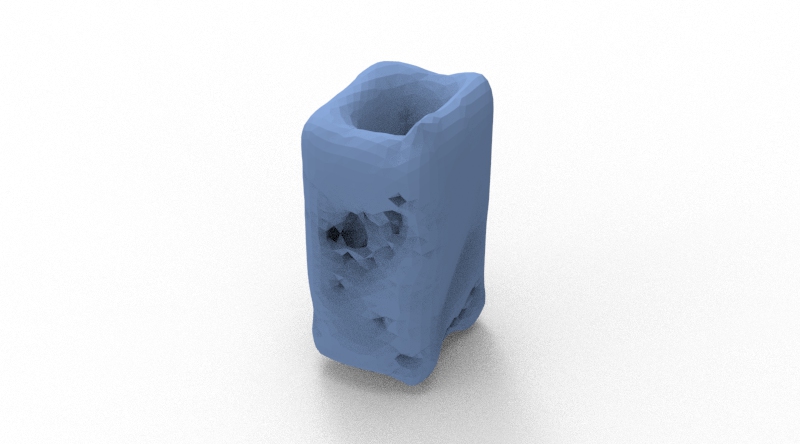}}
	\subfigure{
		\includegraphics[width=0.113\linewidth]{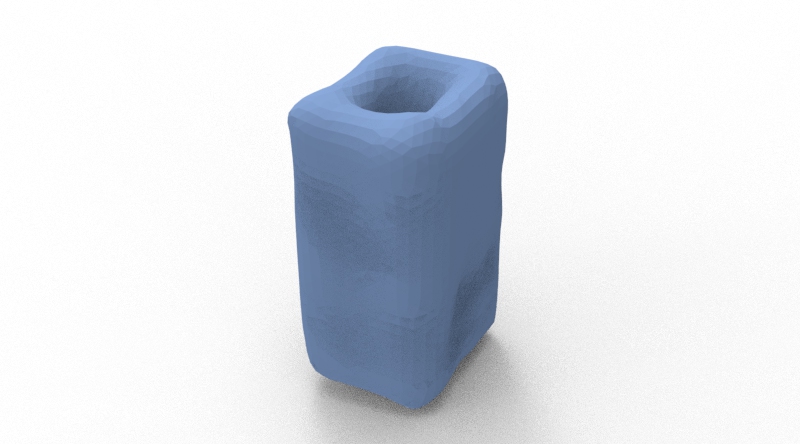}}
	\subfigure{
		\includegraphics[width=0.113\linewidth]{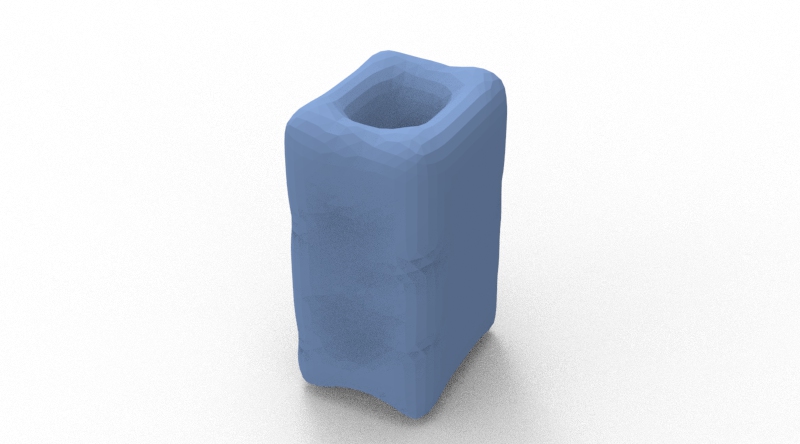}}
	\\
        \setcounter{subfigure}{0}
	%\quad
	\subfigure[]{
		\includegraphics[width=0.113\linewidth]{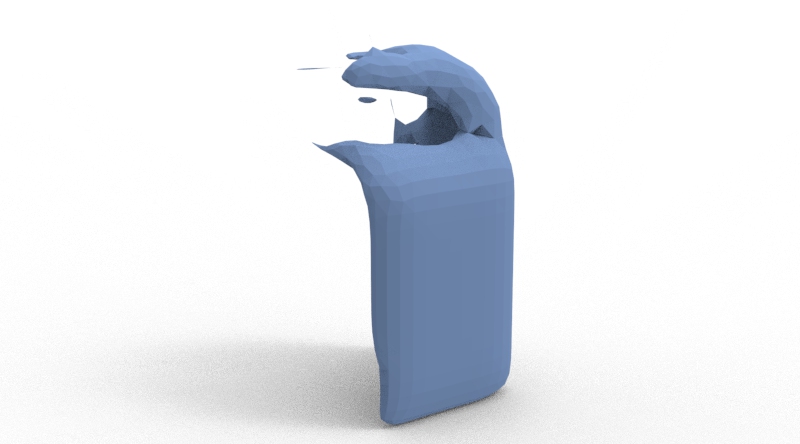}}
	\subfigure[]{
		\includegraphics[width=0.113\linewidth]{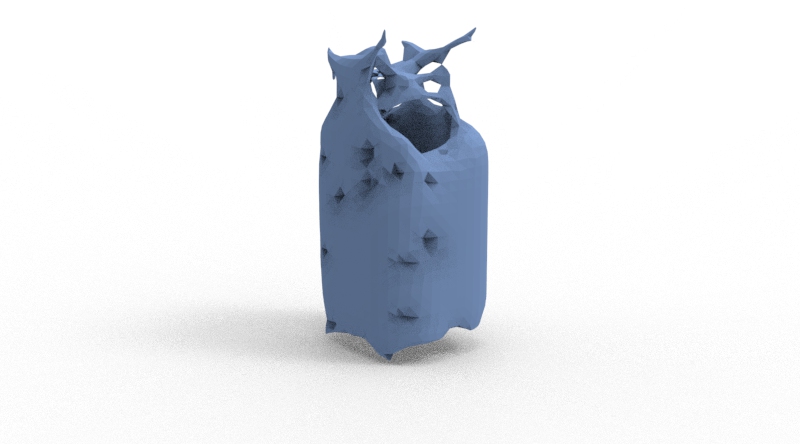}}
	\subfigure[]{
		\includegraphics[width=0.113\linewidth]{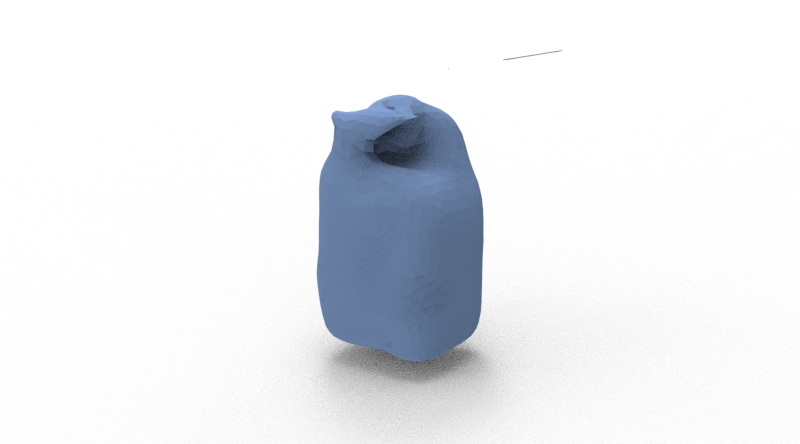}}
	\subfigure[]{
		\includegraphics[width=0.113\linewidth]{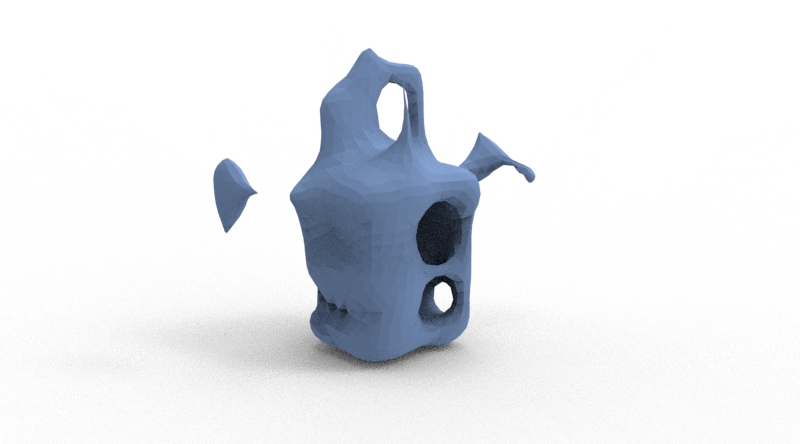}}
        \subfigure[]{
		\includegraphics[width=0.113\linewidth]{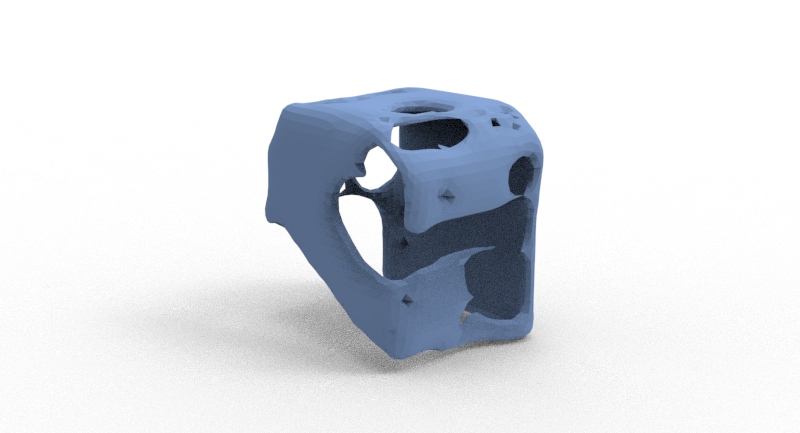}}
	\subfigure[]{
		\includegraphics[width=0.113\linewidth]{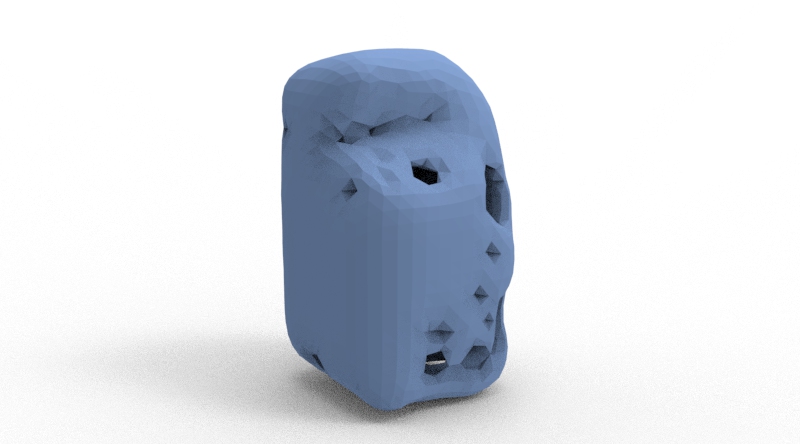}}
	\subfigure[]{
		\includegraphics[width=0.113\linewidth]{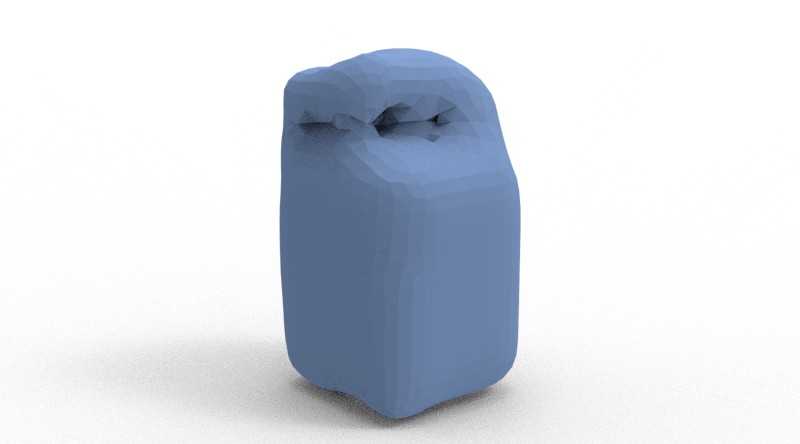}}
	\subfigure[]{
		\includegraphics[width=0.113\linewidth]{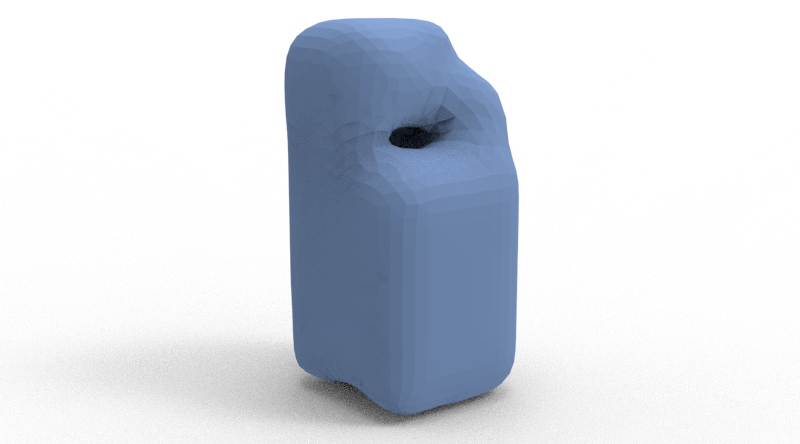}}
	\caption{Comparison of the visual results on ScanNet dataset. From top to bottom, the categories of the objects are bag, lamp, bathtub, bed, basket, and printer. (a) Partial input (b) AutoSDF. (c) IFNet. (d) Fewshot. (e) SDFusion. (f) PatchComplete. (g) Ours. (h) Ground truth.}
	\label{visual_scannet}
\end{figure*}

\subsubsection{Results on synthetic data}
We first conducted experiments on ShapeNet. Table \ref{shapenet_results} shows the quantization results of our method and other state-of-the-art (SOTA) shape completion and reconstruction methods, including Few-Shot \cite{wallace2019few}, IF-Nets \cite{chibane2020implicit}, AutoSDF \cite{mittal2022autosdf}, SDFusion \cite{cheng2023sdfusion} and PatchComplete \cite{raopatchcomplete}. \revise{Both our method and PatchComplete take as input a partial shape along with a set of priors, in contrast to other approaches that only use a single partial shape. However, the priors are selected from the training set without requiring additional data.} We can see that our method achieves the best results among all categories in terms of IoU and F1 score. Specifically, the average IoU and F1 of our model are almost double those of Few-Shot, IF-Nets, and AutoSDF. Compared to PatchComplete, our model improves the average IoU and F1 score by 0.045 and 0.043, respectively. Especially the IoU of our model surpasses that of PatchComplete by more than 0.05 in four categories, i.e., bag, bed, printer, and laptop. In terms of the CD metric, our method achieves the best average result and exceeds the second-place method by 0.33. Besides, our method outperforms all other methods in six categories and only slightly falls behind PatchComplete in the remaining two categories.

Figure \ref{visual_shapenet} displays the qualitative results of our method and the baselines on the ShapeNet dataset. Note that the objects depicted in these images are meshes converted from predicted voxels and have undergone smoothing. As shown in Figure \ref{visual_shapenet}, our method is able to recover the missing part effectively while preserving the partial input. AutoSDF and Few-Shot tend to modify the partial and face challenges in accurately predicting missing parts. Although SDFusion better preserves the partial input than AutoSDF and Few-Shot, it struggles to recover missing regions. IFNet is susceptible to generating noises during reconstruction that do not belong to the objects themselves. The shape reconstructed by PatchComplete is incomplete, with some holes resulting from missing details. However, our method can reconstruct a more complete shape and restore finer details, such as the corners of the printer and the basket.

\begin{table*}[htbp]
	\centering
    \renewcommand\arraystretch{1.25}
	\caption{Results comparison with the point cloud completion methods on ShapeNet in terms of CD\(\downarrow\) (scaled by \(10^{2}\)).  The best results are highlighted in bold.}
	\label{shapenet_completion}
	\begin{tabular}{l|c|c|c|c|c|c|c|c|c}
		\toprule[1.2pt]
		Methods & Average & Bag & Lamp & Bathtub & Bed & Basket	& Printer	& Laptop	& Bench \\ \hline  
	PCN \cite{yuan2018pcn}  & 7.57   &7.27 & 11.08&6.01     &8.43 & 7.34   &8.28  &6.71 & 5.42 \\
	GRNet \cite{xie2020grnet}&4.96    &4.65 & 5.74 &4.44     &5.59 &5.13    &5.84  & 4.62 &  3.63    \\ 
	AnchorFormer\cite{chen2023anchorformer} & 4.64 & 4.64 & 6.33 & 3.70 & 5.50 & 4.69 & 5.48 & 3.90 &  2.87  \\ \hline
	CS-SC (ours)        & \textbf{3.02} & \textbf{2.75} & \textbf{5.01} & \textbf{2.18} & \textbf{3.40} & \textbf{2.53} & \textbf{3.54} & \textbf{2.18}	& \textbf{2.57}	 \\ \bottomrule[1.2pt]
	\end{tabular}
\end{table*}

\begin{table*}[htbp]
	\centering
	\renewcommand\arraystretch{1.25}
	\caption{Results comparison with the point cloud completion methods on ScanNet in terms of CD\(\downarrow\) (scaled by \(10^{2}\)).  The best results are highlighted in bold.}
	\label{scannet_completion}
	\begin{tabular}{l|c|c|c|c|c|c|c}
		\toprule[1.2pt]
		Methods & Average & Bag & Lamp & Bathtub & Bed & Basket	& Printer\\ \hline  
	PCN \cite{yuan2018pcn}  & 6.38 &7.27 & 6.66  &5.75  & 5.73  & 5.93  & 6.96 \\
	GRNet \cite{xie2020grnet} & 4.47 & 4.58 & 3.86 & 4.27 & 4.17  & 4.54 & 5.37  \\ 
	AnchorFormer \cite{chen2023anchorformer} &4.21 &4.88 & \textbf{3.87} &3.54 &3.90 &3.85 &5.26\\ \hline
	CS-SC (ours)        & \textbf{3.81} & \textbf{4.41} & 3.93 & \textbf{2.99} & \textbf{3.54} & \textbf{3.11} & \textbf{4.92}	 \\ \bottomrule[1.2pt]
	\end{tabular}
\end{table*}

\subsubsection{Results on real-world data}
Our method is effective not only on synthetic datasets but also demonstrates strong performance on real-world datasets. Table \ref{scannet_results} shows the quantization results of our method and other SOTA shape completion and reconstruction methods. Overall, our method outperforms all baselines by large margins in terms of average IoU, F1 score, and CD, especially with a nearly 16\% lead over the second-ranked method in terms of IoU metric. In addition, our method yields a comprehensive lead over other methods in all categories in terms of CD, while the IoU and F1 score are only slightly behind those of PatchComplete in one category. It is worth noting that our method's IoU and F1 score in the bathtub and bed categories both demonstrate a lead of more than 0.1 over those of the second-ranked method.

Figure \ref{visual_scannet} depicts the visual results of our method and the baselines on the ScanNet dataset. The visual results indicate that all methods perform similarly on the ScanNet dataset as they did on the ShapeNet dataset. Our method is capable of predicting missing parts that are more reasonable and finely detailed while more effectively preserving the input parts compared to other methods.

\subsubsection{Comparison with point cloud completion methods}
\begin{figure}[htbp]
	\centering  
	\subfigbottomskip=1pt 
	\subfigure{
		\includegraphics[width=0.19\linewidth]{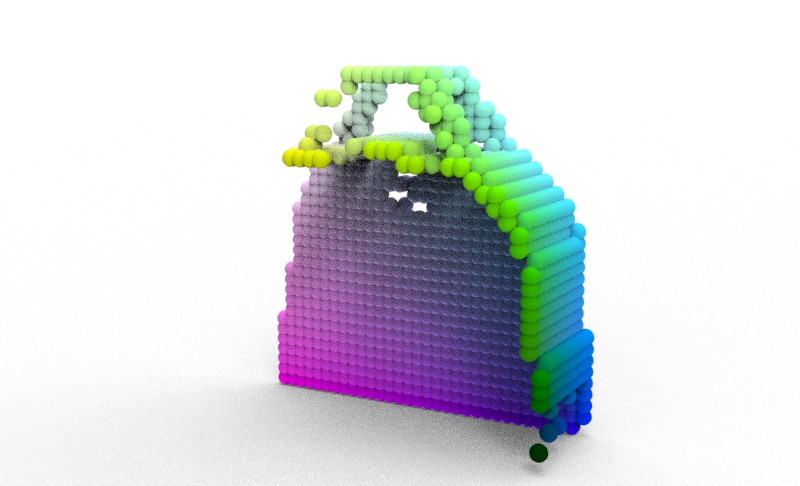}}
	\hspace{-3mm}
	\subfigure{
		\includegraphics[width=0.19\linewidth]{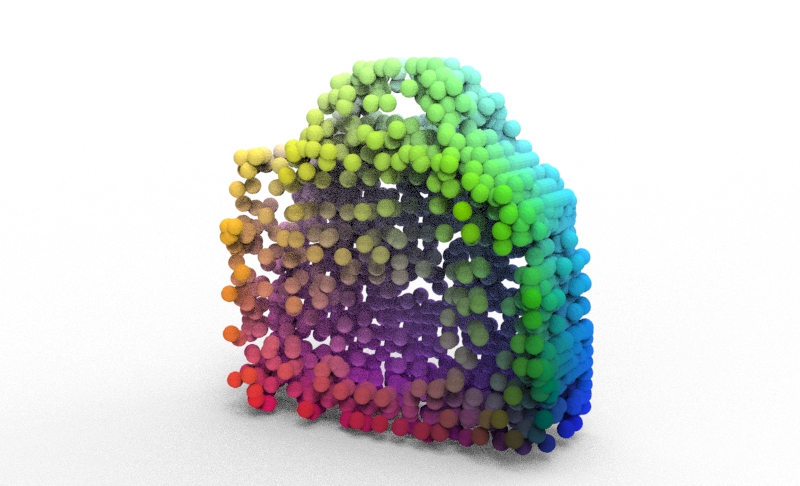}}
	\hspace{-3mm}
	\subfigure{
		\includegraphics[width=0.19\linewidth]{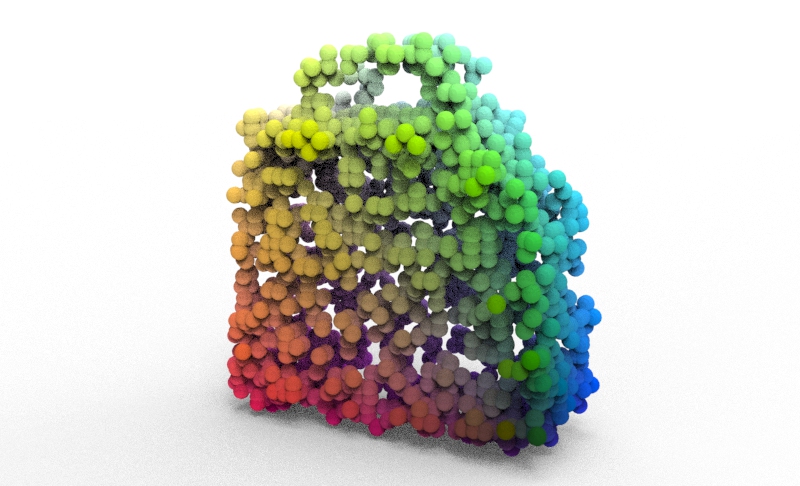}}
	\hspace{-3mm}
	\subfigure{
		\includegraphics[width=0.19\linewidth]{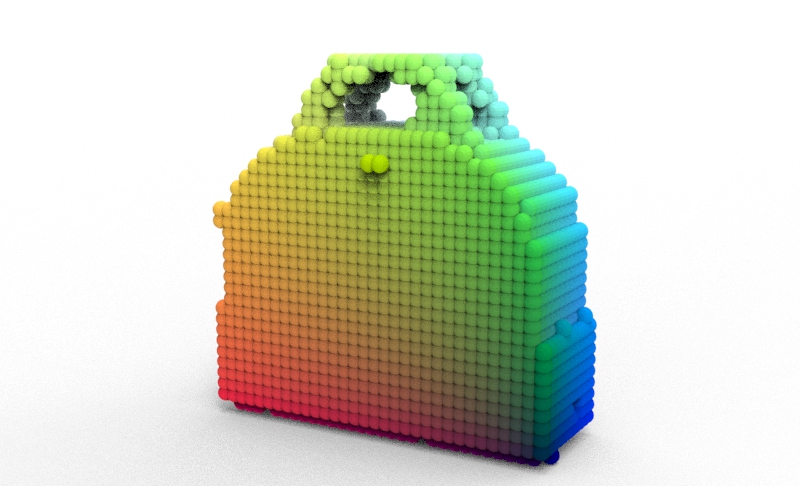}}
        \hspace{-3mm}
	\subfigure{
		\includegraphics[width=0.19\linewidth]{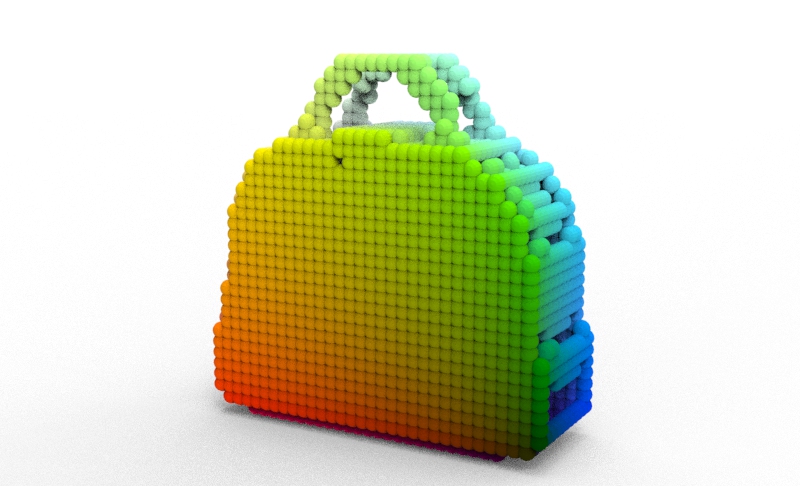}}
	\\
        \subfigure{
		\includegraphics[width=0.19\linewidth]{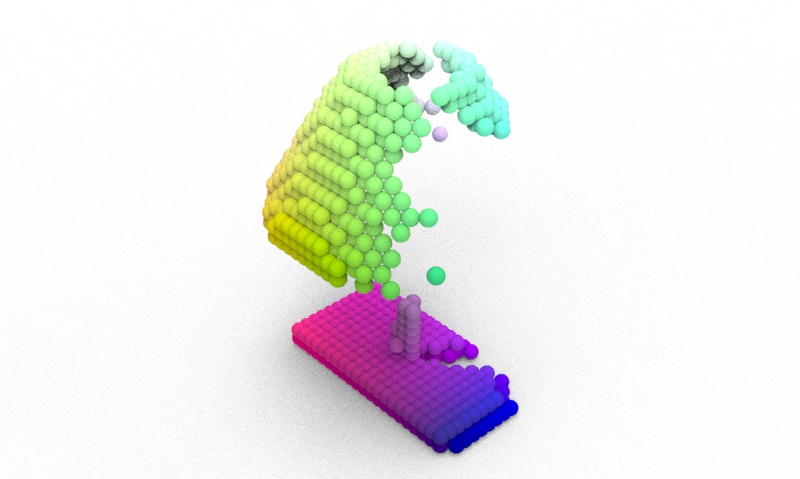}}
	\hspace{-3mm}
	\subfigure{
		\includegraphics[width=0.19\linewidth]{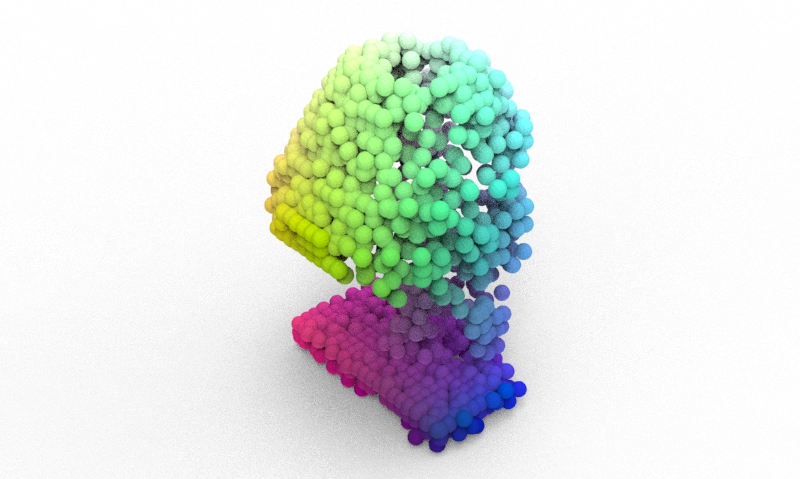}}
	\hspace{-3mm}
	\subfigure{
		\includegraphics[width=0.19\linewidth]{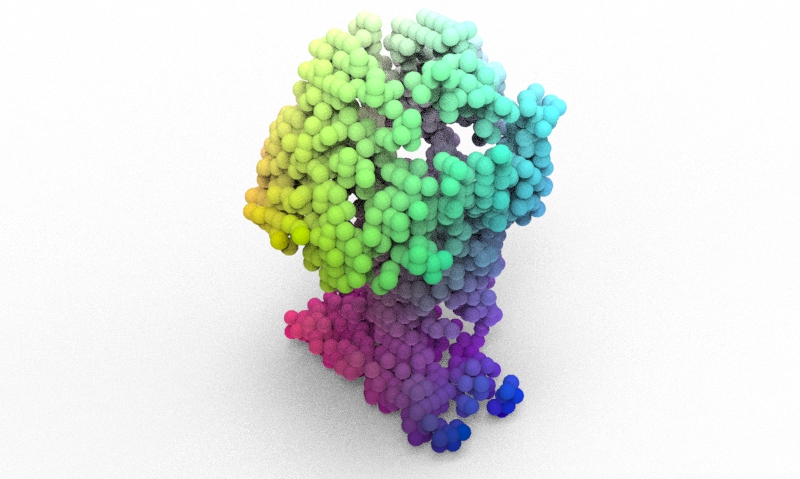}}
	\hspace{-3mm}
	\subfigure{
		\includegraphics[width=0.19\linewidth]{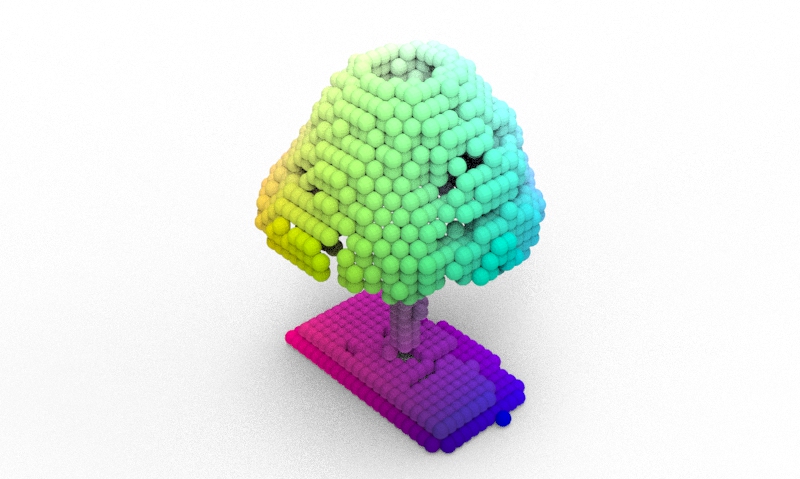}}
        \hspace{-3mm}
	\subfigure{
		\includegraphics[width=0.19\linewidth]{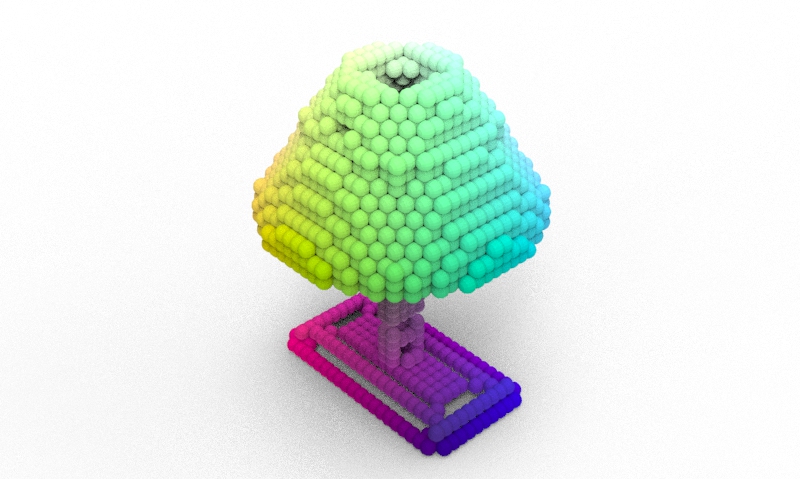}}
	\\
        \subfigure{
		\includegraphics[width=0.19\linewidth]{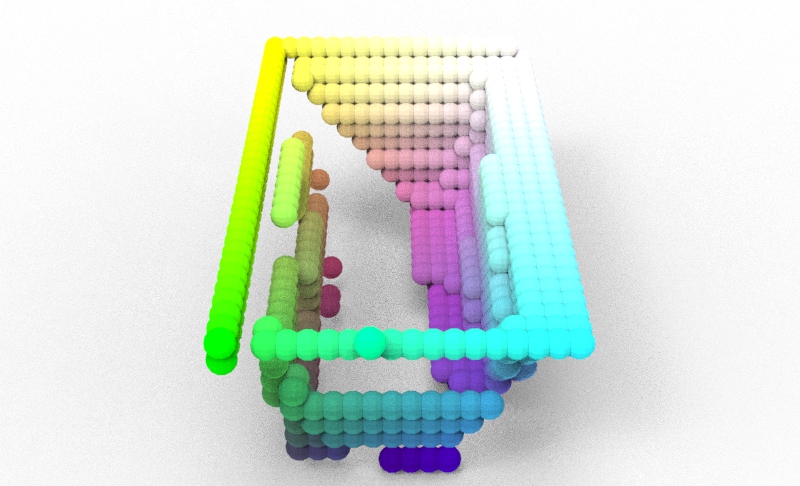}}
	\hspace{-3mm}
	\subfigure{
		\includegraphics[width=0.19\linewidth]{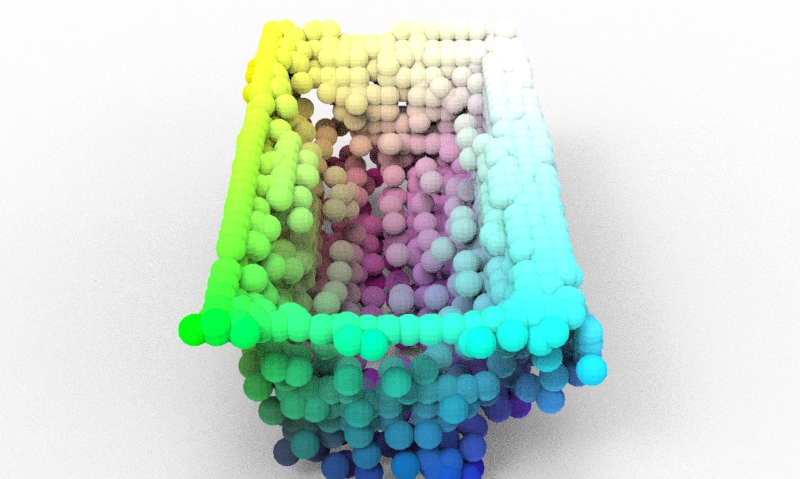}}
	\hspace{-3mm}
	\subfigure{
		\includegraphics[width=0.19\linewidth]{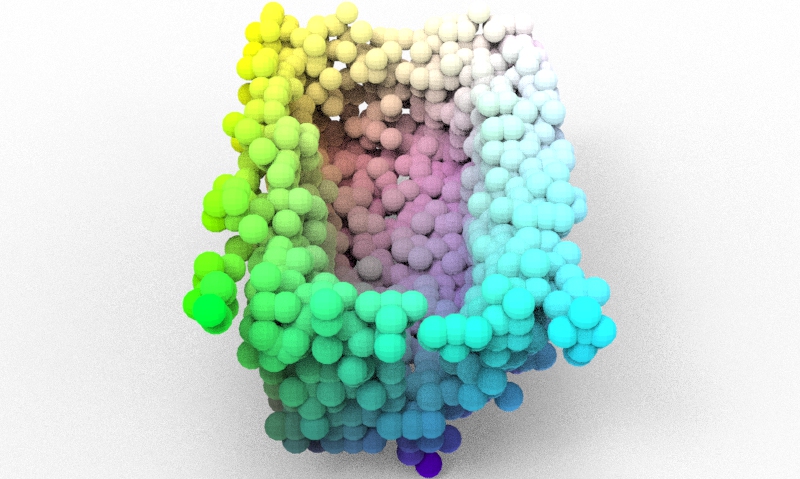}}
	\hspace{-3mm}
	\subfigure{
		\includegraphics[width=0.19\linewidth]{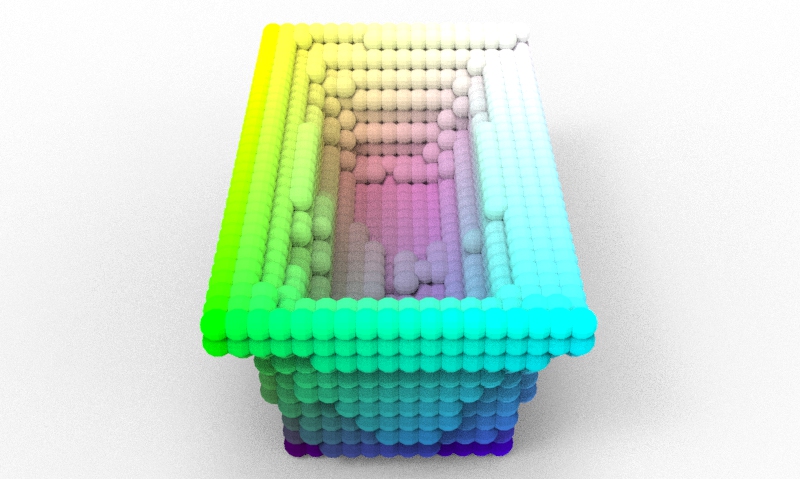}}
        \hspace{-3mm}
	\subfigure{
		\includegraphics[width=0.19\linewidth]{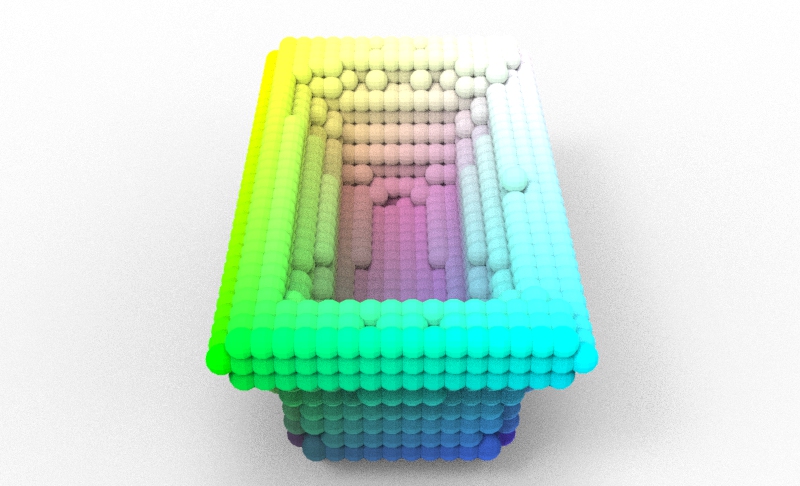}}
	\\
        \subfigure{
		\includegraphics[width=0.19\linewidth]{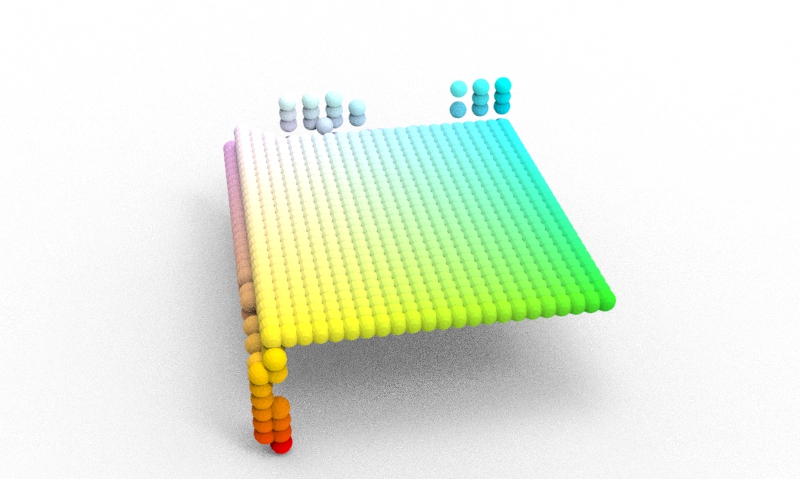}}
	\hspace{-3mm}
	\subfigure{
		\includegraphics[width=0.19\linewidth]{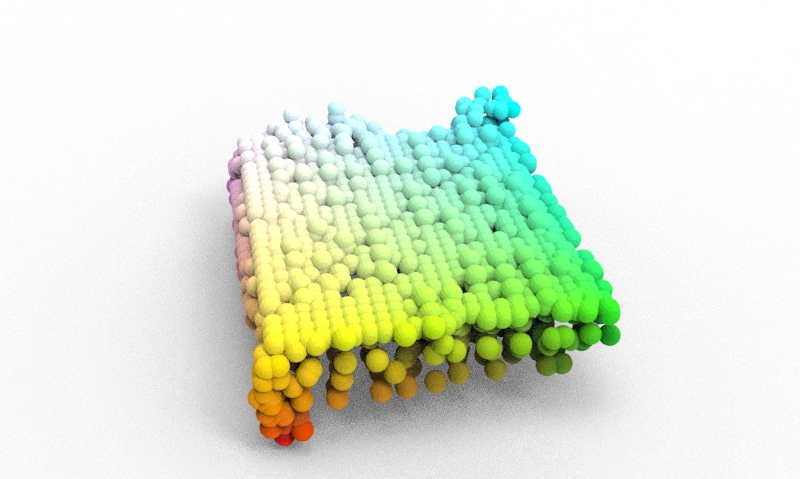}}
	\hspace{-3mm}
	\subfigure{
		\includegraphics[width=0.19\linewidth]{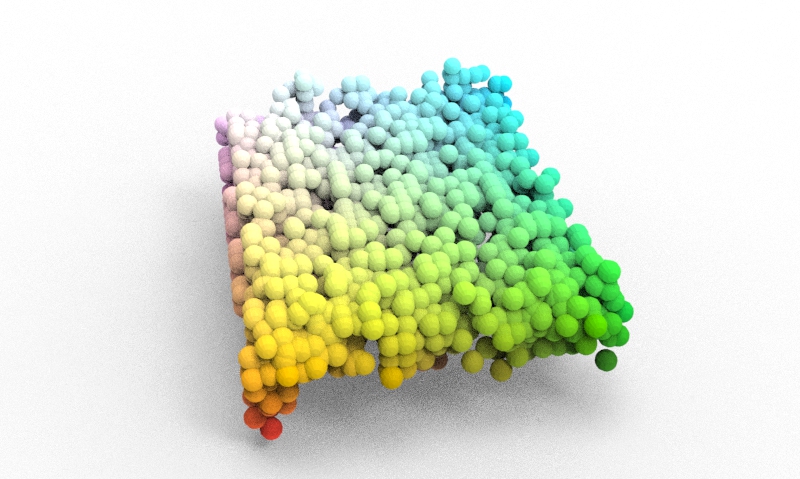}}
	\hspace{-3mm}
	\subfigure{
		\includegraphics[width=0.19\linewidth]{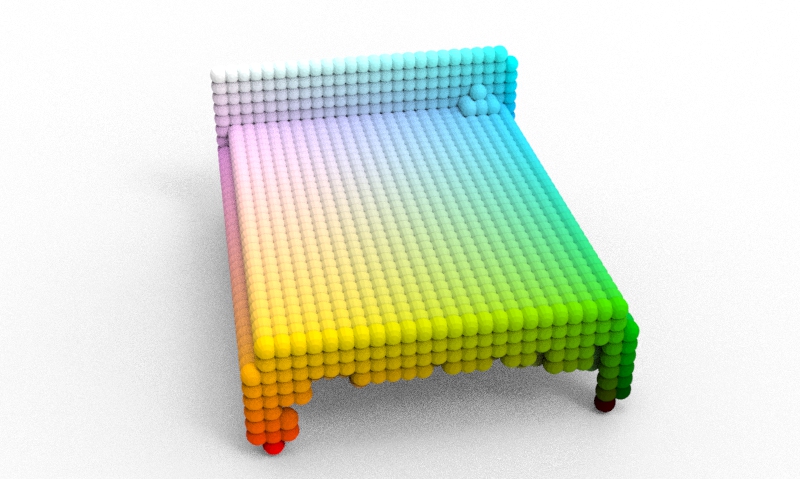}}
        \hspace{-3mm}
	\subfigure{
		\includegraphics[width=0.19\linewidth]{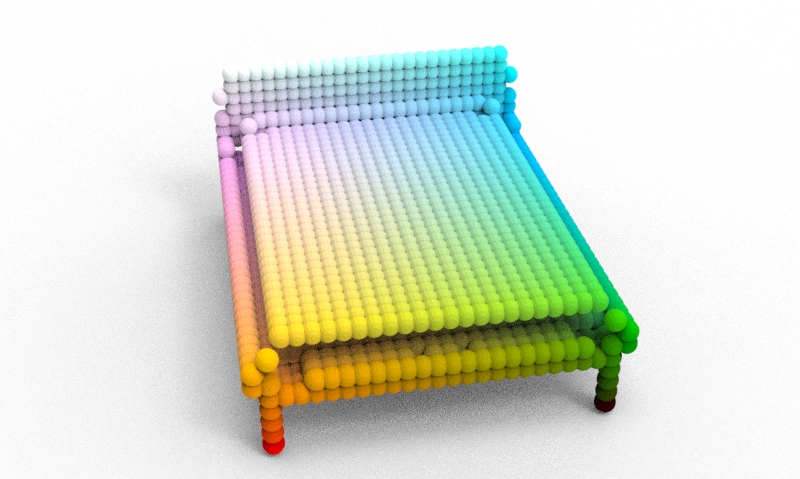}}
	\\
        \subfigure{
		\includegraphics[width=0.19\linewidth]{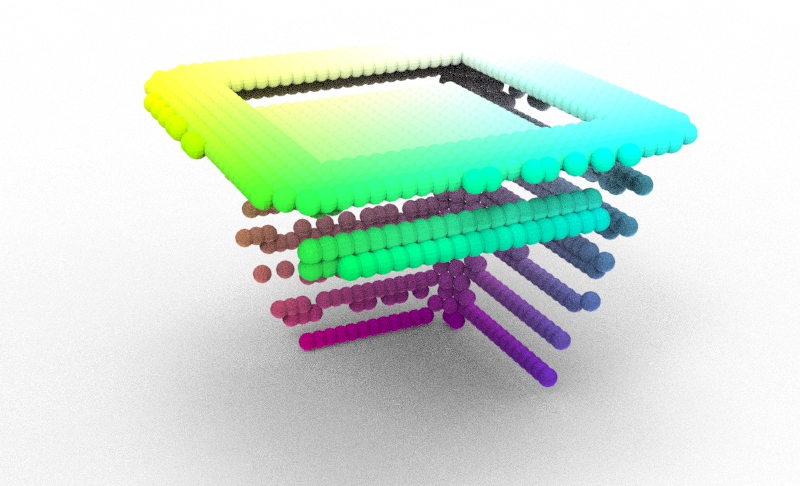}}
	\hspace{-3mm}
	\subfigure{
		\includegraphics[width=0.19\linewidth]{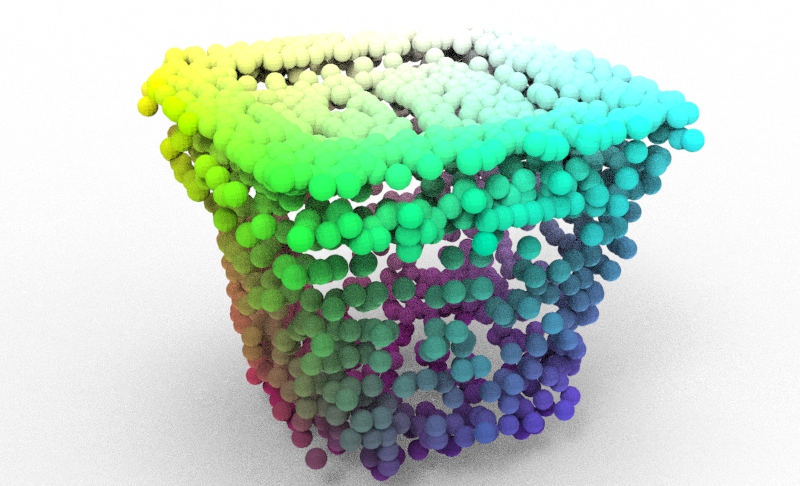}}
	\hspace{-3mm}
	\subfigure{
		\includegraphics[width=0.19\linewidth]{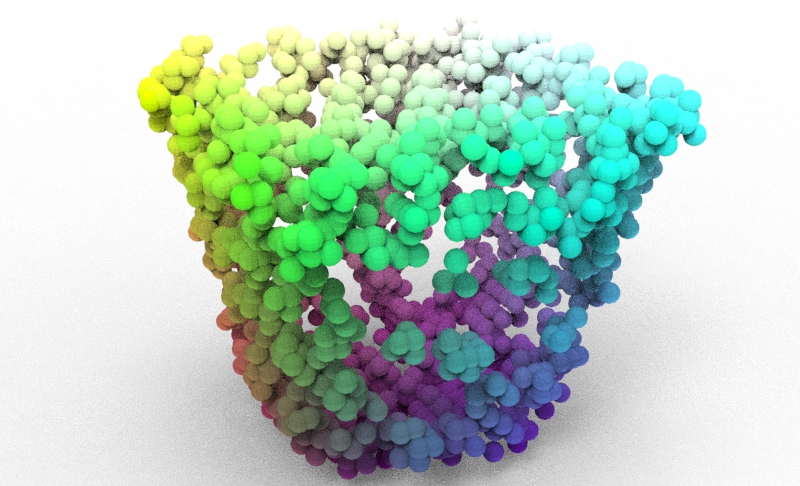}}
	\hspace{-3mm}
	\subfigure{
		\includegraphics[width=0.19\linewidth]{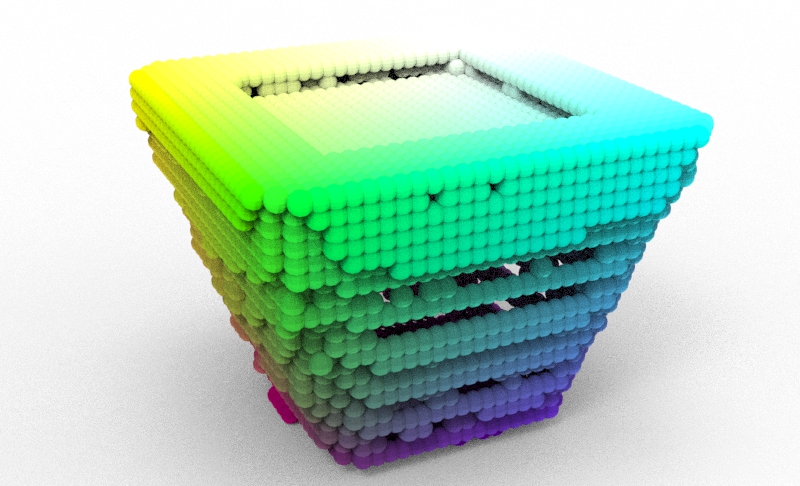}}
        \hspace{-3mm}
	\subfigure{
		\includegraphics[width=0.19\linewidth]{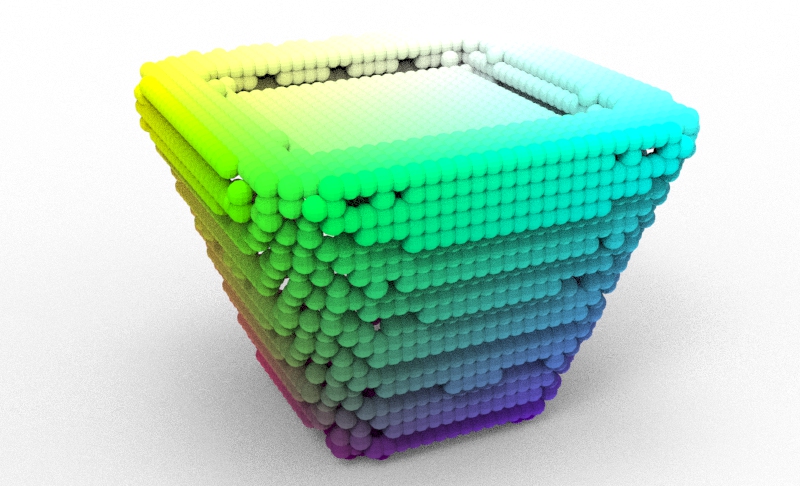}}
	\\
        \subfigure{
		\includegraphics[width=0.19\linewidth]{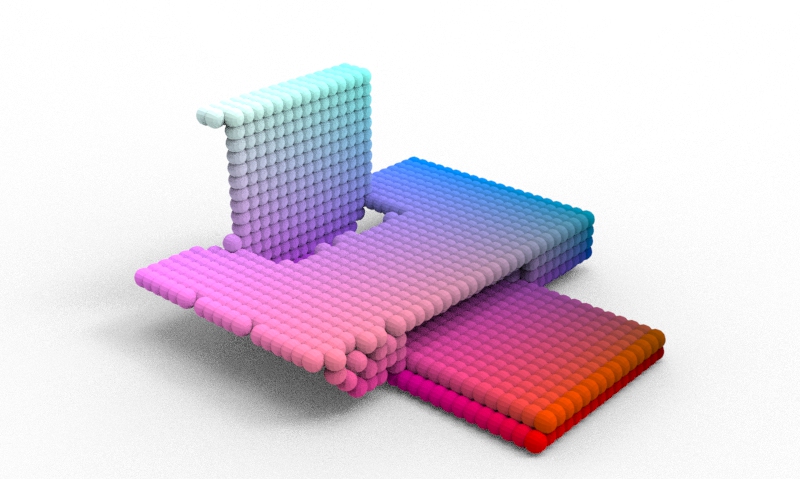}}
	\hspace{-3mm}
	\subfigure{
		\includegraphics[width=0.19\linewidth]{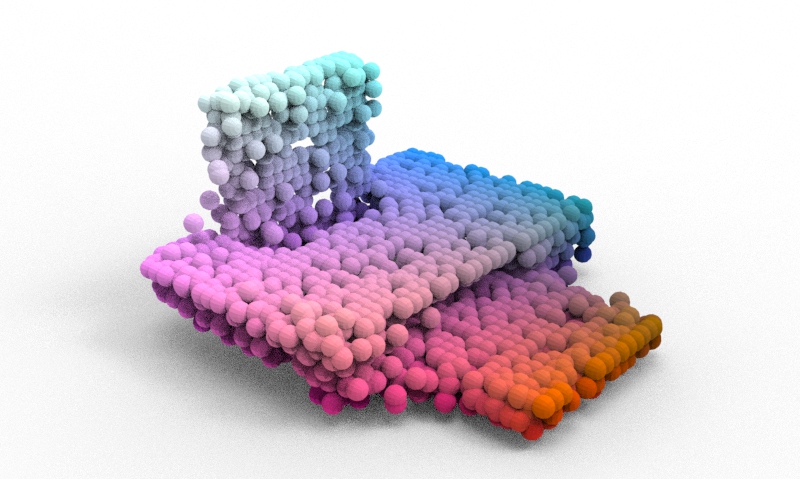}}
	\hspace{-3mm}
	\subfigure{
		\includegraphics[width=0.19\linewidth]{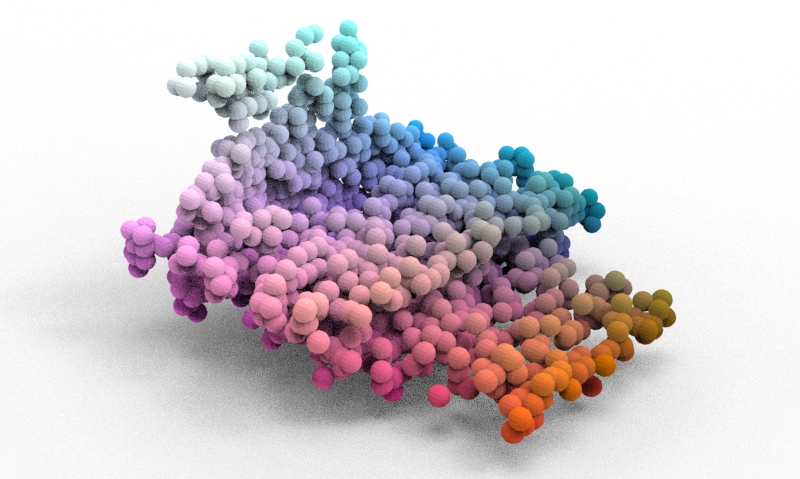}}
	\hspace{-3mm}
	\subfigure{
		\includegraphics[width=0.19\linewidth]{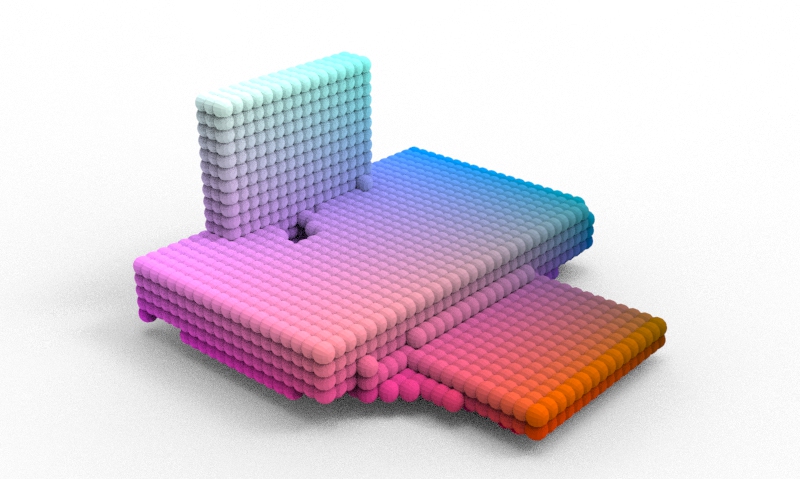}}
        \hspace{-3mm}
	\subfigure{
		\includegraphics[width=0.19\linewidth]{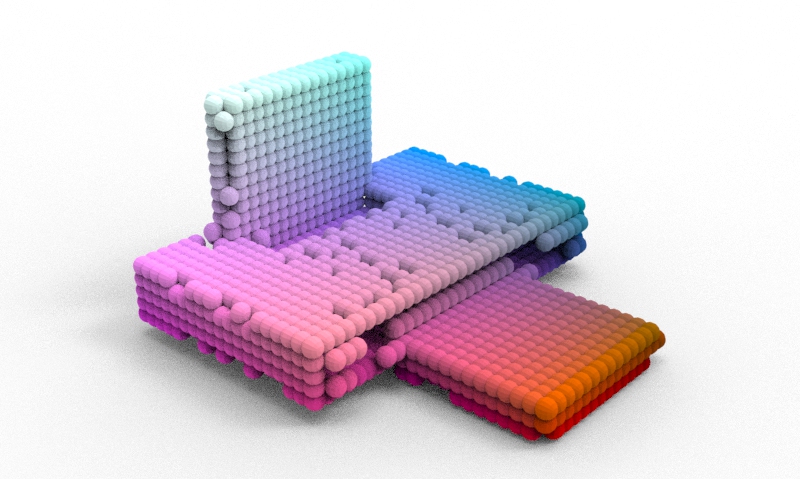}}
	\\
        \subfigure{
		\includegraphics[width=0.19\linewidth]{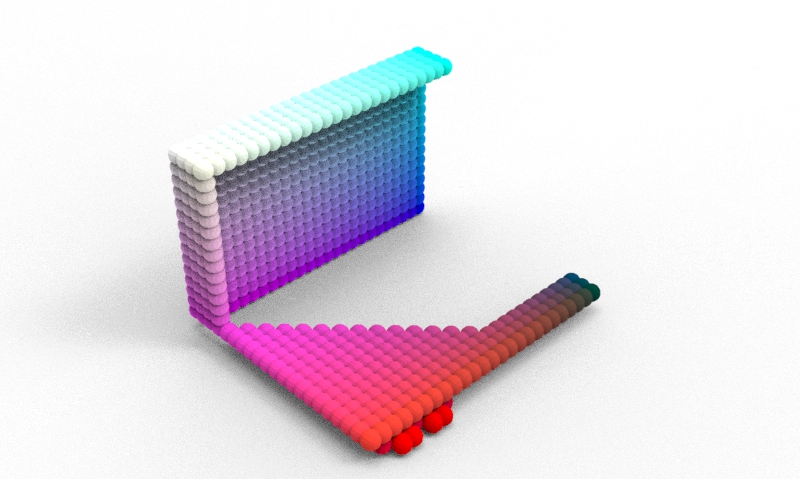}}
	\hspace{-3mm}
	\subfigure{
		\includegraphics[width=0.19\linewidth]{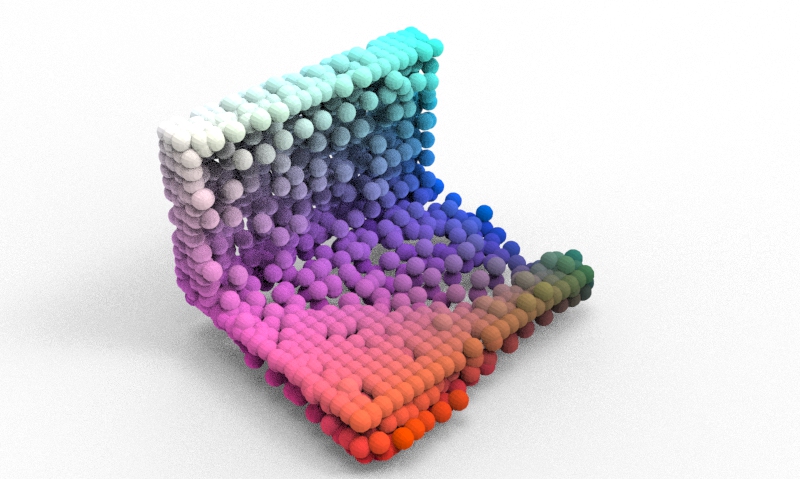}}
	\hspace{-3mm}
	\subfigure{
		\includegraphics[width=0.19\linewidth]{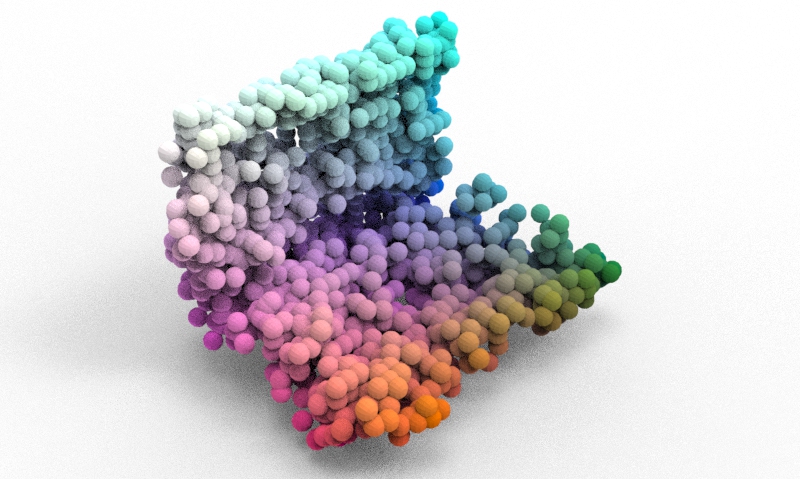}}
	\hspace{-3mm}
	\subfigure{
		\includegraphics[width=0.19\linewidth]{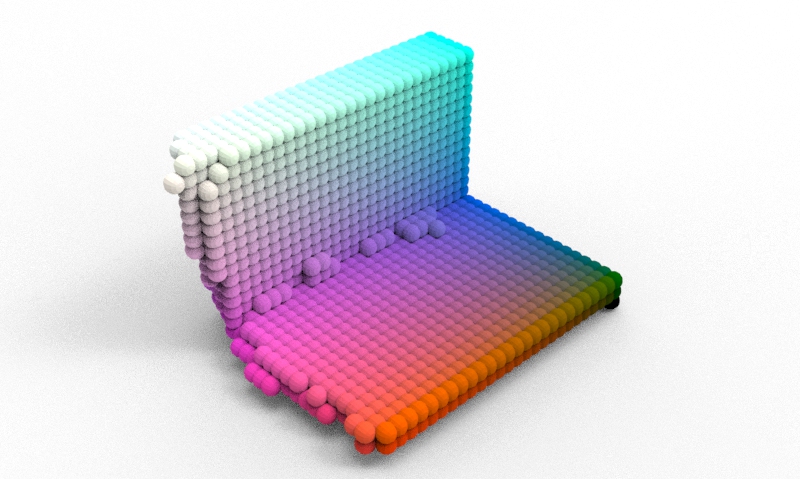}}
        \hspace{-3mm}
	\subfigure{
		\includegraphics[width=0.19\linewidth]{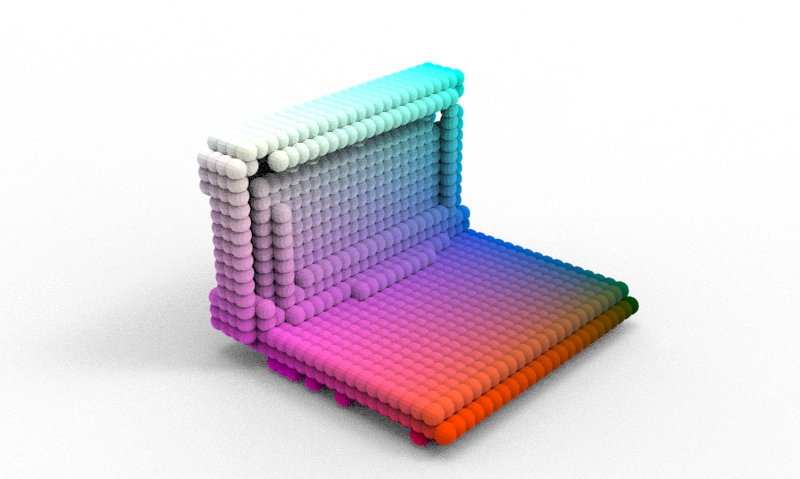}}
	\\
        \setcounter{subfigure}{0}
	\subfigure[]{
		\includegraphics[width=0.19\linewidth]{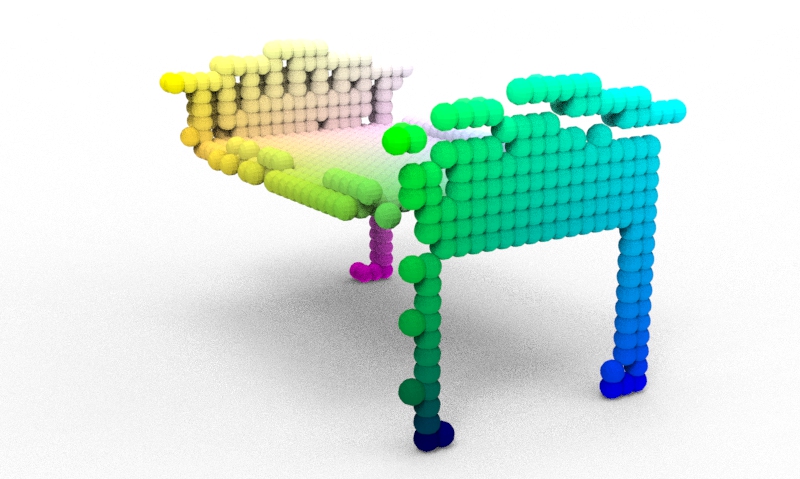}}
	\hspace{-3mm}
	\subfigure[]{
		\includegraphics[width=0.19\linewidth]{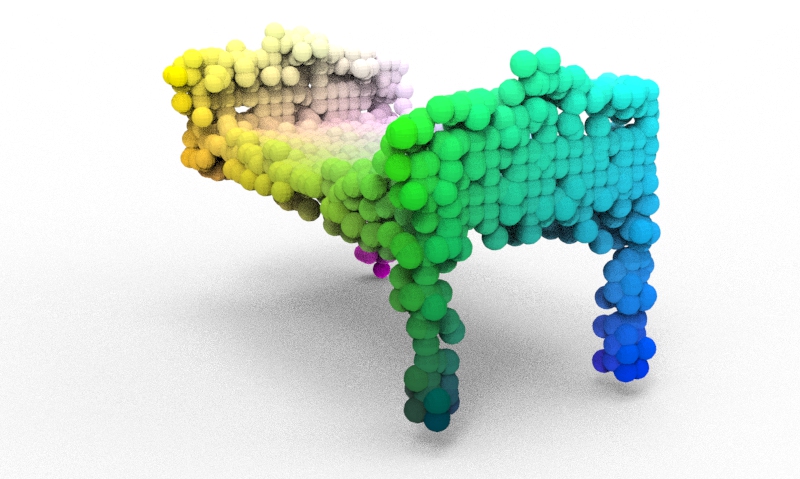}}
	\hspace{-3mm}
	\subfigure[]{
		\includegraphics[width=0.19\linewidth]{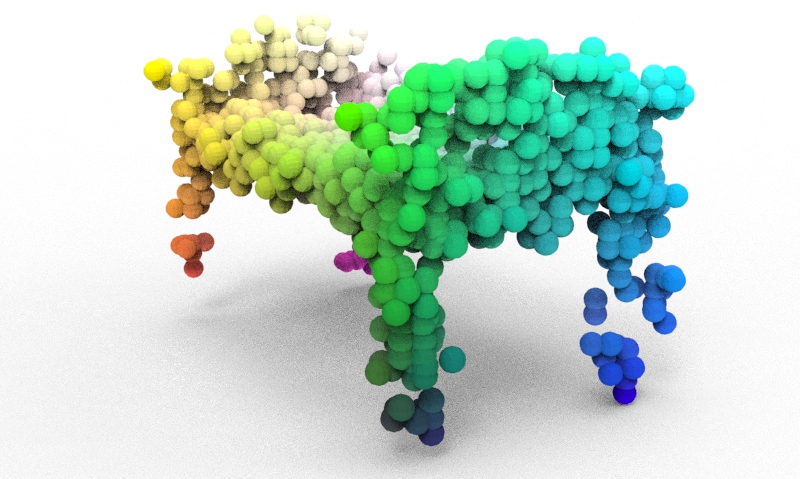}}
	\hspace{-3mm}
	\subfigure[]{
		\includegraphics[width=0.19\linewidth]{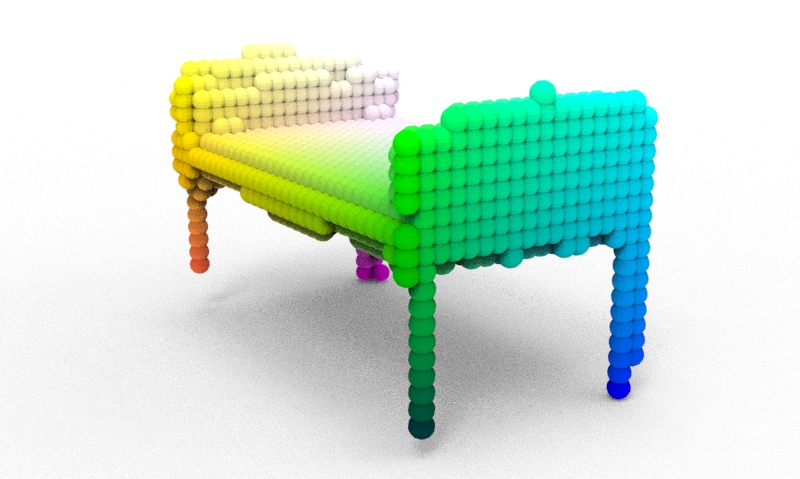}}
        \hspace{-3mm}
	\subfigure[]{
		\includegraphics[width=0.19\linewidth]{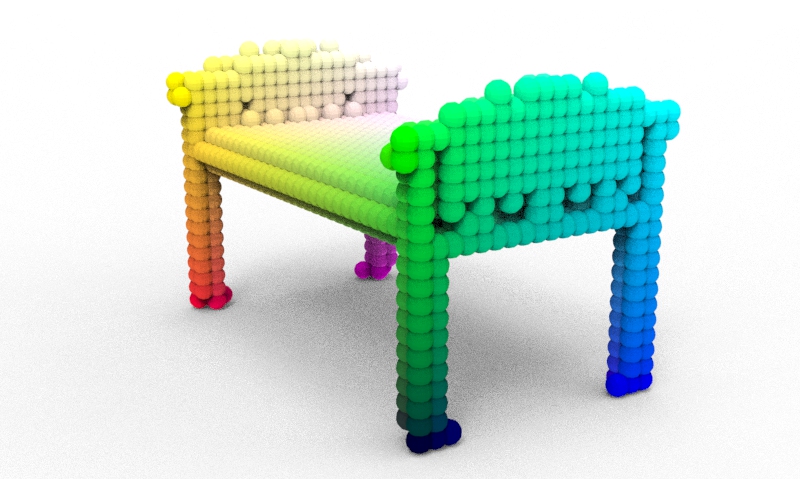}}
	\\
	
	\caption{Comparison of visual results between our methods and point cloud completion methods. (a) Partial input, (b) GRNet \cite{xie2020grnet}, (c) AnchorFormer \cite{chen2023anchorformer}, (d) Ours, (e) Ground truth.}
	\label{point_completion_shapenet}
\end{figure}

To provide a more comprehensive evaluation of our method, we also compared it with three SOTA point cloud completion approaches, i.e., PCN \cite{yuan2018pcn}, GRNet \cite{xie2020grnet} and AnchorFormer \cite{chen2023anchorformer}. For a fair comparison, the number of output points is set to 2048, which is close to the mean point number (1922) of the complete point clouds converted from voxel grids in our dataset. All methods were trained using data from seen categories and subsequently tested on unseen categories. As depicted in Table \ref{shapenet_completion}, our method outperforms all other point cloud completion methods across all categories on ShapeNet data. In terms of average CD, our method surpasses the second-ranked method by 1.62. Regarding the ScanNet dataset, as shown in Table \ref{scannet_completion}, our method significantly outperforms point cloud completion methods in 5/6 categories on the ScanNet dataset, and the average CD is 0.4 lower than the AnchorFormer.
Figure \ref{point_completion_shapenet} presents the visual results of our model and point cloud approaches. It is evident that our method accurately reconstructs the complete shapes while recovering the details, such as the corners of the bathtub and the legs of the bed. Moreover, our model generates smoother surfaces than the compared point cloud methods.

\subsubsection{Analysis of computational complexity and model size}
\revise{Table \ref{flops_comparison} shows the FLOPs (Floating Point Operations) and model sizes of our PSLN and PatchComplete under different resolutions of the used data. It is evident that at a resolution of 32, our PSLN achieves superior results compared to PatchComplete while consuming only half the number of FLOPs and a quarter of the model size. Increasing to a resolution of 64, the FLOPs of our PSLN decrease to approximately 1/25 of those of PatchComplete, accompanied by a model size that is merely 1/7. The increase in model size is only 6M in comparison to the values observed at a resolution of 32. These minimal increases are attributed to our model's ability to downsample priors and partial inputs at the feature level, reducing the computational cost of attention maps. Moreover, our PSLN can be trained end-to-end, whereas PatchComplete requires four-time training.}

\begin{table}[htbp]
	\centering
    \renewcommand\arraystretch{1.25}
	\caption{ \revise{Comparison of FLOPs and model size.``Res." means the resolution of the used voxel grids. The best results are highlighted in bold}.}
	\label{flops_comparison}
	\begin{tabular}{l|c|c|c}
		\toprule[1.2pt]
		Methods         & FLOPs (G)   & Model size (M)  & IoU \\ \hline  
		PatchComplete (Res. 32)   & 1335.1 & 697.7 & 0.429 \\   
		PSLN (Res. 32)   & \textbf{652.1} & \textbf{180.8} & \textbf{0.441} \\ \hline
            PatchComplete (Res. 64)   & 30748.5 & 1339.1 & 0.562 \\ 
            PSLN  (Res. 64)    & \textbf{1235.8} & \textbf{186.7} & \textbf{0.598} \\ \bottomrule[1.2pt] \hline
	\end{tabular}
\end{table}

\begin{figure*}[htbp]
	\centering  
	\subfigbottomskip=1pt 
        \setcounter{subfigure}{0}
	%\quad
	\subfigure[]{
		\includegraphics[width=0.23\linewidth]{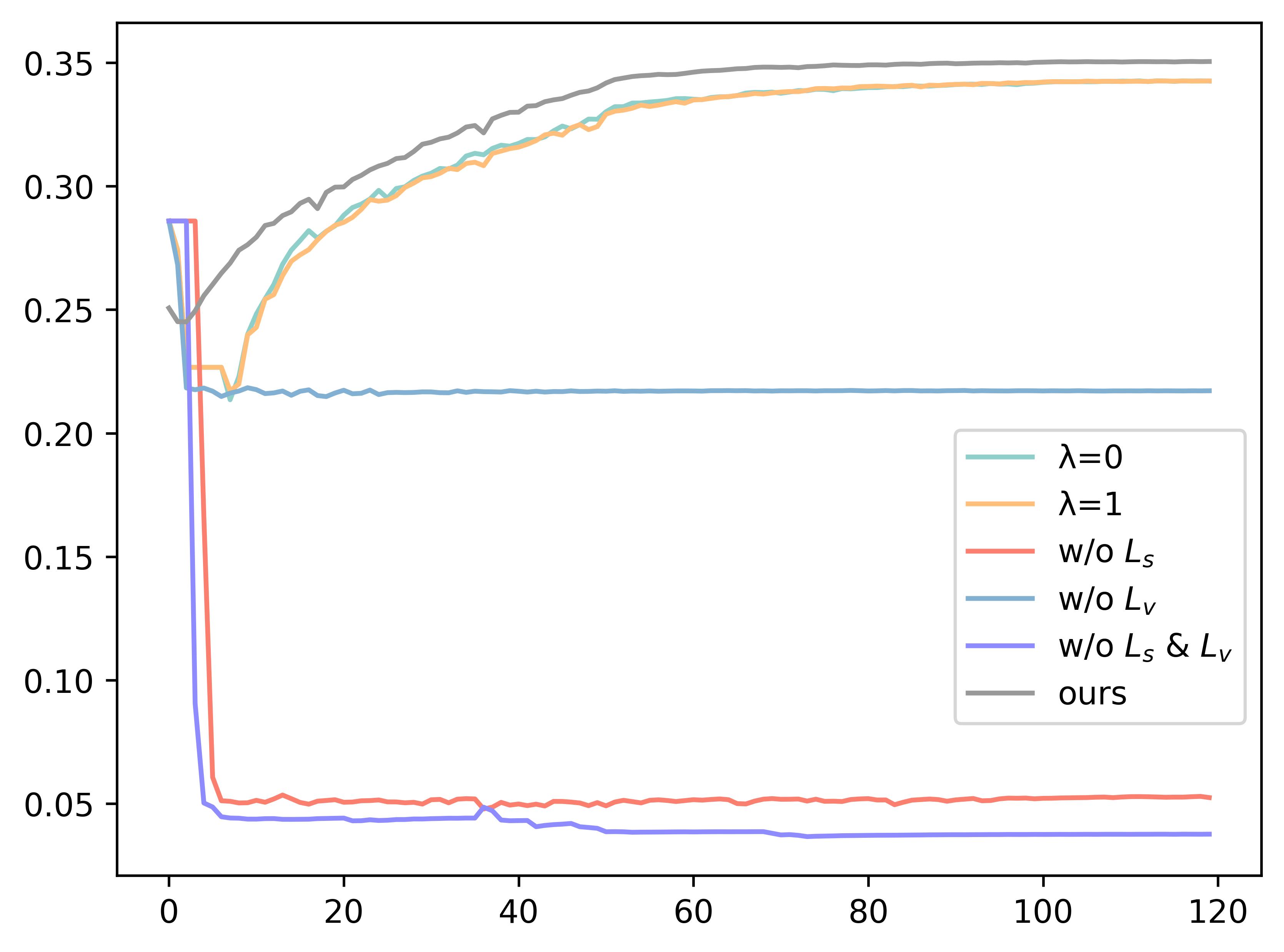}}
	\subfigure[]{
		\includegraphics[width=0.23\linewidth]{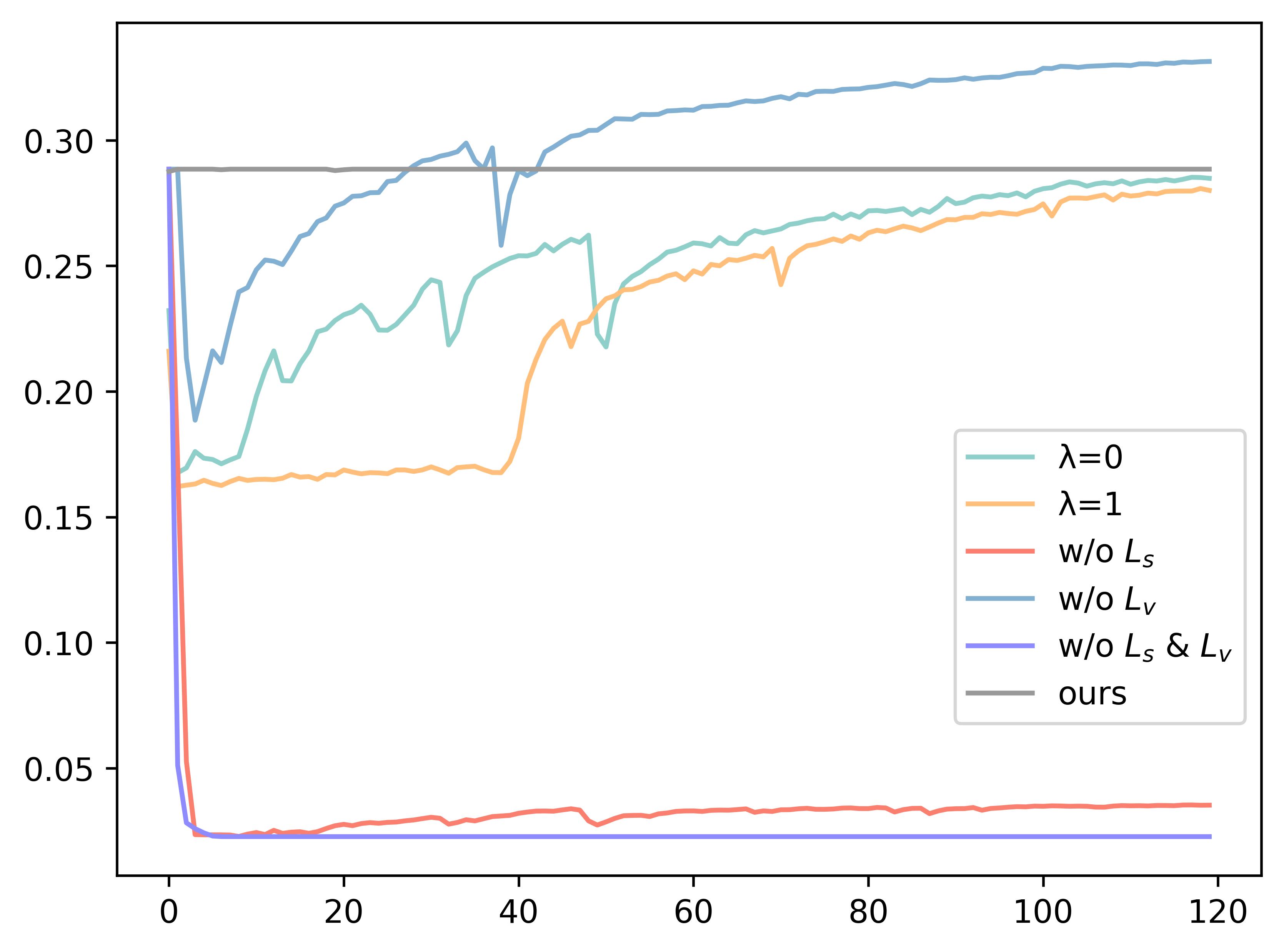}}
	\subfigure[]{
		\includegraphics[width=0.23\linewidth]{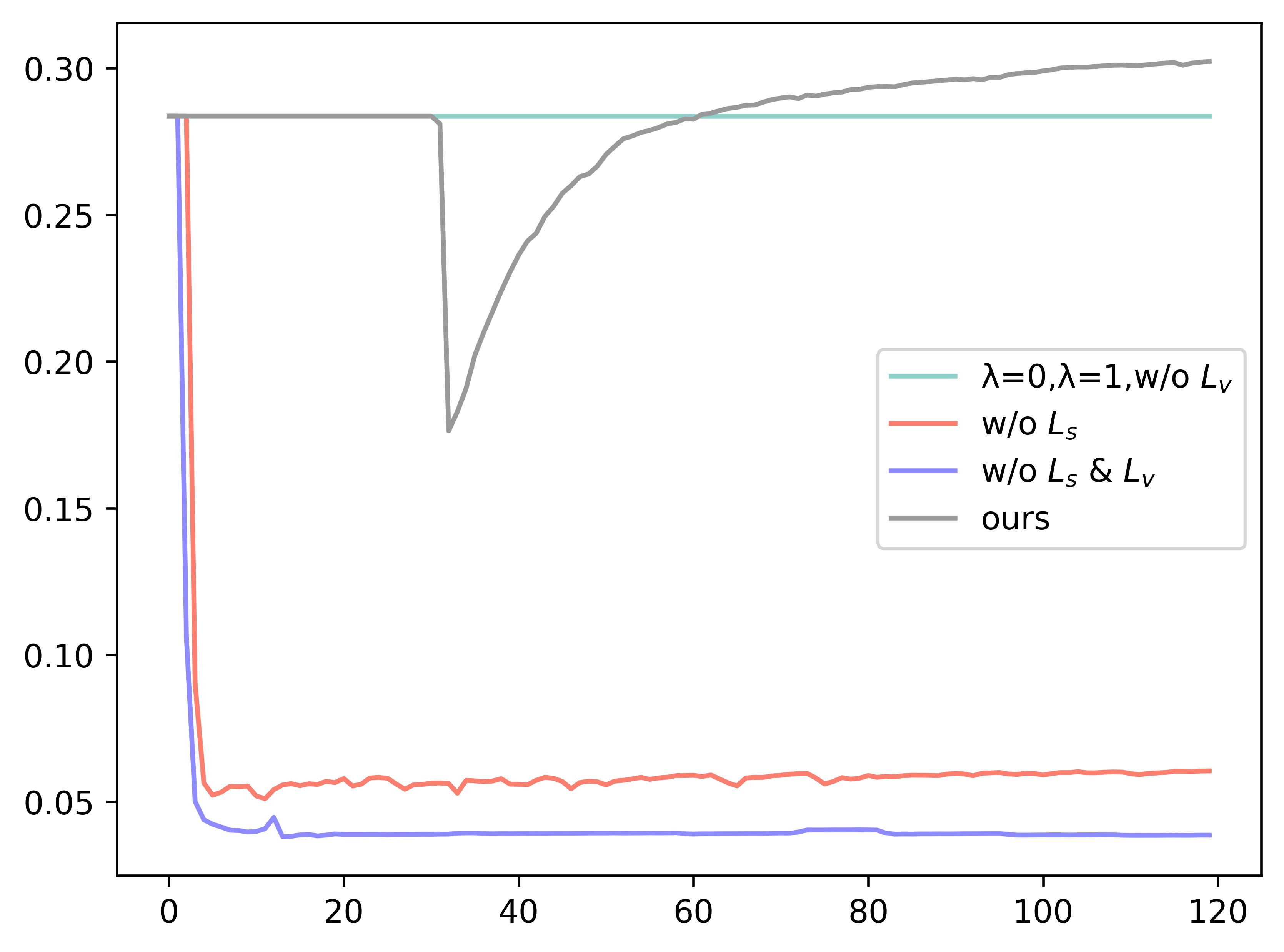}}
	\subfigure[]{
		\includegraphics[width=0.23\linewidth]{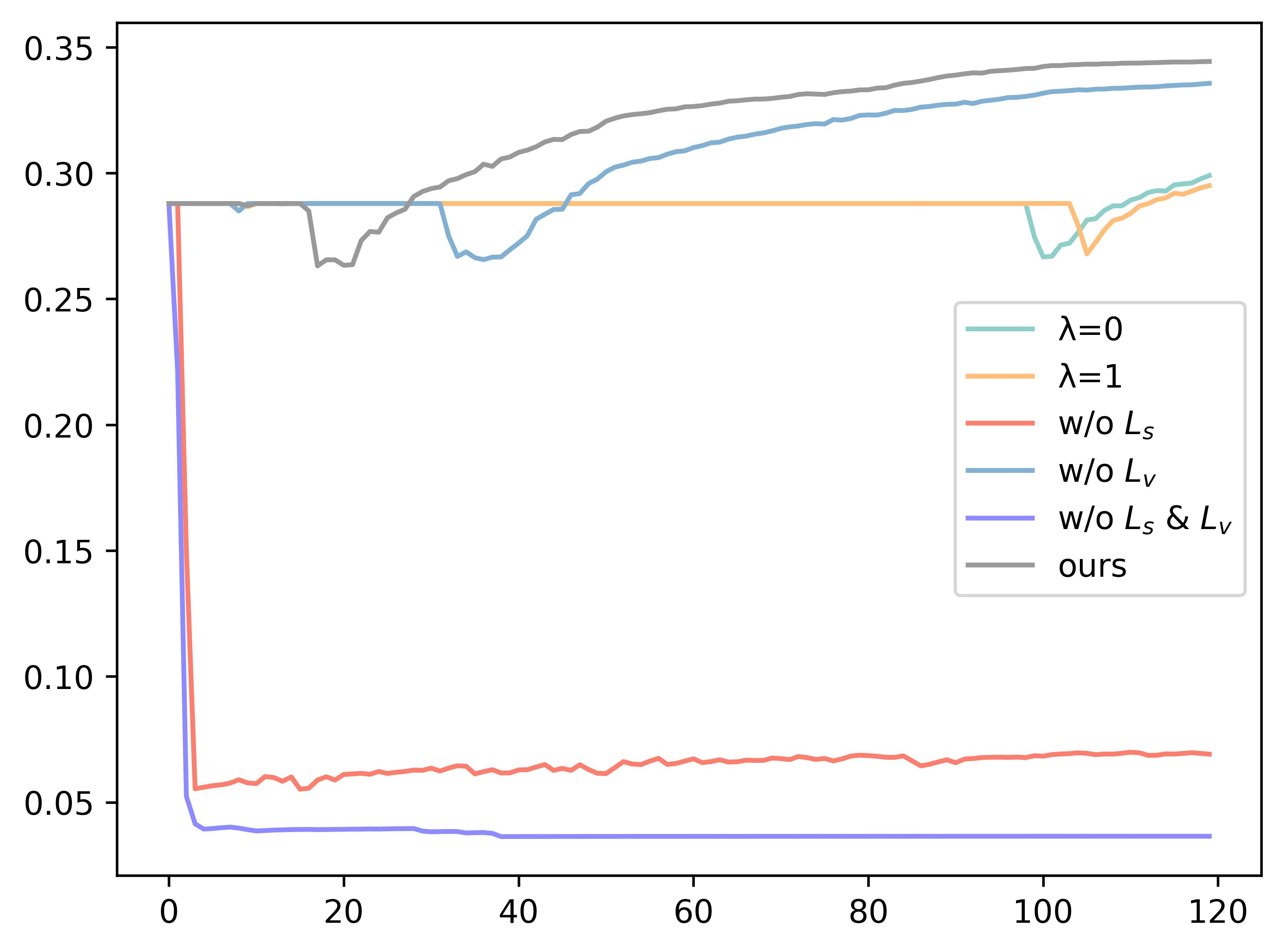}}
        \\
        \subfigure[]{
		\includegraphics[width=0.23\linewidth]{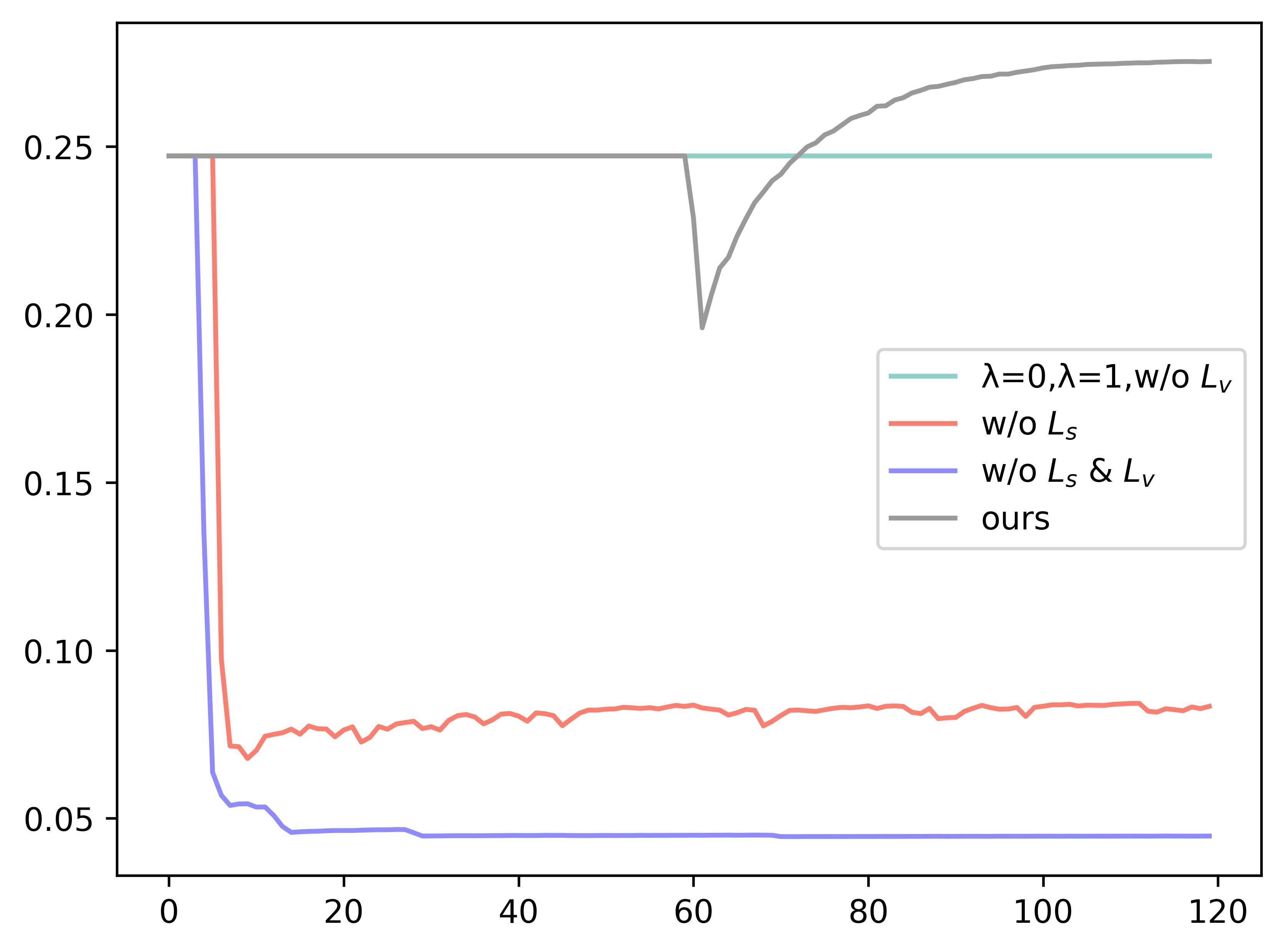}}
	\subfigure[]{
		\includegraphics[width=0.23\linewidth]{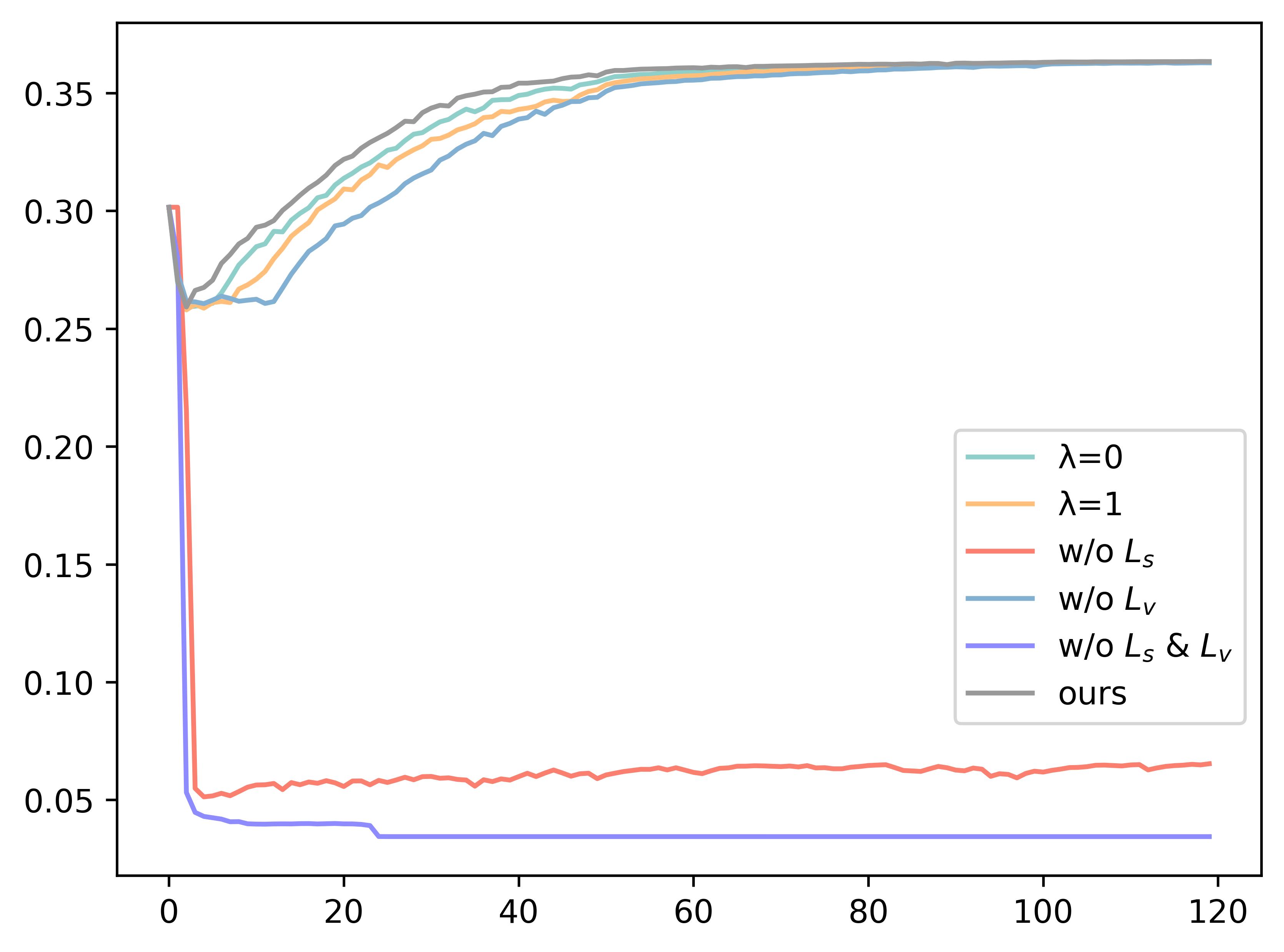}}
	\subfigure[]{
		\includegraphics[width=0.23\linewidth]{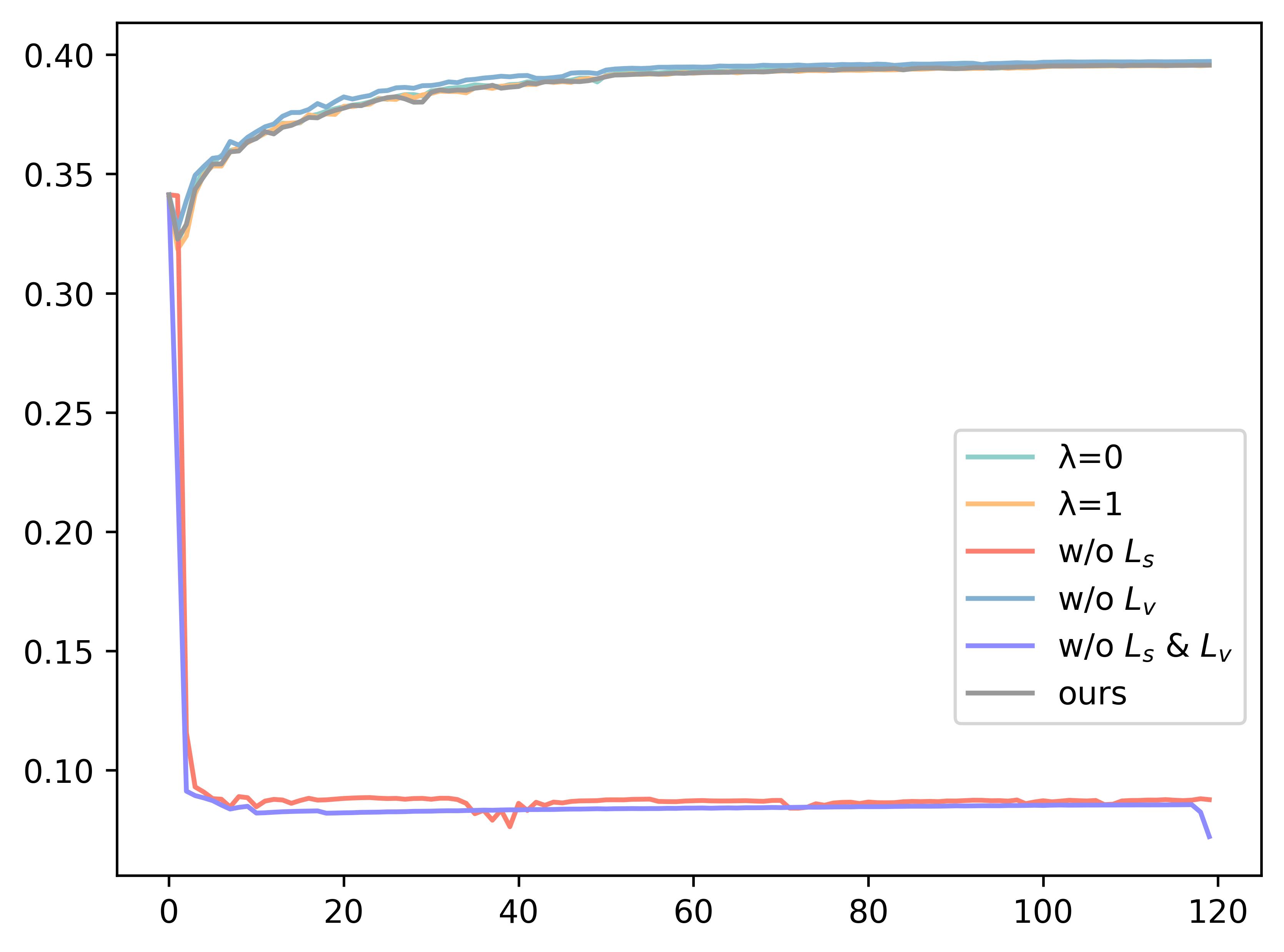}}
	\subfigure[]{
		\includegraphics[width=0.23\linewidth]{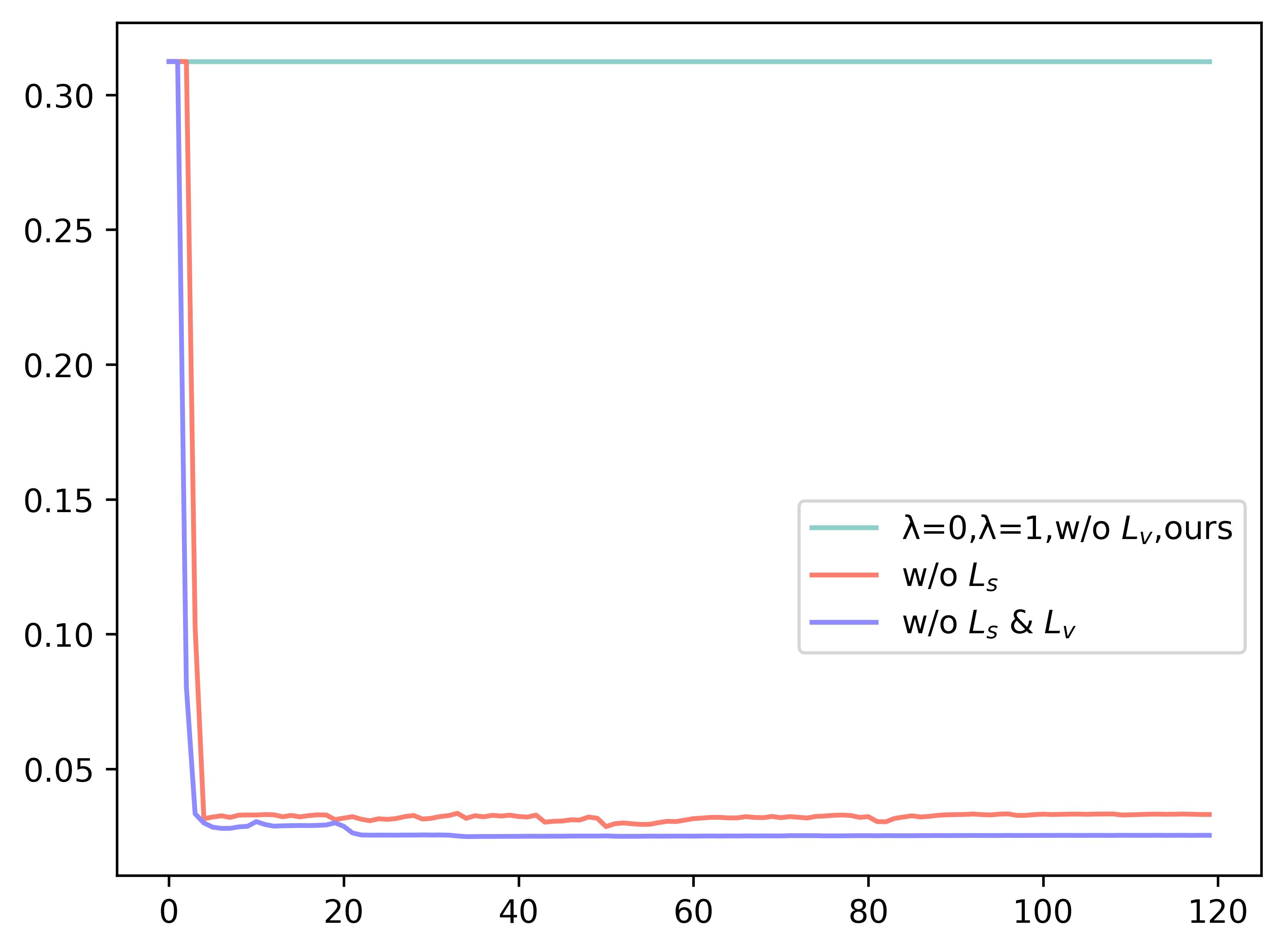}}
	\caption{The IoU curve during the training process of CaSR on the ShapeNet dataset. The y-axis represents the IoU between the predicted complete shape and the partial input, while the x-axis represents the training epoch. (a) bag, (b) lamp, (c) bathtub, (d) bed, (e) basket, (f) printer, (g) laptop, (h) bench.}
	\label{training_iou}
\end{figure*}

\begin{table*}[htbp]
	\centering
        \setlength{\tabcolsep}{4pt}
	\renewcommand\arraystretch{1.25}
	\caption{Effects of the CoSL and CaSR. The reported results are IoU\(\uparrow\). ``Avg." means average results across all categories. The best results are highlighted in bold. ``MSL" means Multi-scale Structure Learning unit.}
	\label{ablation_pln}
        \scalebox{0.975}{
	\begin{tabular}{c|ccccccccc|ccccccc}
		\toprule[1.2pt]
           \multirow{2}{*}{Methods} & \multicolumn{9}{c|}{ShapeNet}                                                                                                                                                                            & \multicolumn{7}{c}{ScanNet}                                                                                                                             \\ \cline{2-17} 
                  & \multicolumn{1}{c|}{Avg.} & \multicolumn{1}{c|}{Bag} & \multicolumn{1}{c|}{Lamp} & \multicolumn{1}{c|}{Bathtub} & \multicolumn{1}{c|}{Bed} & \multicolumn{1}{c|}{Basket} & \multicolumn{1}{c|}{Printer} & \multicolumn{1}{c|}{Laptop} & Bench & \multicolumn{1}{c|}{Avg.} & \multicolumn{1}{c|}{Bag} & \multicolumn{1}{c|}{Lamp} & \multicolumn{1}{c|}{Bathtub} & \multicolumn{1}{c|}{Bed} & \multicolumn{1}{c|}{Basket} & Printer \\ \hline
                  PatchComplete & \multicolumn{1}{c|}{0.429} & \multicolumn{1}{c|}{0.391} & \multicolumn{1}{c|}{0.374} & \multicolumn{1}{c|}{0.493} & \multicolumn{1}{c|}{0.368} & \multicolumn{1}{c|}{0.490} & \multicolumn{1}{c|}{0.396} & \multicolumn{1}{c|}{0.461} & 0.456 & \multicolumn{1}{c|}{0.370} & \multicolumn{1}{c|}{0.351} & \multicolumn{1}{c|}{\textbf{0.369}} & \multicolumn{1}{c|}{0.407} & \multicolumn{1}{c|}{0.337} & \multicolumn{1}{c|}{0.423} & 0.336 \\
                  CoSL w/o MSL& \multicolumn{1}{c|}{0.409} & \multicolumn{1}{c|}{0.380} & \multicolumn{1}{c|}{0.355} & \multicolumn{1}{c|}{0.450} & \multicolumn{1}{c|}{0.345} & \multicolumn{1}{c|}{0.453} & \multicolumn{1}{c|}{0.368} & \multicolumn{1}{c|}{0.424} & \textbf{0.497} & \multicolumn{1}{c|}{0.331} & \multicolumn{1}{c|}{0.322} & \multicolumn{1}{c|}{0.347} & \multicolumn{1}{c|}{0.361} & \multicolumn{1}{c|}{0.319} & \multicolumn{1}{c|}{0.352} & 0.286 \\ 
                  CoSL w/ MSL& \multicolumn{1}{c|}{0.441} & \multicolumn{1}{c|}{0.391} & \multicolumn{1}{c|}{\textbf{0.380}} & \multicolumn{1}{c|}{0.525} & \multicolumn{1}{c|}{0.383} & \multicolumn{1}{c|}{0.493} & \multicolumn{1}{c|}{0.392} & \multicolumn{1}{c|}{0.472} & 0.492 & \multicolumn{1}{c|}{0.385} & \multicolumn{1}{c|}{0.354} & \multicolumn{1}{c|}{0.362} & \multicolumn{1}{c|}{0.438} & \multicolumn{1}{c|}{0.369} & \multicolumn{1}{c|}{0.445} & \textbf{0.343} \\ 
                  CoSL+CaSR & \multicolumn{1}{c|}{\textbf{0.474}} & \multicolumn{1}{c|}{\textbf{0.450}} & \multicolumn{1}{c|}{\textbf{0.380}} & \multicolumn{1}{c|}{\textbf{0.542}} & \multicolumn{1}{c|}{\textbf{0.426}} & \multicolumn{1}{c|}{\textbf{0.509}} & \multicolumn{1}{c|}{\textbf{0.458}} & \multicolumn{1}{c|}{\textbf{0.533}} & 0.492 & \multicolumn{1}{c|}{\textbf{0.432}} & \multicolumn{1}{c|}{\textbf{0.384}} & \multicolumn{1}{c|}{0.362} & \multicolumn{1}{c|}{\textbf{0.529}} & \multicolumn{1}{c|}{\textbf{0.448}} & \multicolumn{1}{c|}{\textbf{0.523}} & \textbf{0.343}
			 \\ \bottomrule[1.2pt]
	\end{tabular}
    }
\end{table*}

\subsubsection{Relationship between the training results and hyperparameters}
\label{hyper_selections}
During the training process, our CaSR determines hyperparameters by evaluating the IoU between the predicted complete shape and the partial input. This approach can ensure that the selected hyperparameters are optimized to achieve the highest IoU. Figure \ref{training_iou} illustrates the impact of different hyperparameters on the training results. We discovered that the best training results were obtained when simultaneously using \(L_s\) and \(L_v\) and \(\lambda=0.5\). Moreover, the rankings of the training results align with the rankings of the test results across all categories, as shown in Table \ref{component}.
These findings affirm the rationality of selecting optimal hyperparameters based on the IoU between the predicted complete shape and the partial input.

\subsection{Ablation Study}

\begin{table*}[htbp]
	\centering
	\renewcommand\arraystretch{1.25}
	\caption{Results comparison of components of our refinement module on ShapeNet dataset in terms of IoU\(\uparrow\). The best results are highlighted in bold.}
	\label{component}
	\begin{tabular}{c|c|c|c|c|c|c|c|c|c|c|c|c}
		\toprule[1.2pt]
		\multicolumn{4}{c|}{Components} & \multicolumn{9}{c}{Categories}                   \\ \hline
		 \(L_s\)	& \(L_v\)	& \(\lambda\) & prior    & Average & Bag & Lamp & Bathtub & Bed & Basket	& Printer	& Laptop	& Bench \\ \hline
		\XSolid	& \XSolid	& 0.5 & specific   & 0.099  & 0.080 & 0.026 & 0.100 & 0.080 & 0.173 & 0.085	& 0.178	& 0.072 \\
		\Checkmark	& \XSolid	& 0.5  & specific  & 0.463  & 0.391 & \textbf{0.394} & 0.525 & 0.416 & 0.493 & \textbf{0.458}	& \textbf{0.534}	& \textbf{0.492} \\
		\XSolid	& \Checkmark	& 0.5 & specific   &  0.127 & 0.090 & 0.049 & 0.144 & 0.135 & 0.219 & 0.126	& 0.176	& 0.073 \\
  \hline
		\Checkmark	& \Checkmark	& 1.0 & specific &  0.458 & 0.440 & 0.349 & 0.525 & 0.376 & 0.493 & \textbf{0.458} & 	0.532 & \textbf{0.492} \\
		\Checkmark	& \Checkmark	& 0.0 & specific & 0.459 & 0.441 & 0.355 & 0.525 & 0.376 & 0.493 & \textbf{0.458} & 0.532 & \textbf{0.492} \\
            \Checkmark	& \Checkmark	& 0.5 & fixed & 0.460  & 0.441 &0.356  & 0.525 & 0.383 & 0.493	& \textbf{0.458} & 0.532 & \textbf{0.492}
          \\
        \Checkmark	& \Checkmark	& 0.5 & specific &  \textbf{0.474} & \textbf{0.450} & 0.380 & \textbf{0.542} & \textbf{0.426} & \textbf{0.509}	& \textbf{0.458} & 0.533 & \textbf{0.492} \\ 
        
        \bottomrule[1.2pt]
	\end{tabular}
\end{table*}

\subsubsection{Effects of CoSL and CaSR}
To evaluate the effectiveness of the CoSL and CaSR model, we conducted ablation studies on both ShapeNet and ScanNet. As Table \ref{ablation_pln} shows, without the Multi-scale Structure Learning unit, the CoSL's performance falls behind that of PatchComplete on both datasets. Nevertheless, after using the MSL unit, the model improves the IoU by 0.032 and 0.054 in ShapeNet and ScanNet, respectively, surpassing that of PatchComplete. This observation highlights the significant role of our MSL unit in learning complete shapes. This can be attributed to the incorporation of the correlations between multi-scale local patterns in the partial shape and priors.
Furthermore, after refinement by our  CaSR,  our model increases the IoU by 0.033 and 0.047 in ShapeNet and ScanNet, respectively. These results prove that both our CoSL and CaSR can achieve good performance.

\begin{figure}[htbp]
	\centering  
	\subfigbottomskip=1pt 
	\subfigure{
		\includegraphics[width=0.225\linewidth]{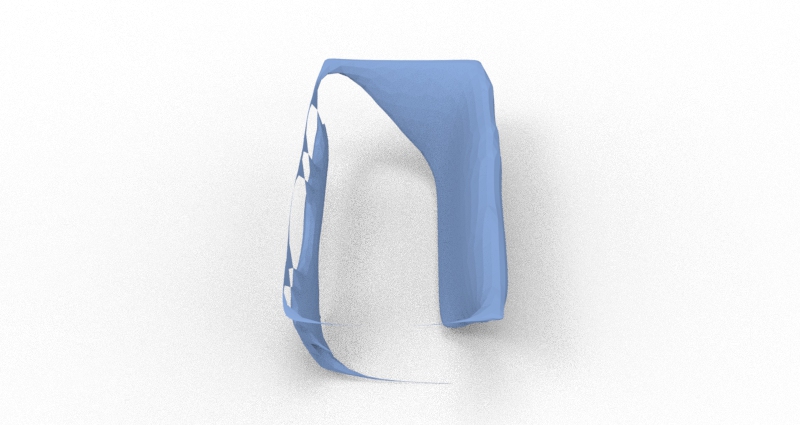}}
	\subfigure{
		\includegraphics[width=0.225\linewidth]{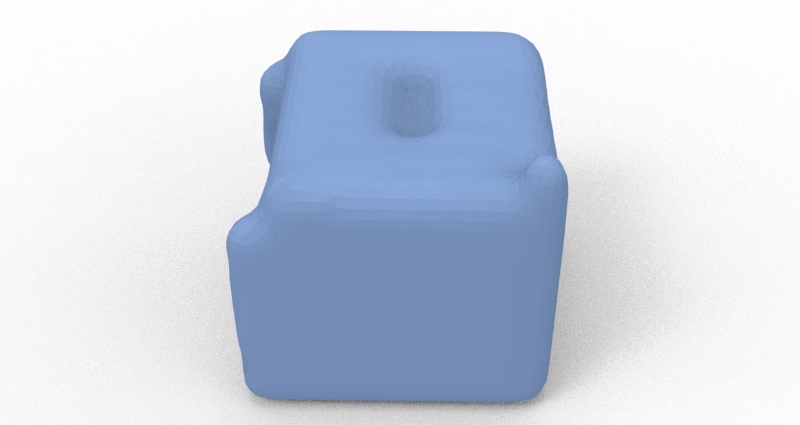}}
	\subfigure{
		\includegraphics[width=0.225\linewidth]{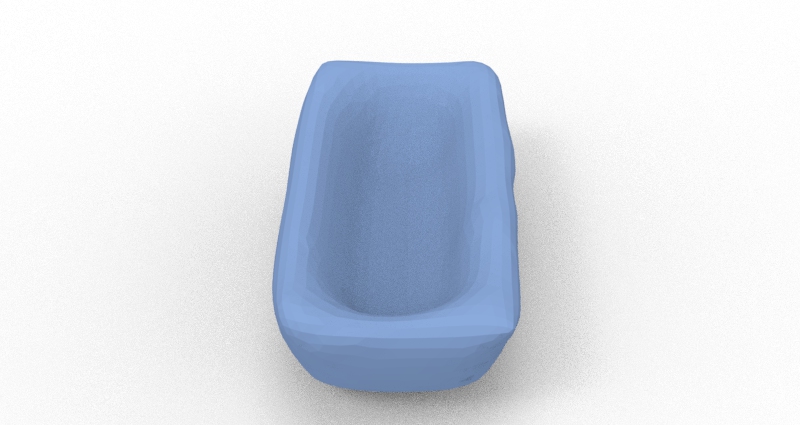}}
	\subfigure{
		\includegraphics[width=0.225\linewidth]{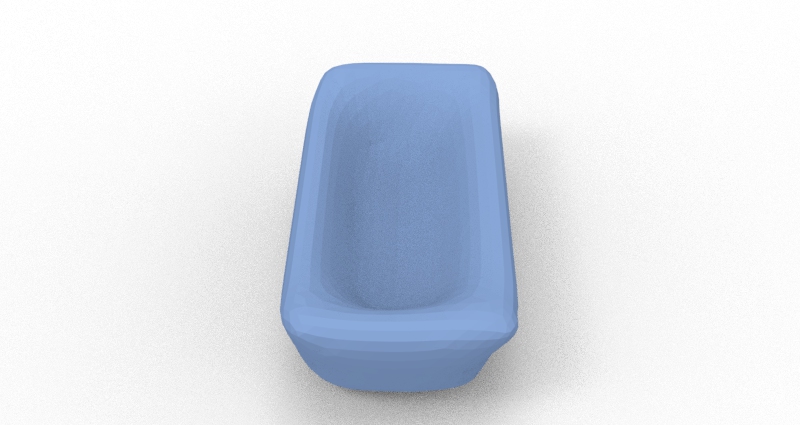}}
	\\
        \setcounter{subfigure}{0}
	%\quad
	\subfigure[]{
		\includegraphics[width=0.225\linewidth]{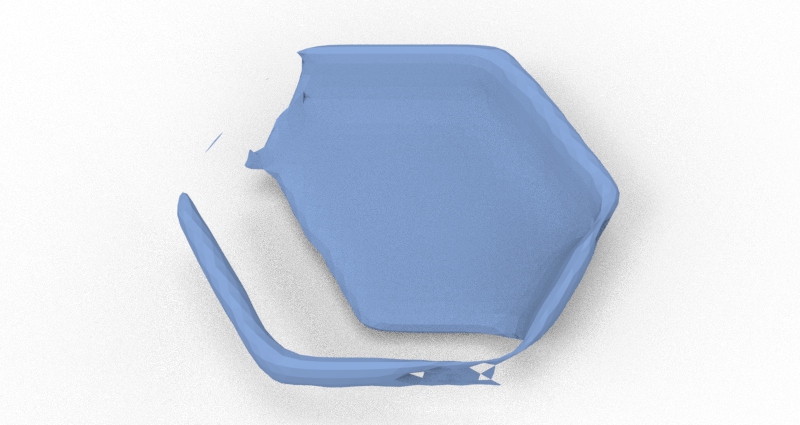}}
	\subfigure[]{
		\includegraphics[width=0.225\linewidth]{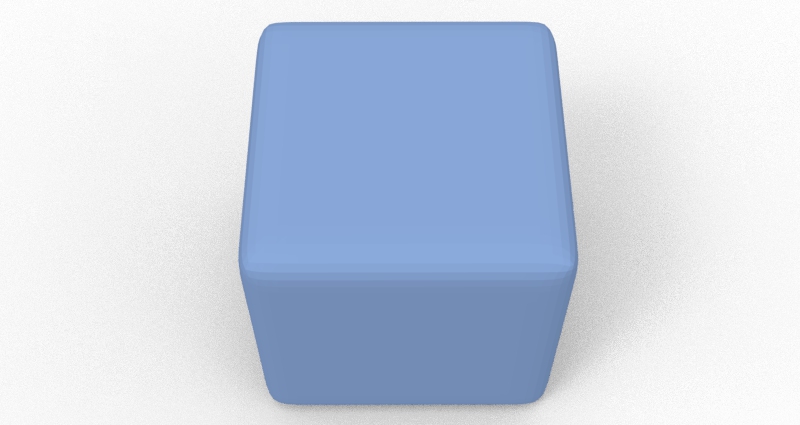}}
	\subfigure[]{
		\includegraphics[width=0.225\linewidth]{figure/ablation/our.jpg}}
	\subfigure[]{
		\includegraphics[width=0.225\linewidth]{figure/ablation/gt.jpg}}
	\caption{Visual comparison of the results with or without \(L_v\) and \(L_s\). (a) Partial input (b) w/o \(L_v\) and \(L_s\). (c) w/ \(L_v\) and \(L_s\). (d) Ground truth. The model tends to fill most voxels with foreground values without \(L_v\) and \(L_s\).}
	\label{ablation_visual_loss}
\end{figure}

\begin{figure}[htbp]
	\centering  
	\subfigbottomskip=1pt 
	\subfigure{
		\includegraphics[width=0.3\linewidth]{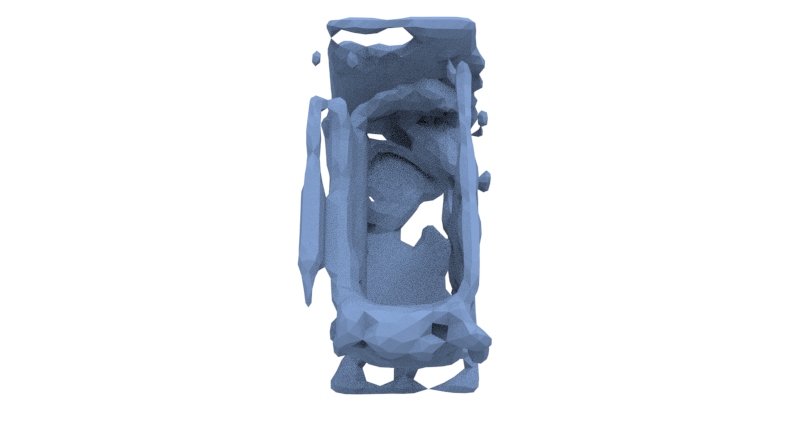}}
	\subfigure{
		\includegraphics[width=0.3\linewidth]{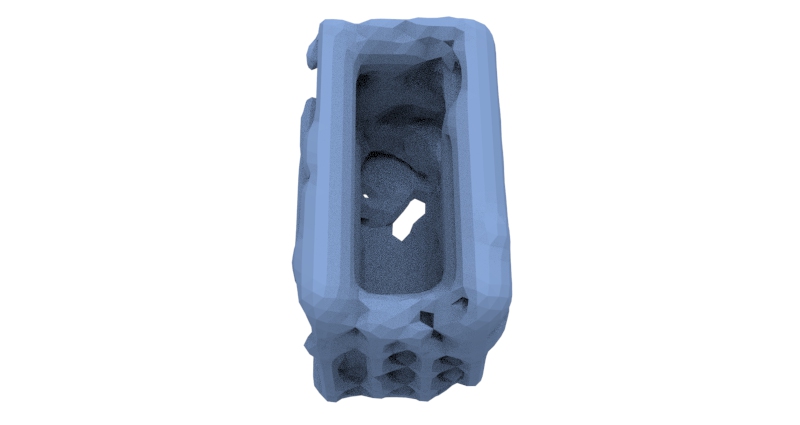}}
	\subfigure{
		\includegraphics[width=0.3\linewidth]{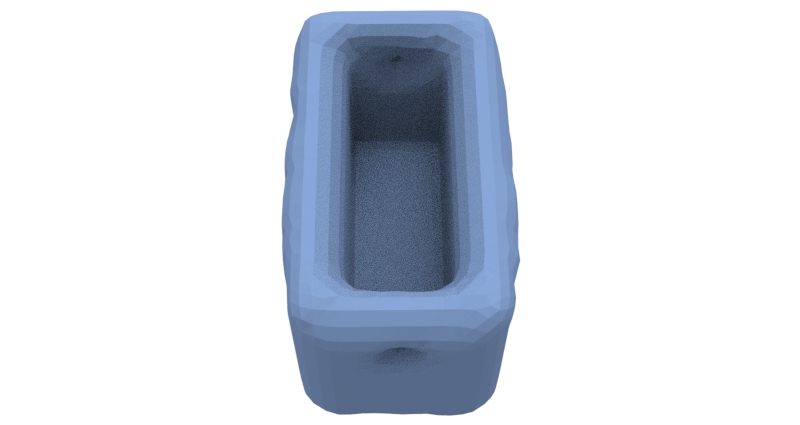}}
	\\
        \setcounter{subfigure}{0}
	%\quad
	\subfigure[]{
		\includegraphics[width=0.3\linewidth]{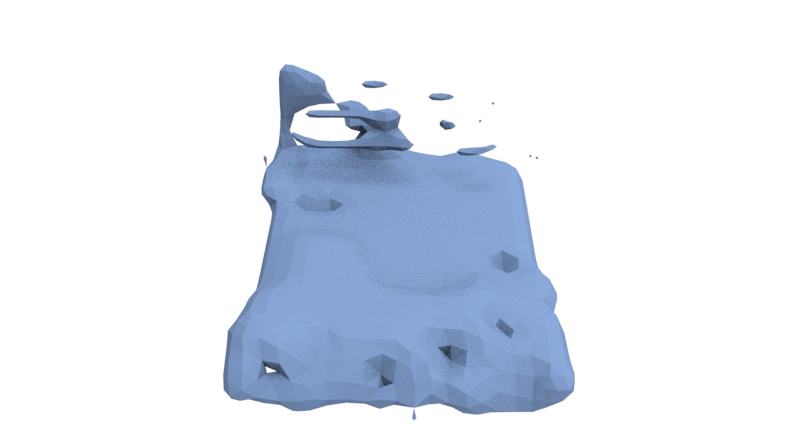}}
	\subfigure[]{
		\includegraphics[width=0.3\linewidth]{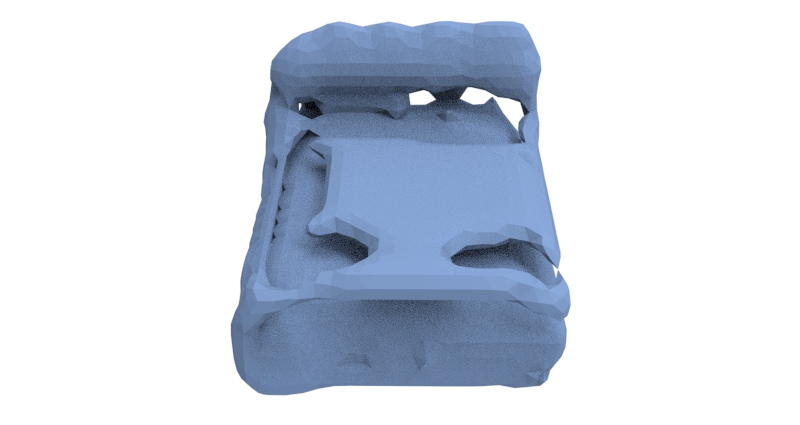}}
	\subfigure[]{
		\includegraphics[width=0.3\linewidth]{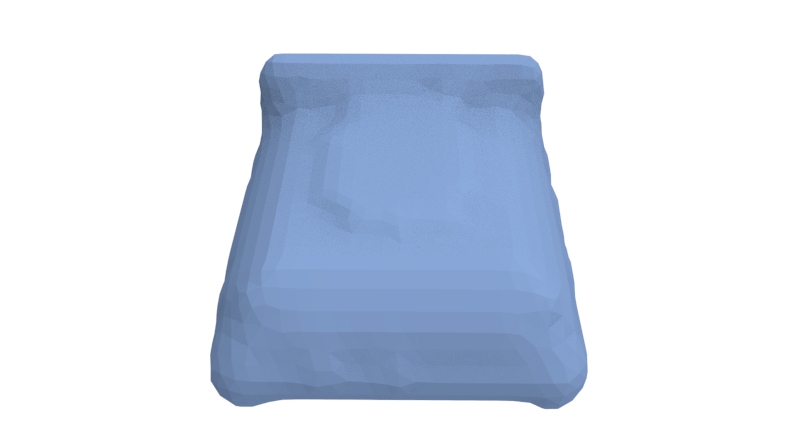}}
	\caption{Visualization of the coarse shapes outputted by CoSL and the results refined by CaSR. (a) Coarse shapes outputted by CoSL (b) Results after refinement with \(\lambda =0.5\). (c) Ground truth with complete shape. Note that this figure only shows the missing parts of the coarse shapes and refined shapes since we only use the missing parts of the coarse shapes as weak supervision. The model tends to reconstruct shapes that are more reasonable and close to the ground truths rather than the coarse shapes.}
	\label{ablation_visual_coarse}
\end{figure}

\begin{table*}[htbp]
	\centering
	\renewcommand\arraystretch{1.25}
	\caption{\revise{Comparison of results using different sizes and construction methods of the prior bank. The reported metric is IoU\(\uparrow\). ``Random" denotes that the priors are randomly sampled from the training set. ``Novel" indicates that the priors come from novel categories not seen in the training set. The best results of each stage are highlighted in bold.}}
	\label{prior_ablation}
	\begin{tabular}{c|c|c|c|c|c|c|c}
		\toprule[1.2pt]
		Stage & Prior size & Prior source & Average & Bag & Lamp & Bathtub & Bed \\ \hline 
            \multirow{7}*{CoSL} &
		32  & Mean-shift & 0.385 & 0.376 & 0.341 & 0.474 & 0.350\\
		~ & 64 & Mean-shift & 0.395 & 0.374 & 0.355 & 0.484 & 0.365 \\
		~ & 112 & Mean-shift & \textbf{0.420} & \textbf{0.391} & \textbf{0.380} & \textbf{0.525} & \textbf{0.383} \\ 
            ~ & 150 & Mean-shift &0.394 & 0.386 & 0.345 & 0.488 & 0.357 \\ 
		~ & 112 & Random \#1 & 0.382 & 0.358 & 0.341 & 0.480 & 0.348\\   
            ~ & 112 & Random \#2 & 0.399 & 0.370 & 0.368 & 0.503 & 0.356 \\
            ~ & 112 & Novel \#1 & 0.389 & 0.366 & 0.353 & 0.482 & 0.356 \\
            ~ & 112 & Novel \#2 & 0.392 & 0.370 & 0.357 & 0.484 & 0.356 \\
            \hline
            \multirow{3}*{CaSR} &
		32 & Specific & 0.426 & 0.439 & 0.358 & 0.525 & 0.383\\
		~ & 64 & Specific & 0.437 & 0.442 & 0.360 & 0.525 & 0.420 \\
            ~ & 112 & Specific & \textbf{0.450} & \textbf{0.450} & \textbf{0.380} & \textbf{0.542} & \textbf{0.426} \\
            ~ & 150 & Specific & 0.437 & 0.441 & 0.358 & 0.525 & 0.424 \\
             \bottomrule[1.2pt]
	\end{tabular}
\end{table*}

\subsubsection{Effects of the components of CaSR}
As CaSR plays a critical role in our model, we conducted ablation experiments to verify the impact of each component in CaSR, including \(L_s\) and \(L_v\) of voxel-based partial matching loss, the weight \(\lambda\) of the coarse shape loss and the priors. As shown in Table \ref{component}, without \(L_s\) and \(L_v\), the model's performance is significantly poor, resulting in an average IoU of less than 0.1. Upon adding \(L_v\), the IoU increases slightly. However, after adding \(L_s\), the IoU improves nearly three times. That means both \(L_s\) and \(L_v\) have an impact on the model's performance, with \(L_s\) having a considerably larger impact. In order to visually demonstrate the effect of \(L_s\) and \(L_v\), we present some qualitative results before and after incorporating \(L_s\) and \(L_v\). As Figure \ref{ablation_visual_loss} shows, the model tends to occupy most of the voxel grids without \(L_s\) and \(L_v\). From our perspective, this behavior can be attributed to the lack of occupancy constraints, leading the model to predict most voxels as foregrounds in order to cover all objects' shapes and thereby minimize the loss.

\begin{table*}[htbp]
	\centering
	\renewcommand\arraystretch{1.25}
	\caption{\revise{Quantitative results at a resolution of 64 on the ShapeNet dataset. The best results under each metric are highlighted in bold. ``$\uparrow$" (resp. ``$\downarrow$") indicates the higher (resp. lower), the better.}}
	\label{result_res64}
	\begin{tabular}{c|c|c|c|c|c|c|c|c|c|c}
		\toprule[1.2pt]
		Metric & Methods             & Average & Bag & Lamp & Bathtub & Bed & Basket	& Printer	& Laptop	& Bench \\ \hline 
            \multirow{2}*{IoU $\uparrow$} &
		 PatchComplete \cite{raopatchcomplete}       & 0.562 & 0.490 & 0.478 & 0.610 & 0.504 & 0.623 & 0.517 & 0.676 & 0.601\\   
            ~ & Ours & \textbf{0.598} & \textbf{0.530} & \textbf{0.511} & \textbf{0.662} & \textbf{0.527} & \textbf{0.649} & \textbf{0.544} & \textbf{0.711} & \textbf{0.650} \\ \hline
            \multirow{2}*{F1 $\uparrow$} &
             PatchComplete \cite{raopatchcomplete}       & 0.710 & 0.651 & 0.638 & 0.753 & 0.666 & 0.751 & 0.676 & 0.803 & 0.745 \\
            ~ & Ours & \textbf{0.739} & \textbf{0.686} & \textbf{0.666} & \textbf{0.791} & \textbf{0.685} & \textbf{0.769} & \textbf{0.700} & \textbf{0.827} & \textbf{0.783} \\ \hline
            \multirow{3}*{CD $\downarrow$}  &
             PatchComplete \cite{raopatchcomplete}    & 2.65 & 2.54 & 3.47 & 2.29 & \textbf{3.07} & \textbf{2.93} & \textbf{3.08} & \textbf{2.29} & 1.51  \\
             ~ & AnchorFormer \cite{chen2023anchorformer}    & 3.53 & 3.14 & 5.07 & 2.85 & 4.33 & 3.53 & 3.83 & 3.12 & 2.37  \\
            ~ & Ours & \textbf{2.59} & \textbf{2.39} & \textbf{3.20} & \textbf{2.19} & 3.12 & 2.94 & 3.09 & 2.39 &  \textbf{1.36}
			 \\ \bottomrule[1.2pt]
	\end{tabular}
\end{table*}

\begin{figure}[htbp]
	\centering  
	\subfigbottomskip=1pt 
	\subfigure{
		\includegraphics[width=0.225\linewidth]{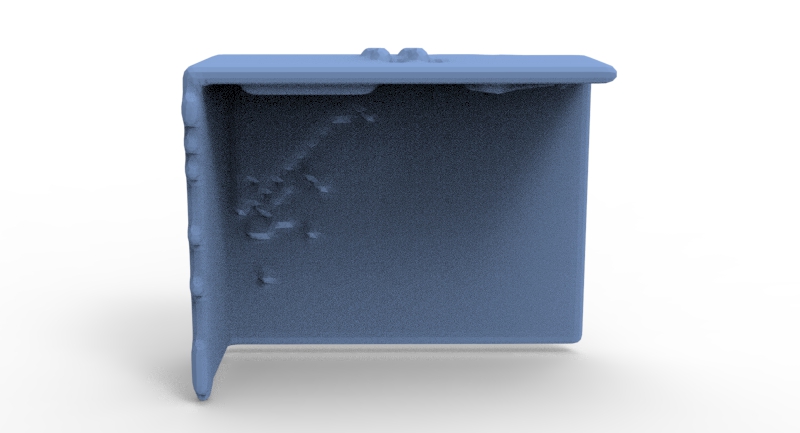}}
	\subfigure{
		\includegraphics[width=0.225\linewidth]{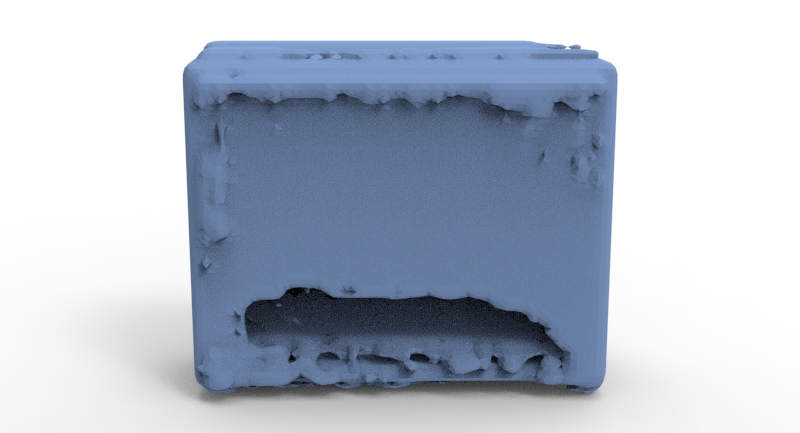}}
	\subfigure{
		\includegraphics[width=0.225\linewidth]{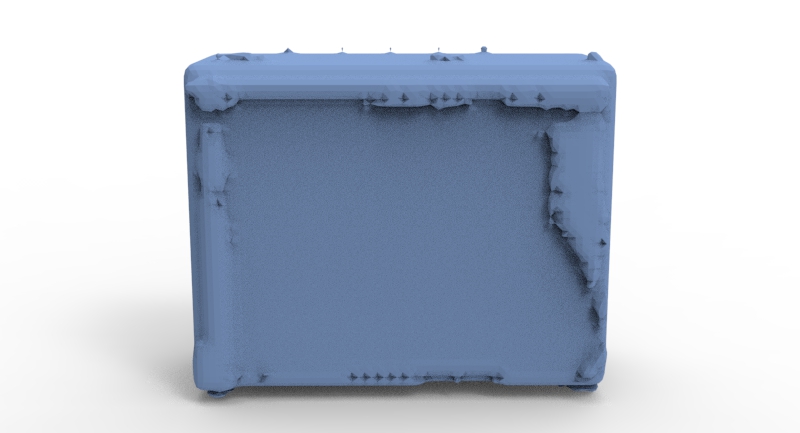}}
	\subfigure{
		\includegraphics[width=0.225\linewidth]{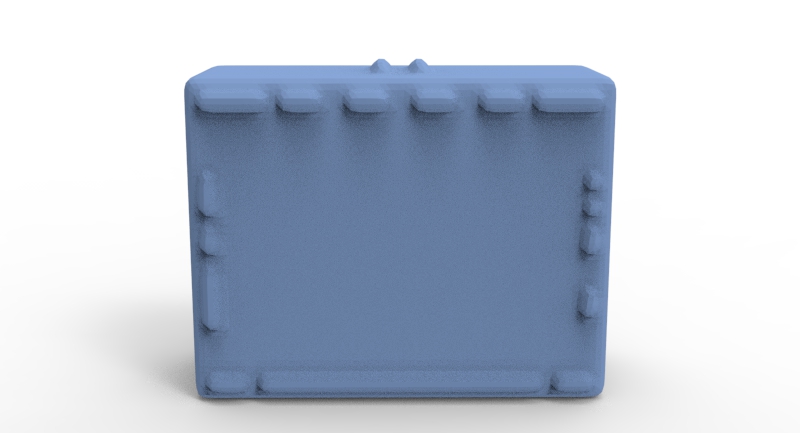}}
	\\
        \subfigure{
		\includegraphics[width=0.225\linewidth]{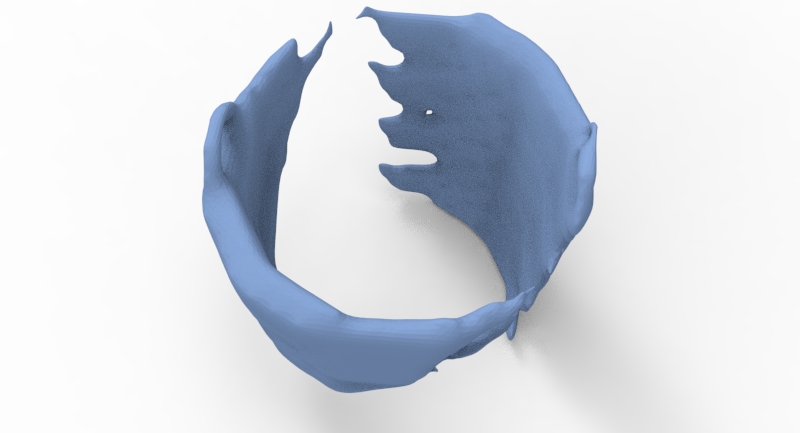}}
	\subfigure{
		\includegraphics[width=0.225\linewidth]{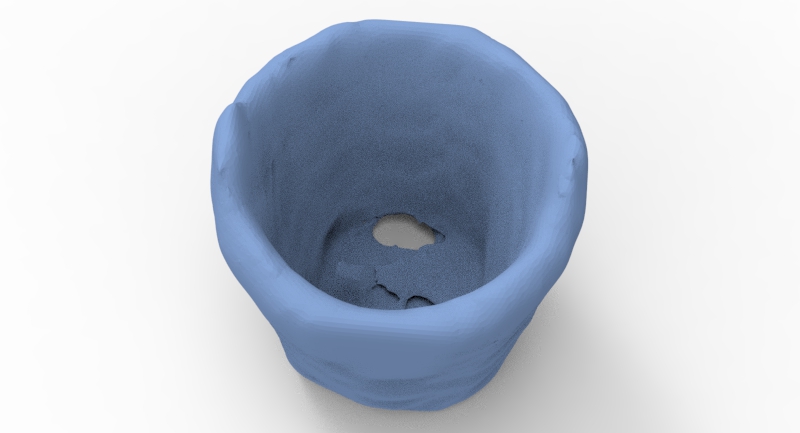}}
	\subfigure{
		\includegraphics[width=0.225\linewidth]{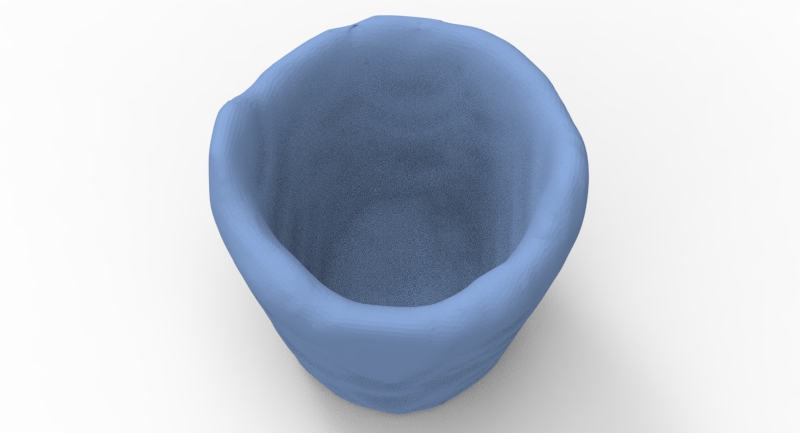}}
	\subfigure{
		\includegraphics[width=0.225\linewidth]{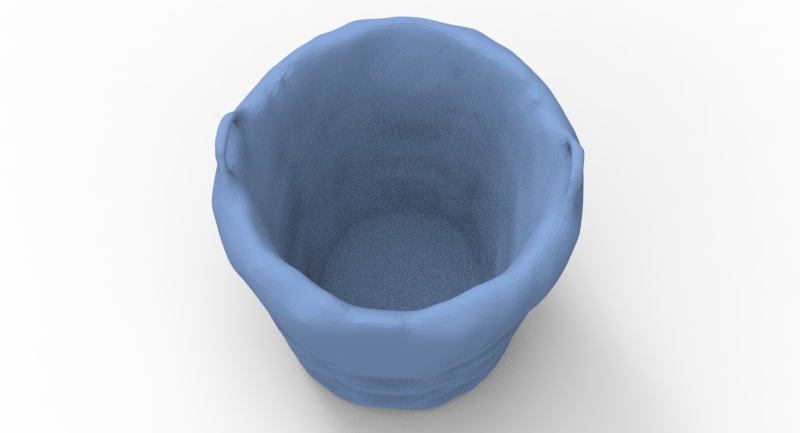}}
	\\
        \subfigure{
		\includegraphics[width=0.225\linewidth]{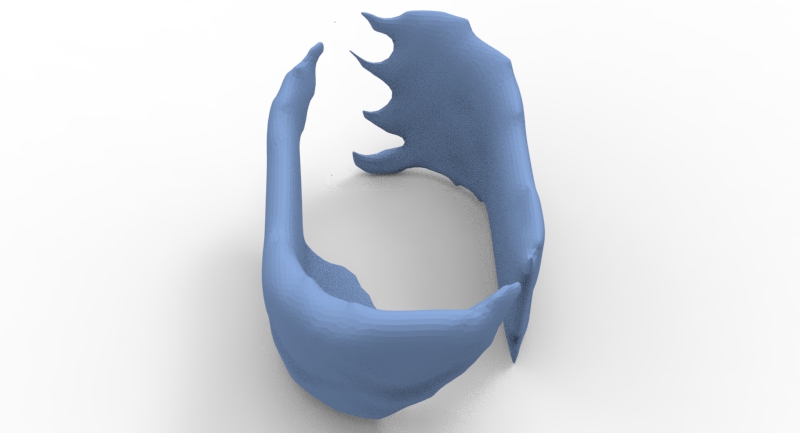}}
	\subfigure{
		\includegraphics[width=0.225\linewidth]{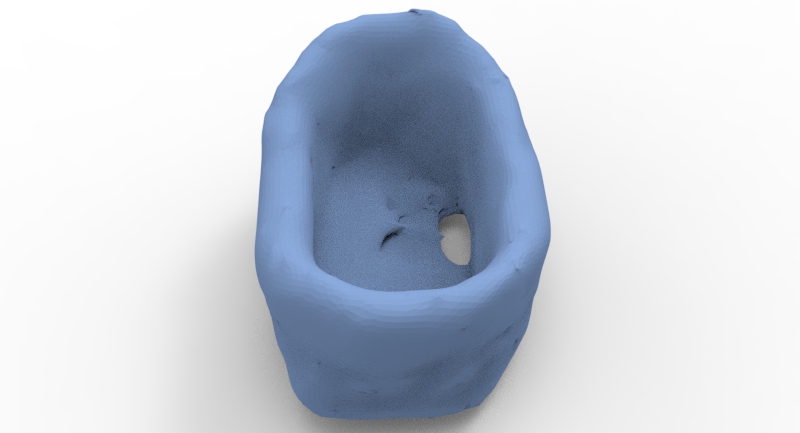}}
	\subfigure{
		\includegraphics[width=0.225\linewidth]{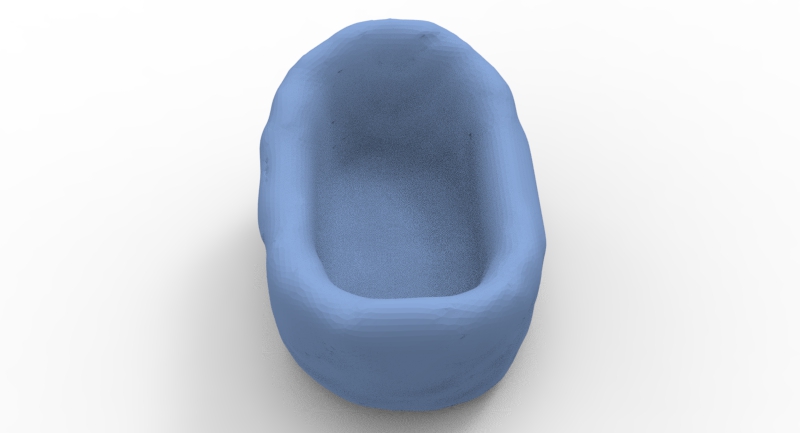}}
	\subfigure{
		\includegraphics[width=0.225\linewidth]{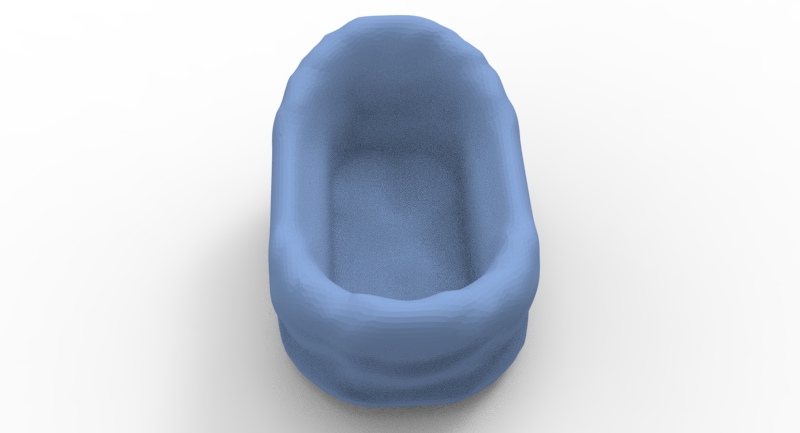}}
	\\
        \setcounter{subfigure}{0}
	%\quad
	\subfigure[]{
		\includegraphics[width=0.225\linewidth]{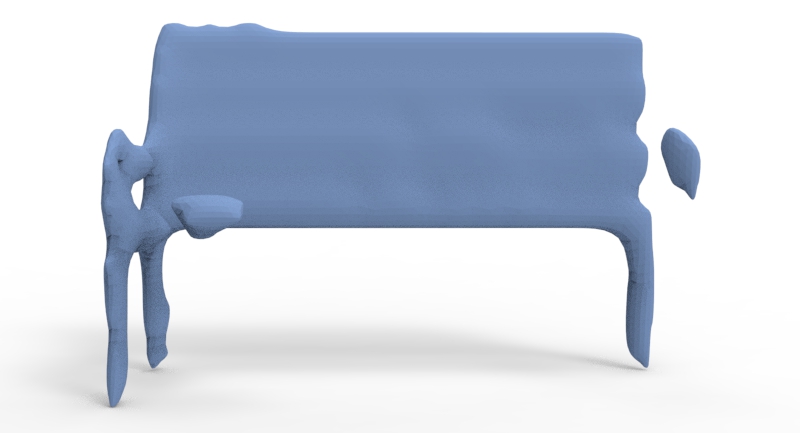}}
	\subfigure[]{
		\includegraphics[width=0.225\linewidth]{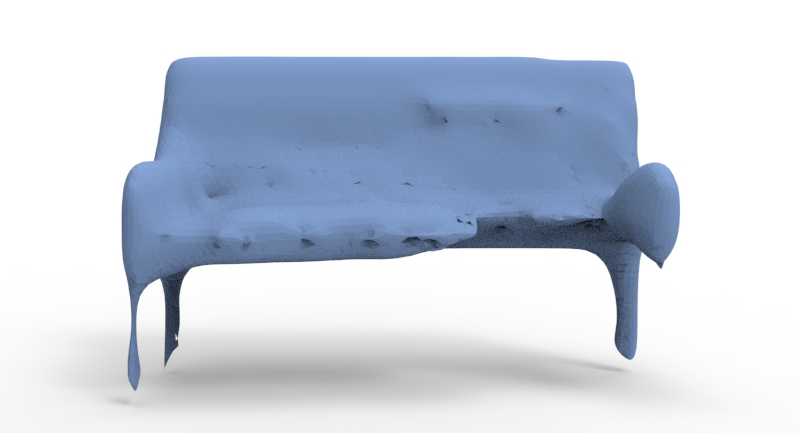}}
	\subfigure[]{
		\includegraphics[width=0.225\linewidth]{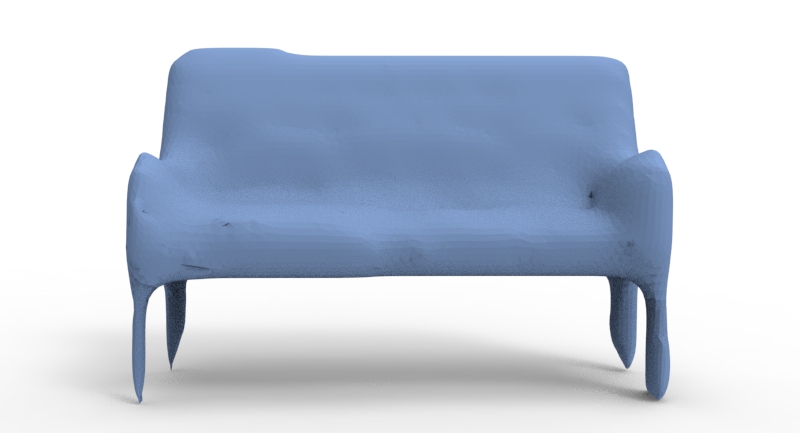}}
	\subfigure[]{
		\includegraphics[width=0.225\linewidth]{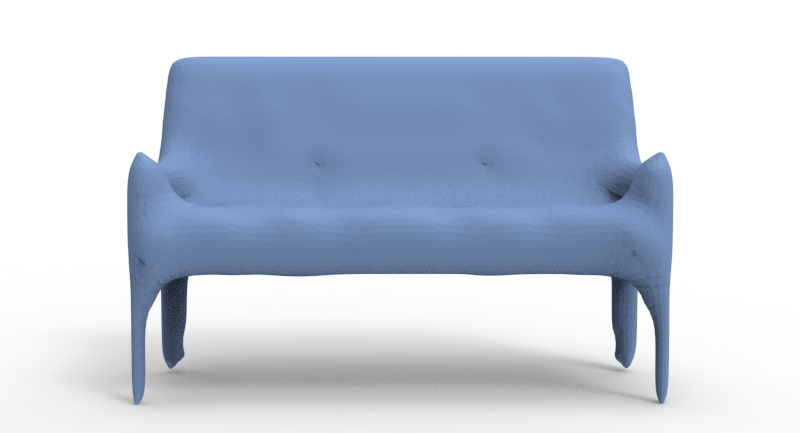}}
	\caption{\revise{Visual comparison of the results at a resolution of 64. (a) Partial input (b) PatchComplete. (c) Ours. (d) Ground truth.}}
	\label{res64_vis}
\end{figure}

In addition, based on the data presented in rows 4, 5, and 7 of Table \ref{component}, it is evident that the model yields optimal results when \(\lambda\) is 0.5. When the supervision of the coarse shape is excessively strong (\(\lambda=1\)) or absent (\(\lambda=0\)), the reconstruction ability of the model is compromised. To provide a more intuitive analysis of the role of the coarse shape loss, we compare the visual results of both the coarse shapes and the outputs refined with \(\lambda=0.5\). As Figure \ref{ablation_visual_coarse} shows, employing the coarse shapes as weak supervision resulted in a more precise prediction. Based on the comprehensive quantification and visualization results, using the coarse shape as weak supervision (\(\lambda=0.5\)) effectively enhances the reconstruction ability of the model. 
On the one hand, the coarse shapes can offer effective weak supervision to the model. On the other hand, this weak supervision does not confine the model to reconstruct the same results as the coarse shapes but rather facilitates the model learning to infer shapes that are more accurate and complete.

\begin{figure*}[htbp]
	\centering  
	\subfigbottomskip=1pt 
	%\quad
        \subfigure{
		\includegraphics[width=0.13\linewidth]{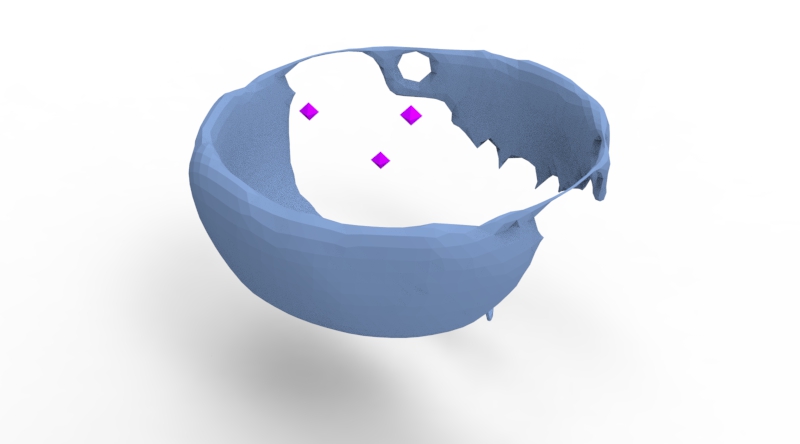}}
	\subfigure{
		\includegraphics[width=0.13\linewidth]{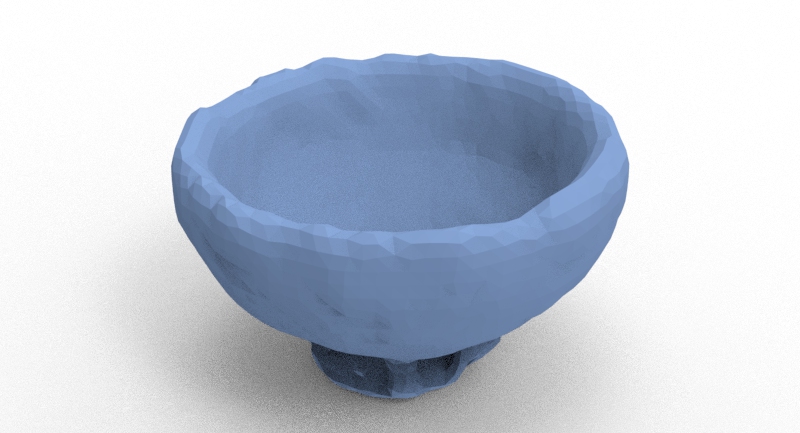}}
	\subfigure{
		\includegraphics[width=0.13\linewidth]{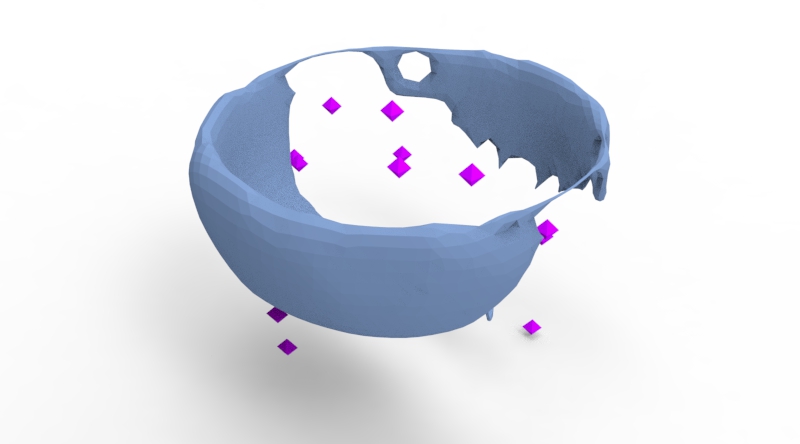}}
	\subfigure{
		\includegraphics[width=0.13\linewidth]{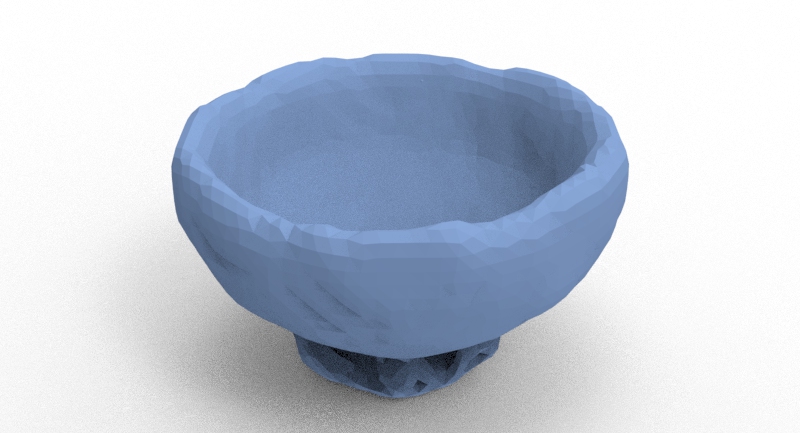}}
        \subfigure{
		\includegraphics[width=0.13\linewidth]{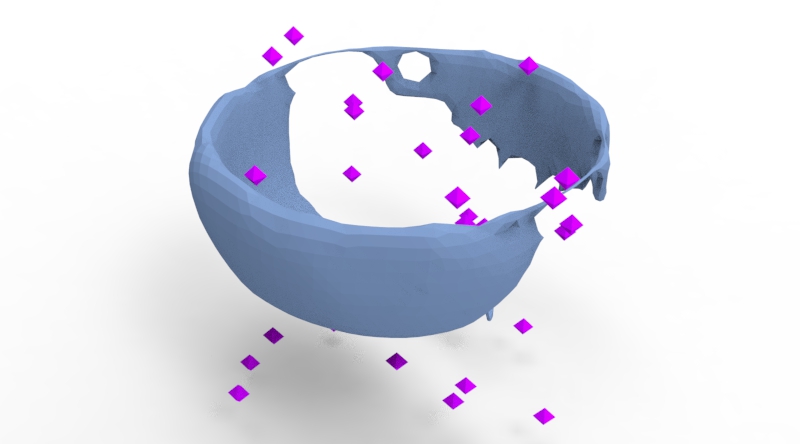}}
	\subfigure{
		\includegraphics[width=0.13\linewidth]{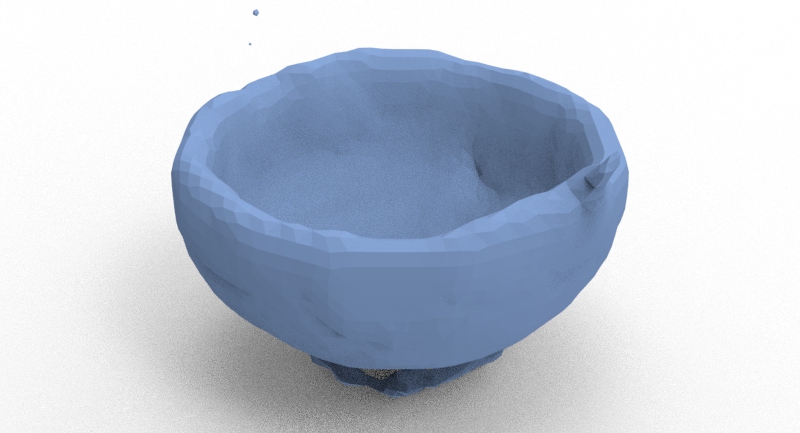}}
	\subfigure{
		\includegraphics[width=0.13\linewidth]{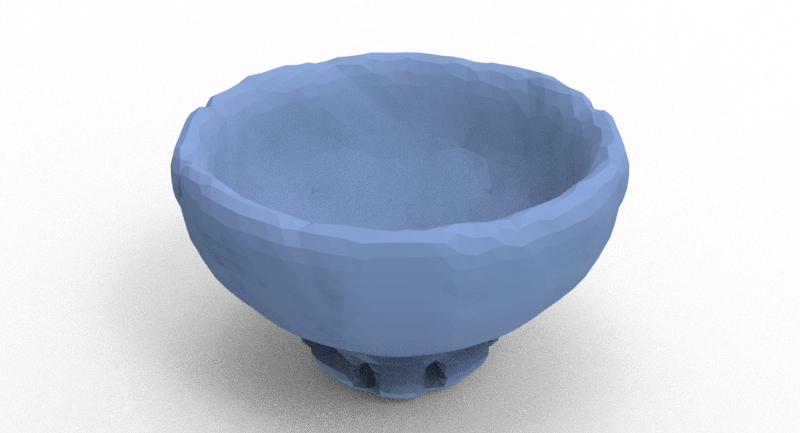}}
	\\
        \setcounter{subfigure}{0}
	\subfigure[]{
		\includegraphics[width=0.13\linewidth]{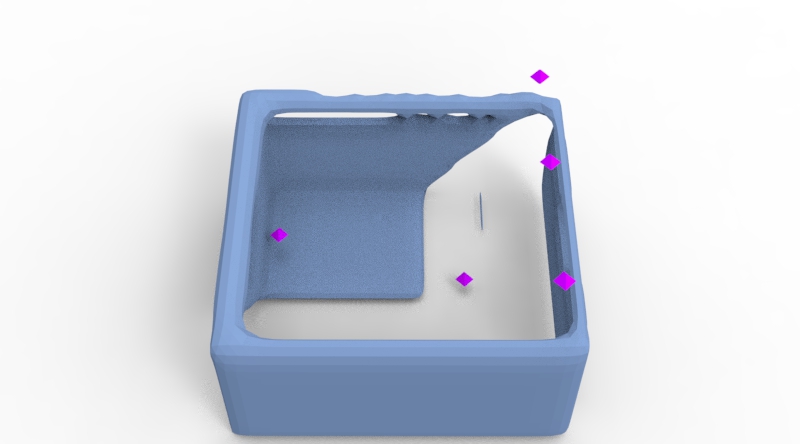}}
	\subfigure[]{
		\includegraphics[width=0.13\linewidth]{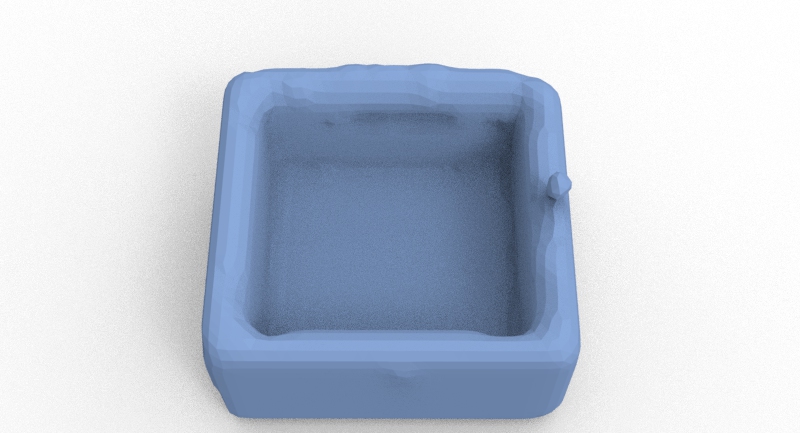}}
	\subfigure[]{
		\includegraphics[width=0.13\linewidth]{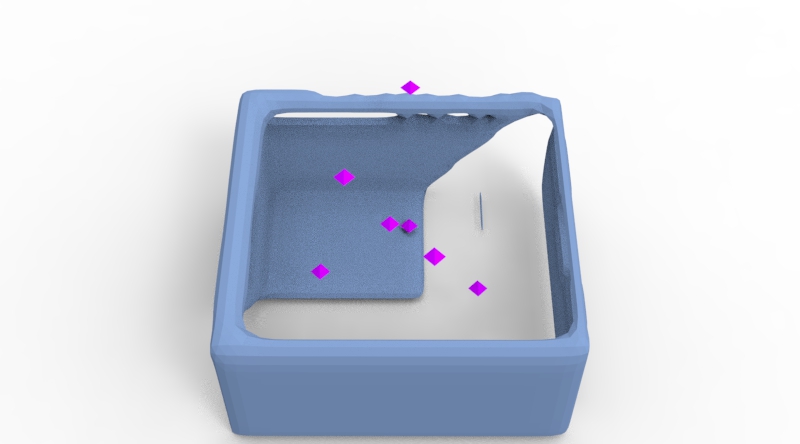}}
	\subfigure[]{
		\includegraphics[width=0.13\linewidth]{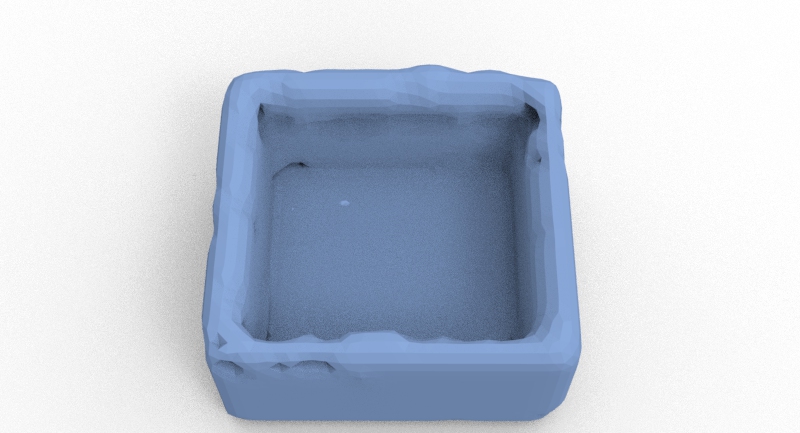}}
        \subfigure[]{
		\includegraphics[width=0.13\linewidth]{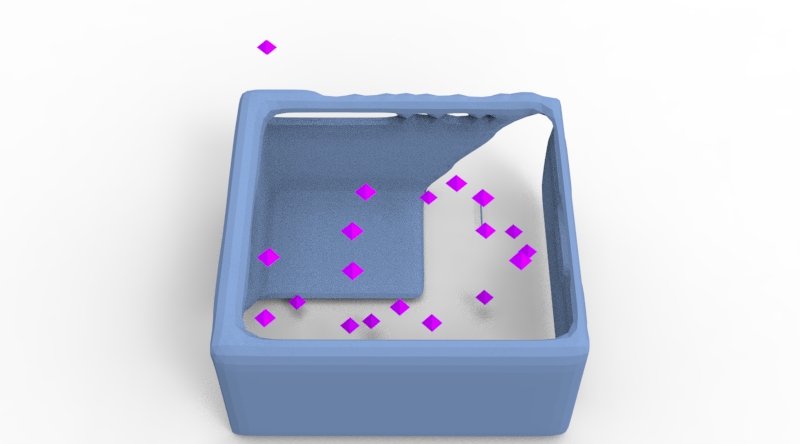}}
        \subfigure[]{
		\includegraphics[width=0.13\linewidth]{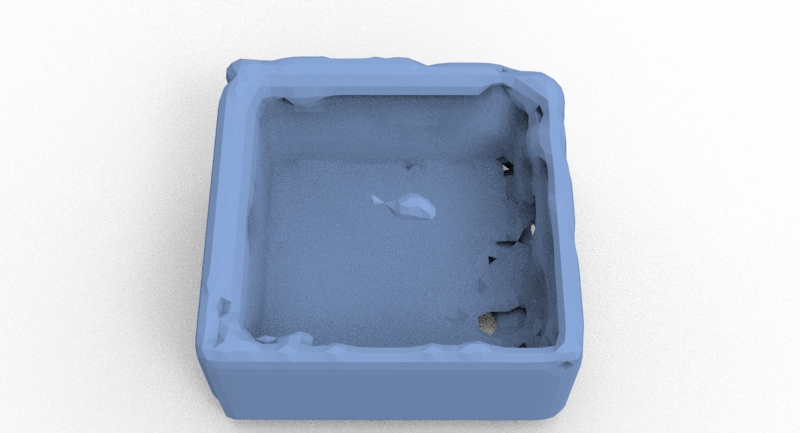}}
        \subfigure[]{
		\includegraphics[width=0.13\linewidth]{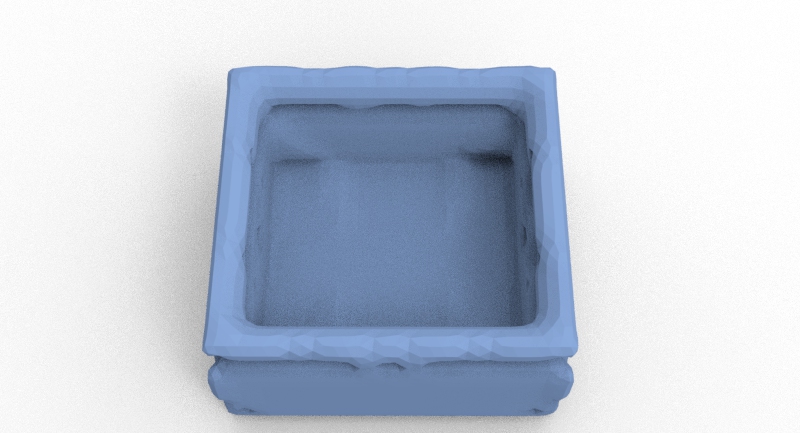}}
	\caption{\revise{Visual comparison of the results with different noise levels. Noises are highlighted with purple points. (a)$\sim$(b) Noisy input with \(\sigma=0.14\) and its output. (c)$\sim$(d) Noisy input with \(\sigma=0.15\) and its  output. (e)$\sim$(f) Noisy input with \(\sigma=0.16\) and its  output. (g) Ground truth.}}
	\label{noise_vis}
\end{figure*}

Besides, from the last two rows of Table \ref{component}, we can see that the IoU improves by 0.014 when using our category-specific priors instead of the fixed priors designed by \cite{raopatchcomplete}. That means our category-specific priors are better suited for refining data from different categories. 
\revise{
Furthermore, we investigated the impact of both the size and construction method of the prior bank on the results. First, we varied the size of the prior bank to 32, 64, and 150, respectively, and conducted experiments across both the CoSL and CaSR stages. As detailed in Table \ref{prior_ablation}, using 112 priors achieves the best results for both stages. Reducing the number of priors leads to performance degradation, while increasing the number to 150 also results in decreased performance. In our opinion, this is because the local structures provided by the priors already cover most shapes with 112 objects. Adding more priors does not provide additional useful structural priors for the model but rather introduces irrelevant information into the learning process, which increases the difficulty of computing prior weights and weakens the role of important priors.
In addition, we used different methods to construct the prior bank and explore its impact. Concretely, on the one hand, we randomly sampled two sets of priors from the training set. On the other hand, we selected two groups of priors from four categories in the test set (one from basket, printer, and another from laptop, bench) and tested on the remaining four categories. Compared to the original prior bank, these priors differ significantly from the shape distribution of the training set. Table \ref{prior_ablation} shows that the random priors result in significantly poorer performance compared to the priors obtained from mean-shift clustering. This performance gap arises because clustering ensures more diverse local shapes among priors. Additionally, from rows 3, 7, and 8 of Table \ref{prior_ablation}, it can be seen that when the shape distribution of priors differs greatly from that of the training set, the performance of the CoSL's model significantly decreases. Furthermore, according to the ablation results presented in Table \ref{component}, CaSR also demonstrates a performance decrease when using priors (fixed priors) that differ from the distribution of the target shapes. We believe that due to significant differences in local structures, the model learns few shape-related structures between partial observations and priors through cross-attention, making it difficult to accurately reconstruct complete shapes.}

\begin{table}[htbp]
	\centering
    \renewcommand\arraystretch{1.25}
	\caption{\revise{Effects of the diversity between the training set and test set. ``Diverse split" denotes the group with the most significant shape differences between the training set and test set. ``Average" means the average IOU of the control group.}}
	\label{diversity_comparison}
	\begin{tabular}{c|c|c|c|c|c}
		\toprule[1.2pt]
            \multicolumn{2}{c|}{Shape diversity} & \multicolumn{4}{c}{Noise diversity} \\ \hline
		          Diverse split   & Average  & Clean & \(\sigma=0.14\) & \(\sigma=0.15\) & \(\sigma=0.16\)\\ \hline     
		   0.426 & 0.429 & 0.441 & 0.431 & 0.410 & 0.370 \\ \bottomrule[1.2pt]
	\end{tabular}
\end{table}

\subsubsection{Effects of the resolution}
\revise{
Higher resolutions usually preserve finer details. To evaluate our model's performance on high-resolution data, we conducted experiments using data at a resolution of \(64\times 64\times 64\). Table \ref{result_res64} presents a comparison of the quantitative results among our approach, PatchComplete, and AnchorFormer. Note that converting the voxel grids of \(64\times 64\times 64\) resolution into point clouds yields an average of nearly 24,000 points. For a fair comparison, we set the output point number of AnchorFormer to 16,384 and randomly sampled our results to the same number of points when calculating the CD value. As indicated in Table \ref{result_res64}, our method significantly outperforms PatchComplete and AnchorFormer at the \(64\times 64\times 64\) resolution. Figure \ref{res64_vis} visually illustrates these results, showing that our method excels in preserving input details while generating more complete and reasonable shapes.}

\subsubsection{Evaluation of training/test set diversity}
\revise{We also investigated the impact of diversity between the training and test sets, including the shape diversity and noise levels, on the model's performance. Specifically, we first employed an auto-encoder to learn shape codes for all objects. Subsequently, we utilized the K-means algorithm to cluster the average shape codes of the 26 categories. From these clusters, we selected two groups of data with the most significant shape differences as the training set and the test set, respectively. The training set contains 18 categories, i.e., trash bin, bag, bookshelf, cabinet, chair, dishwasher, display, faucet, file cabinet, guitar, lamp, laptop, microwave, piano, printer, stove, table, washing machine, while the test set contains 8 categories, i.e., basket, bed, pots, bathtub, bench, bowl, keyboard, sofa. To ensure a more objective and fair comparison, we also set up a control group. The control group consists of three pairs of training and test sets obtained through random combinations of categories, including the initial category split in Section \ref{dataset_metric}. We compared the IOU of the diverse-shape group with the average IOU of the control group. As shown in Table \ref{diversity_comparison}, the gap between these two groups is only 0.03. This indicates that our method is highly robust to shape diversity between the training set and the test set.}

\revise{
In addition, considering that real-world partial scans may contain noises, we introduced various levels of Gaussian noise (\(\mu=0,\sigma=\{0.14,0.15,0.16\}\)) to the partial input of the test set to verify its impact on our method. As shown in Table \ref{diversity_comparison} and Figure \ref{noise_vis}, the performance of our method is not significantly affected under conditions of low to moderate noise. Significant performance degradation is observed under extremely severe noise levels. This finding validates the robustness of our method in handling noisy data.}

\section{Conclusion}
\label{conclusion}
In this paper, we present a novel approach to address the challenging problem of completing 3D shapes from unseen categories. Unlike the previous method that only relies on prior information from seen categories, our method takes advantage of both shape priors from seen categories and partial scans from unseen categories to learn complete shapes in a weakly-supervised manner. We introduce a prior-assisted shape learning network featuring a multi-scale pattern correlation module to reconstruct coarse shapes. This model effectively captures the correlations between the local patterns of the partial input and priors at various scales and uses them to infer missing structures. Upon refinement with the category-specific priors, partial scans, and the voxel-based partial matching loss, the reconstructed shapes are more complete and reasonable. Notably, the voxel-based partial matching loss paves the way for self-supervised shape completion of voxel-based shapes. In the future, we will try to extend our method to shape completion tasks with more types of data, such as point clouds, meshes, and continuous implicit functions. We hope that this study can inspire future research in the field of 3D shape reconstruction and scene understanding.
 
\ifCLASSOPTIONcaptionsoff
  \newpage
\fi

\bibliographystyle{IEEEtran}
%\bibliography{mybibfile}

%\if 0
%\end{IEEEbiography}


% \fi 


\end{document}